\documentclass[]{infiniai}
\usepackage{microtype}
\usepackage{amsfonts}
\usepackage{graphicx}
\usepackage{booktabs}
\usepackage{wrapfig}
\usepackage{amsmath}
\usepackage{amsthm}
\usepackage{subcaption}

\usepackage{algpseudocode}
\usepackage[linesnumbered,lined,boxed,commentsnumbered,ruled,longend]{algorithm2e}
\usepackage[capitalize,noabbrev]{cleveref}
\theoremstyle{plain}

\theoremstyle{definition}

\theoremstyle{remark}

\usepackage{enumitem}

\newcommand{\sparselaw}{\textsc{Sparse Kinetics}\xspace}
\newcommand{\Law}{\textsc{Kinetics}\xspace}
\newcommand{\law}{\textsc{Kinetics}\xspace}

\newcommand{\bon}{{Best-of-$N$}\xspace}

\newcommand{\longcot}{Long-CoTs\xspace}

\definecolor{lavenderpink}{rgb}{0.95, 0.85, 0.95}
\definecolor{softlavender}{rgb}{0.64, 0.50, 0.68}
\title{Kinetics: Rethinking Test-Time Scaling Laws}

\author{Ranajoy Sadhukhan$^*$}
\author{Zhuoming Chen$^*$}
\author{Haizhong Zheng}
\author{Yang Zhou, Emma Strubell, Beidi Chen}
\contact{\{rsadhukh,zhuominc,haizhonz,yangzho6,estrubel,beidic\}@andrew.cmu.edu}
\affiliation{Carnegie Mellon University}
\abstract{We rethink test-time scaling laws from a \textit{practical efficiency} perspective, revealing that the effectiveness of smaller models is significantly overestimated. Prior work, grounded in compute-optimality, overlooks critical memory access bottlenecks introduced by inference-time strategies (e.g., Best-of-$N$, long CoTs). 
Our holistic analysis, spanning models from 0.6B to 32B parameters, reveals a new \textsc{Kinetics} scaling law that better guides resource allocation by incorporating both computation and memory access costs. \textsc{Kinetics} suggests that test-time compute is more effective when used on models above a parameter threshold than with smaller ones. A key reason is that in test-time scaling, attention, rather than parameter count,  emerges as the dominant cost factor. Motivated by this, we propose a new scaling paradigm centered on \textit{sparse attention}, which lowers per-token cost and enables longer generations and more parallel samples within the same resource budget. Empirically, we show that sparse attention models consistently outperform dense counterparts, achieving over \textbf{60 points} gains in low-cost regimes and over \textbf{5 points} gains in high-cost regimes for problem-solving accuracy on AIME, encompassing evaluations on state-of-the-art MoEs. These results suggest that sparse attention is essential and increasingly important with more computing invested, for realizing the full potential of test-time scaling where, unlike training, 
accuracy has yet to saturate as a function of computation, and continues to improve through increased generation. 
}
\metadata[Github]{\url{https://github.com/Infini-AI-Lab/Kinetics}}
\metadata[Website]{\url{https://infini-ai-lab.github.io/Kinetics}}

\begin{document}

\maketitle
\begingroup
\renewcommand{\thefootnote}{}%
\footnote{$^*$ indicates equal contributions.}%
\addtocounter{footnote}{-1}%
\endgroup
\begin{figure}[h]
    \centering
    \subfloat[\law scaling law]{
\includegraphics[width=0.48\linewidth]{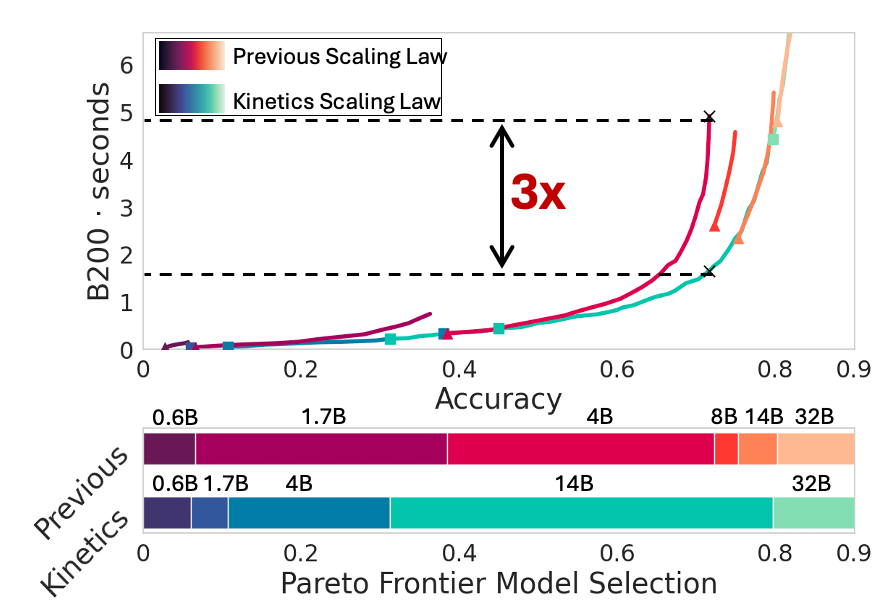}
    \label{fig: scaling law}
    }
  \subfloat[\sparselaw]{
\includegraphics[width=0.48\linewidth]{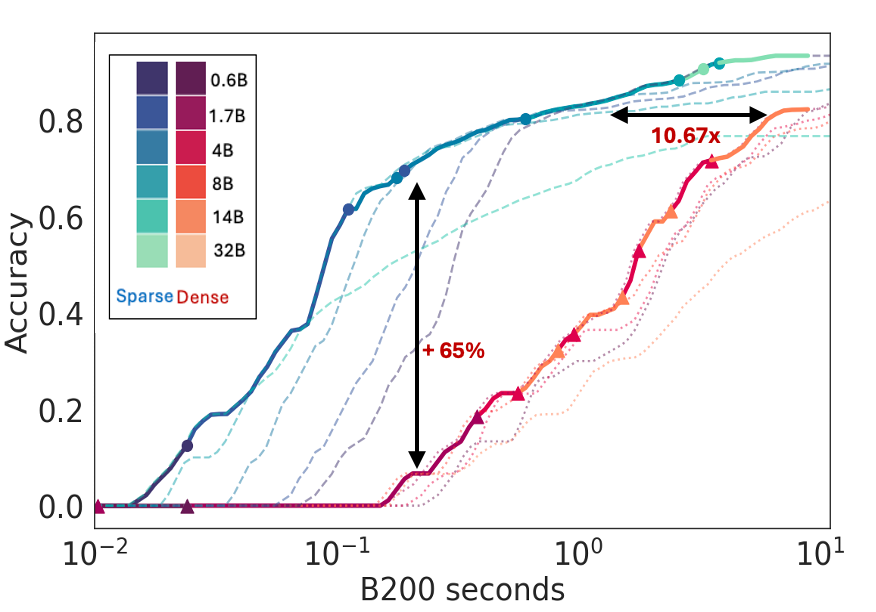}
    \label{fig: sparse attention} 
    }
    \caption{
    \textbf{(a) Pareto Frontier for Qwen3 series on AIME24 (\longcot).} Previous test-time scaling laws~\citep{brown2024large,snell2024scaling,wu2024inference} focus solely on compute optimality, neglecting the significant bottleneck of memory access in long-sequence generation. This leads to suboptimal resource utilization. By incorporating memory access, \law reduces resource demands by up to $3\times$ to achieve the same accuracy. \textbf{(b) Top-$K$ Sparse attention on AIME24 (\bon).} Inspired by \law, we show that \textit{sparse attention} models scale significantly better than dense models, achieving over $50$-point improvements in AIME24 in the low-cost regime and consistently outperforming dense models in the high-cost regime, in addition to substantial efficiency gains. B200 seconds represent the amount of work performed by a single NVIDIA B200 at full utilization per second.
} 
\end{figure}

\section{Introduction}

\begin{figure*}
    \centering
    \subfloat[Attention Cost Dominates]{
\includegraphics[width=0.32\linewidth]{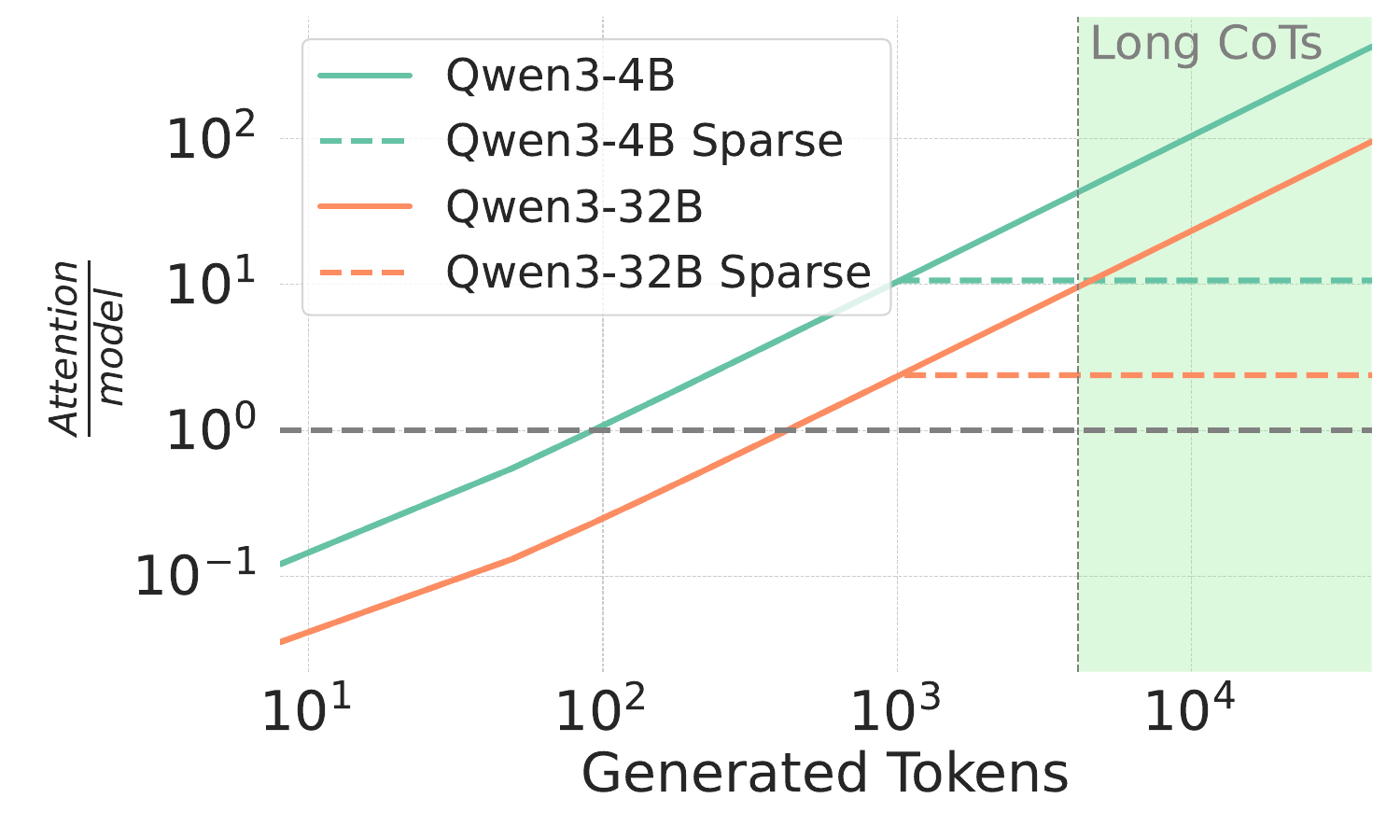}
    \label{fig: Attention-Model-ratio}
    }
    \subfloat[Generated Tokens vs. Cost]{
    \includegraphics[width=0.32\linewidth,height=0.2\linewidth]{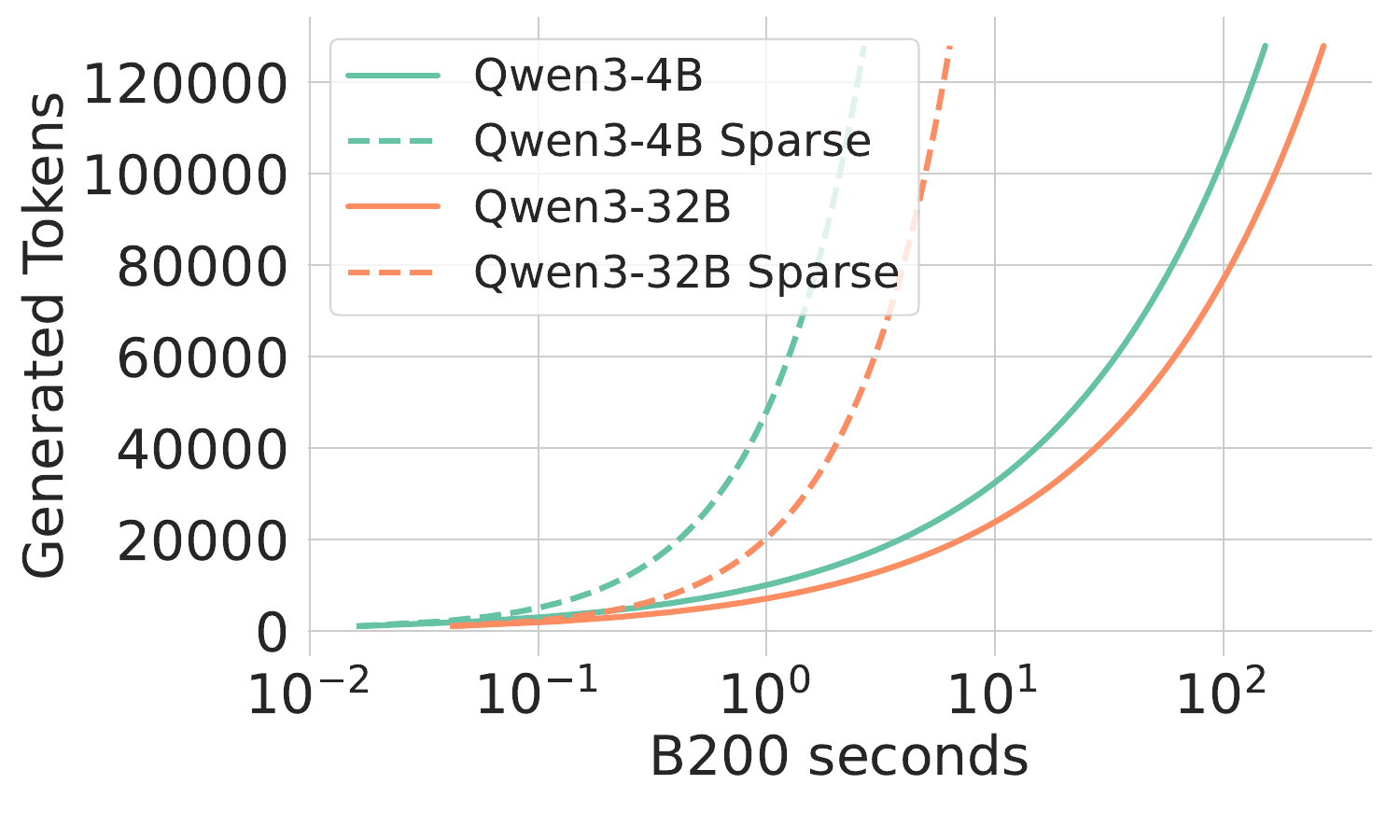}
    \label{fig: latency breakdown}
    }
  \subfloat[Block top-$k$ Attention]{
\includegraphics[width=0.325\linewidth]{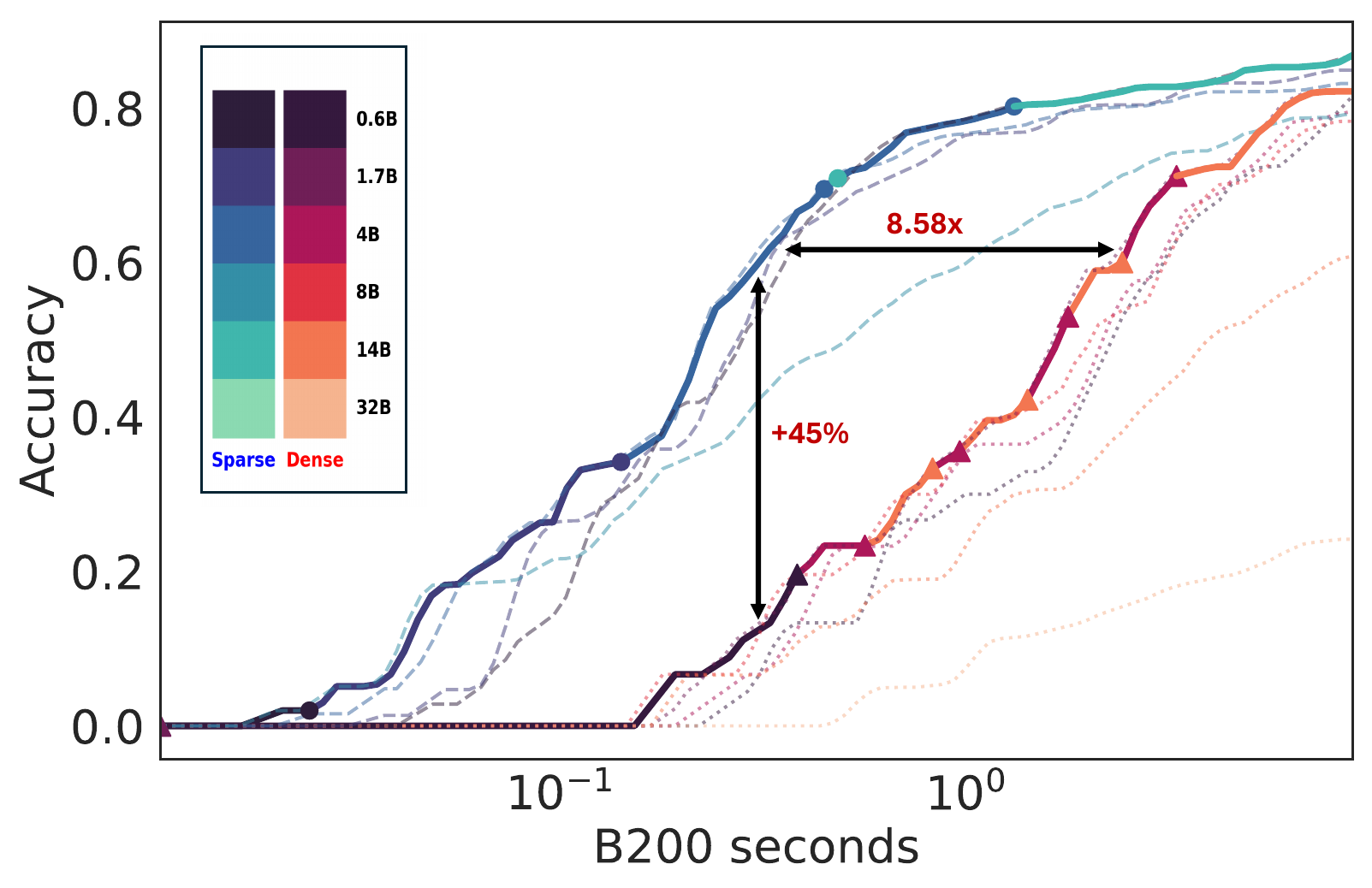}
    \label{fig: Attention-Model-ratio-MoE} 
    }
    \caption{\textbf{(a)} Inference cost is dominated by attention (i.e. \(\text{Softmax}(\frac{qk^T}{\sqrt{d}})v\)), which is 10-1000$\times$ more than model parameter computation (linear modules, including those in MLPs and $W_{Q},W_{K},W_{V}, W_{O}$), sparse attention fundamentally mitigates this bottleneck.
    \textbf{(b)} Under the same resource constraints, sparse attention can generate significantly more tokens than dense models, which has been demonstrated to enhance the effectiveness of test-time scaling. \textbf{(c)} Simple block sparse attention yields substantial gains---improving accuracy by 45 points in the low-cost regime, and achieving equivalent accuracy while using $8.58\times$ fewer resources.}
\end{figure*}

Test-time scaling (TTS) has recently emerged as a powerful strategy (e.g., Best-of-$N$, Long-CoT~\citep{wei2022chain}) for enhancing the reasoning capabilities of large language models (LLMs)~\citep{guo2025deepseek,jaech2024openai,qwq32b}, particularly in scenarios where agents interact with complex environments, e.g., writing code, browsing the web~\citep{nakano2021webgpt,yao2023react} or reinforcement learning (RL) with LLMs-in-the-loop~\citep{huang2022language,driess2023palme,chen2025reinforcement}. These capabilities, however, introduce substantial inference-time costs, making it critical to understand performance scaling in this new paradigm. Existing scaling law studies~\citep{brown2024large,snell2024scaling,wu2024inference} primarily focus on floating-point operations (FLOPs) while ignoring memory access costs, which are often the dominant factor in determining wall-clock latency in TTS regimes. As shown in~\Cref{fig: scaling law}, this gap can lead to sub-optimal deployment decisions. 

In~\Cref{sec:analysis}, we introduce the \law scaling law for TTS, derived from a cost model that explicitly incorporates memory access costs. This new perspective reveals markedly different conclusions about Pareto-optimal strategies for allocating test-time compute (\Cref{fig: scaling law}). Specifically, we find that: (1) prior scaling laws consistently \textbf{overestimate} the effectiveness of small models enhanced with inference-time strategies; and (2) computational resources are best spent first on increasing model size up to a critical threshold (empirically, around 14B parameters) before investing in test-time strategies, such as \bon sampling or \longcot. Guided by \law, our approach yields up to a \textbf{3$\times$} reduction in resource demands to reach the same accuracy on NVIDIA B200 hardware.

Our roofline analysis~\citep{yuan2024llminferenceunveiledsurvey} across a suite of state-of-the-art reasoning models reveals that the shift in optimal test-time compute strategies arises because test-time strategies (e.g., \bon, \longcot) disproportionately increase attention costs rather than parameter costs (\Cref{fig: Attention-Model-ratio}). Our Iso-cost analysis~\citep{kumar2024scaling} shows that the quadratic growth of attention with generation length, combined with the disproportionate scaling of KV memory relative to model parameters, drives a preference for scaling up model size over generations. This imbalance is further exacerbated by MoE architectures~\citep{shazeer2017outrageously,du2021glam,fedus2022switch,llama4modelcard,dai2024deepseekmoe,jiang2024mixtral}, which reduce active parameter count without alleviating attention overhead.

Building on the above analyses, in~\Cref{sec:s3} we introduce a new scaling paradigm, centered on \textbf{sparse} attention, which fundamentally reshapes the scaling law and significantly enhances the scalability of TTS (\Cref{fig: sparse attention,fig: latency breakdown}). According to our \sparselaw scaling law, computational resources are best allocated to test-time strategies rather than reducing sparsity. As more computing is invested at test time, high sparsity becomes increasingly critical to fully leverage the benefits of these strategies. Guided by this principle, we find that sparse attention increases problem-solving rates by up to \textbf{60 points} in the low-cost regime and over \textbf{5 points} in the high-cost regime on AIME (\Cref{fig: sparse attention}) and LiveCodeBench, encompassing state-of-the-art MoEs.

In~\Cref{sec:experiments}, we demonstrate the practicality of \sparselaw using a simple block-sparse attention mechanism (\Cref{fig: Attention-Model-ratio-MoE}). This approach achieves up to a \textbf{$11.2\sim26.2\times$} speedup on H200 GPUs. While sparsity has traditionally been employed either for regularization in small models~\citep{tibshirani1996regression, molchanov2017variational} or to reduce computation in over-parameterized networks~\citep{mishra2021accelerating,chen2021scatterbrain,hoefler2021sparsity,dao2021pixelated,frantar2023sparsegpt,liu2023deja}, our work introduces a fundamentally different perspective: \textbf{sparsity as a central enabler of efficient and scalable test-time compute}. In contrast to pretraining, where scaling is exhibiting diminishing returns~\citep{Ilya}, TTS continues to benefit from increased token generation and more optimized inference paths. Our work highlights the importance of considering hardware concerns alongside model architecture in order to develop a practical understanding of scaling laws. We hope this study will guide and encourage future co-design of model architectures, test-time strategies, and hardware systems to fully unlock the next wave of LLM scaling. 
\section{Related Work and Problem Settings}
\vspace{-0.1cm}
\label{sec: problemformulation} 
In this section, we first review several lines of related work relevant to \law. Then we introduce a cost model accounting for computation and memory access, followed by a roofline analysis uncovering a key departure from traditional scaling laws. Finally, we outline the experimental setup used in the subsequent analysis. Notation is summarized in~\Cref{tab:notation}.

\textbf{Scaling Laws.} Prior work~\citep{kaplan2020scaling,hoffmann2022training,kumar2024scaling} has extensively examined the scaling laws of pretraining, exploring the trade-off between model size and the number of training tokens under a fixed FLOPs budget. More recently, studies such as~\citep{snell2024scaling,wu2024inference,brown2024large,beeching2024scalingtesttimecompute} have extended this analysis to test time scaling, with a focus on compute-optimality. While these works offer a strong theoretical foundation, they largely overlook the critical bottleneck posed by memory access in current inference systems. 

\textbf{Test-Time Scaling.} Recent LLMs such as DeepSeek-R1~\citep{guo2025deepseek}, OpenAI-o1/o3~\citep{jaech2024openai}, and QwQ~\citep{qwq32b} generate extended CoT reasoning~\citep{wei2022chain} to solve complex problems, including those from AIME~\citep{aime24,aime25}. Parallel search through repeated sampling~\citep{brown2024large}, majority voting (self-consistency)~\citep{wang2022self}, and reward-model~\citep{wu2024inference,feng2023alphazero,snell2024scaling} (e.g., \bon, weighted voting, tree search) aims to improve reasoning accuracy. Strategies such as~\citep{fu2024efficiently,arora2502training,sky} and hybrid models~\citep{lieber2024jambahybridtransformermambalanguage, paliotta2025thinkingslowfastscaling, wang2025m1scalabletesttimecompute} have been proposed to reduce the cost of test-time scaling. 

\begin{tcolorbox}[colback=white,colframe=softlavender,title=Note]
Advanced test-time strategies shift evaluation from token-centric metrics (e.g., perplexity, latency) to \textit{task-level throughput}—the number of tasks completed per unit time. This shift is especially relevant for reasoning tasks, where intermediate steps may vary widely depending on the strategy, yet the ultimate utility hinges almost entirely on the correctness of the final output. In contrast, traditional tasks like chat completions focus on token-level quality and throughput.
\end{tcolorbox}

\textbf{Sparse Attention.} A significant line of prior work has focused on overcoming the quadratic computational bottleneck of attention mechanisms during LLM training by leveraging the natural sparsity of attention matrices~\citep{child2019generating,kitaev2020reformer,daras2020smyrf,zaheer2020bigbird,beltagy2020longformer,yuan2025native}. More recently, sparse attention has experienced a resurgence in the context of LLM inference, where methods such as~\citep{zhang2023h2o,xiao2024efficientstreaminglanguagemodels,tang2024quest,liu2024retrievalattention,chen2024magicpig,hu2025efficient} restrict the memory access of the key-value (KV) cache during generation while maintaining strong performance. These advances form a strong and steady foundation for our exploration of a new test-time scaling paradigm.

Extended related work is discussed in~\Cref{relatedwork}.

\begin{table}[ht]
\centering \caption{Notation Used throughout the Paper.}\par
\begin{tabular}{@{}llllll@{}}
\toprule
\textbf{Symbol} & \textbf{Description} & \textbf{Symbol} & \textbf{Description} &
\textbf{Symbol} & \textbf{Description}\\
\midrule
$T,\mathcal{T}$ & Task (set) & $L_{out}$ & \# Gen tokens &$L_{in}$ & Prompt length\\
$M$ & Model & $N, N_T$ & Reasoning trials &$D$ & KV size / token \\
$C, C_{\text{TTS}}(\cdot)$ & Cost function & $n, n_T$ & Max \# tokens & $P$ & Parameters\\
$\mathcal{A}$ & Algorithm & $B, B_T$ & KV budget & $r$ & GQA ratio\\
\bottomrule
\label{tab:notation}
\end{tabular}
\end{table}

\subsection*{Cost Model}
\label{sec:cost-model}
We first calculate the inference cost for the cases where the batch size is 1, and then extend to a more general case in TTS. Finally,  we propose our cost model using equivalent FLOPs. We assume the model weight and KV cache are stored and calculated using the same precision.

\textbf{Computation.}  As discussed in~\citep{brown2024large}, the computation of a transformer architecture layer consists of two parts: linear modules and self-attention, which is given by:
\[
    C_{\text{comp}} =  
    \underbrace{2P L_{out}}_{\substack{\text{model parameters computation} \\ (\text{MLP and }W_{Q},W_{K},W_{V}, W_{O})}} + 
    \underbrace{r (2L_{in} + L_{out}) L_{out} D}_{\substack{\text{self-attention}\\ \text{Softmax}({qk^T}/{\sqrt{d}})v}}
\]

\vspace{-0.2cm}
\paragraph{Memory Access.} Memory access also consists of two parts, model parameters and KV cache:
\[
    C_{\text{mem}} =  
    \underbrace{2P L_{out}}_{\substack{\text{model parameter access} \\ (\text{MLP and }W_{Q},W_{K},W_{V}, W_{O})}} + 
    \underbrace{
        2L_{in} L_{out} D + L_{out}^2 D
    }_{\substack{
        \text{KV cache (prompt + decoding)} \\
        \text{Softmax}({qk^T}/{\sqrt{d}})v
    }}
\]

\begin{wrapfigure}{r}{0.35\linewidth}
\centering

\includegraphics[width=\linewidth]{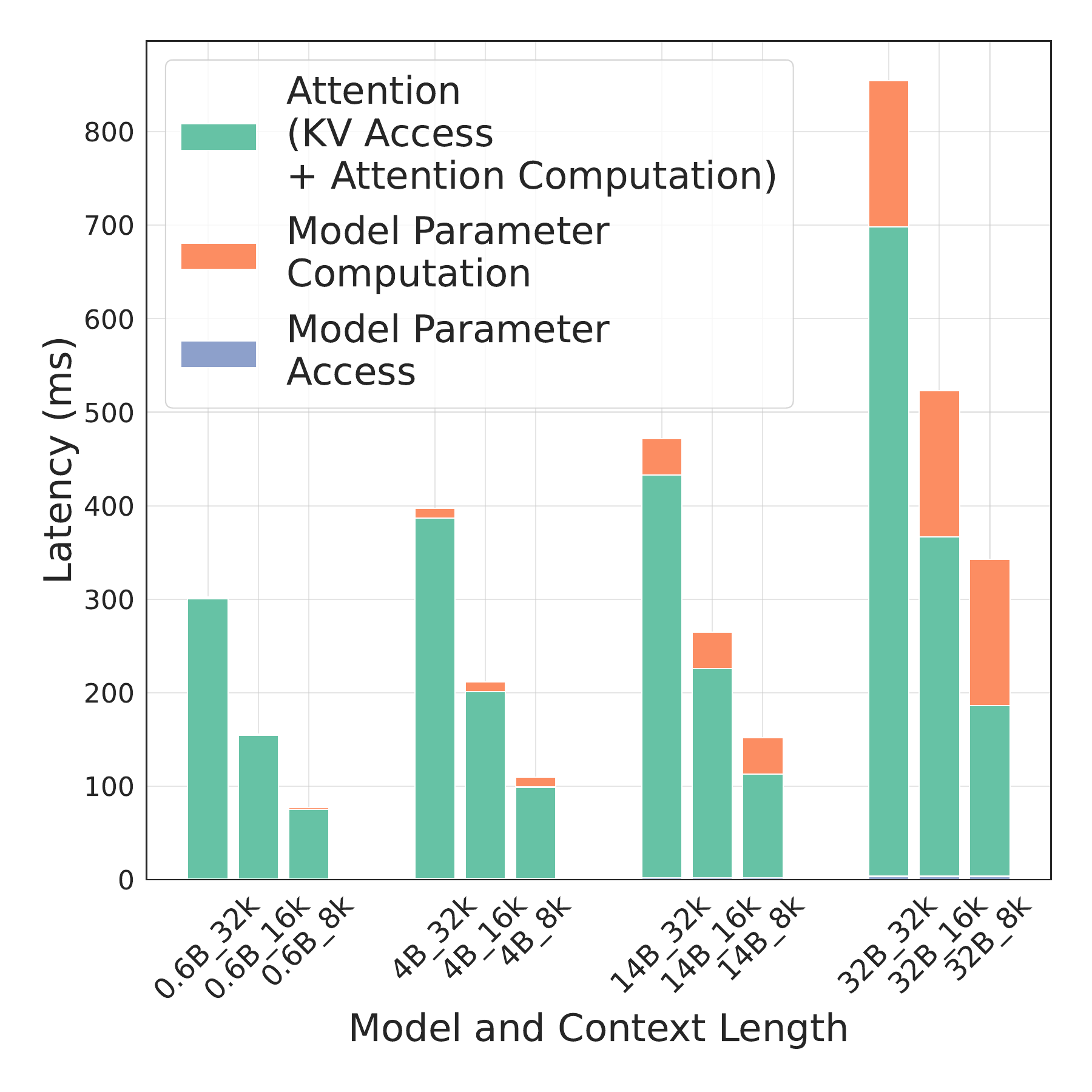}
\captionof{figure}{Latency breakdown for different model sizes (Qwen3 series) and context lengths (batch size 4K).}
\label{fig:sparse_attention_break_down} 
\end{wrapfigure} 

In real serving scenarios, a large batch size will be used~\citep{deepseekinfra} with growing GPU VRAM~\citep{tirumala2024nvidia} and model parallelism~\citep{pope2023efficiently}. The access to the model parameters will be amortized across requests in a batch; \Cref{fig:sparse_attention_break_down} shows parameter access time is negligible when the batch size is large. 
Thus, we only consider the second term (i.e., KV cache loading) in our cost function. 
Furthermore, in the cases that we have $N$ reasoning trials, the prompt cache access~\citep{juravsky2024hydragen,zheng2024sglang} is also shared across these $N$ trials. Thus, 
\begin{align}
 C_{\text{comp}}(N) &=   {2P NL_{out}} + 2rNL_{in}L_{out}D + rNL_{out}^2D \label{comp}\\
     C_{\text{mem}}(N) &=   {2L_{in} L_{out} D} + N {L_{out}^2 D} \label{mem}
\end{align}  
\textbf{eFLOPs.} We propose $\text{eFLOPs}$ (equivalent FLOPs) to capture both compute and memory access cost,
\begin{equation}
  \text{eFLOPs} = C_{\text{comp}} + C_{\text{mem}} \times I \footnote{Max cost model $ \max (C_{\text{comp}},C_{\text{mem}} \times I)$ also works here and favor our claims more since most of the time $C_{\text{mem}} \times I$ dominates the cost. We choose to use an additive cost model because $C_{\text{comp}}$ mainly comes from linear layers while $C_{\text{mem}}$ mainly comes from the self-attention layer. The parallelization of these components during decoding remains an active area of research~\citep{zhu2024nanoflow}. We discuss this max cost model in~\Cref{maxvsadditive}.}
    \label{eflops} 
\end{equation}
where $I$ is the arithmetic intensity of hardware, which reflects that modern accelerators usually have a much larger computation capacity over memory bandwidth, and the gap is growing over the years~\citep{sadhukhan2024magicdec}.

Combining~\Cref{eflops,mem,comp}, we obtain the final cost model:
\begin{equation}
\colorbox{lavenderpink}{$
C_{\mathrm{TTS}} = \underbrace{2NPL_{out}}_{\text{linear modules computation}} + \underbrace{2rNL_{in}DL_{out} + rNDL_{out}^2}_{\text{self-attention computation}} + \underbrace{2IL_{in}DL_{out} + INDL_{out}^2}_{\text{KV access}} 
$}
\label{eq: cost model full equations}
\end{equation}
where $P, r, D$ are hyper-parameters determined by the model $M$. In MoE models, $P$ stands for the number of \textit{active} parameters rather than total parameters.

\textbf{Analysis.}
\label{roofline}
Our key insight is \textbf{attention-related cost dominates in long CoTs.} We show this by estimating the ratio of attention-related cost to parameter-related cost $\Phi$:
\begin{align}
    \Phi = \frac{2rL_{in}D + (rD + ID)L_{out}}{2P} \nonumber
\end{align}
As shown in~\Cref{fig: Attention-Model-ratio}, in the regime of long CoTs, where the generation length exceeds $4096$ tokens, the cost of attention surpasses that of model parameters by a factor of 10-1000$\times$. 

While multi-head latent attention (\textbf{MLA}; \citealp{liu2024deepseek}) reduces KV memory access by a constant factor (similar to $r$ in GQA), it is insufficient for achieving true scalability due to several limitations: (1) MLA does not reduce attention computation; (2) the gap between FLOPs and memory bandwidth is expected to widen in the future; and (3) emerging \textbf{fine-grained MoEs}~\citep{llama4modelcard,dai2024deepseekmoe,snowflake_arctic_2024} drastically reduce FLOPs in linear layers by a factor of 10-20$\times$, further increasing the relative cost of attention.

Under the context of \longcot being widely adopted, we can safely assume generated length $L_{out} \gg L_{in}$ or \textit{at least} proportional to $L_{in}$. Hence, the bottleneck of inference is shifted from linear term $L_{out}P$ to the quadratic term $L_{out}^2D$, motivating our \law scaling law, akin to kinetic energy: $E_k = \frac{1}{2}mv^2$. 

\textbf{Experimental Setup.}
\textbf{Tasks:} we focus on three challenging reasoning benchmarks: AIME24~\citep{aime24}, AIME25~\citep{aime25}, math datasets spanning algebra, combinatorics, and geometry, and LiveCodeBench~\citep{jain2024livecodebench}\footnote{For LiveCodeBench, we sample 50 problems from the \textit{v5} subset (24 hard, 16 medium, 10 easy).}, which includes complex programming problems from recent coding competitions. \textbf{Models:}  We evaluate performance across various model sizes of the Qwen3~\citep{yang2025qwen3technicalreport} and DeepSeek-R1-Distilled-Qwen~\citep{qwen2.5,guo2025deepseek} series. \textbf{Test-time Strategies:} To eliminate the confounding effects introduced by the specific implementations of test-time strategies, such as the quality of reward models, we adopt two representative yet straightforward approaches: \longcot, a practical and widely used method in state-of-the-art reasoning models, and the \textit{oracle} \bon (repeated sampling~\citep{brown2024large}), which measures the solving rate for verifiable problems and suggests an upper bound via TTS. \textbf{Hardware:} We use the specifications of NVIDIA B200 as hardware reference to study the latest serving scenarios. Experiments details are presented in~\Cref{expdetails}.

\vspace{-0.1cm}
\section{Rethinking Test-time Scaling Laws}
\vspace{-0.1cm}
\label{sec:analysis}
\label{sec: model arch and TTS}
In~\Cref{sec:kinetic_scaling_law}, we first introduce \law, derived from empirical investigations across the Qwen3 model series. Then, we explore the underlying reasons for the divergence between \law and prior scaling laws through an Iso-Cost analysis in~\Cref{sec:iso}.


\subsection{\Law} 
\label{sec:kinetic_scaling_law} 
In this section, We study the scaling behavior of the Qwen3~\citep{qwen2,qwen2.5} considering the following problem:
\begin{center}
    \itshape
    For each fixed maximum inference budget, eFLOPs per question, what is the Pareto frontier of achievable accuracy across different LLM configurations?
\end{center}

With the refined cost model in~\Cref{sec:cost-model}, we first formulate the objective of the test-time scaling law, focusing on the tradeoff between model size and the number of generated tokens.

\textbf{\law (for dense models).} Given a problem instance \( T \) and a total inference budget \( C \), our goal is to explore the optimal tradeoff between two key factors: the choice of language model \( M \), and the number of reasoning trials \( N \) or the maximum generation length \(n\). More precisely,
\begin{equation}
    \colorbox{lavenderpink}{$
    (N, n)_{*}, M_{*} = \arg\max_{(N, n), M} \; \mathrm{Acc}(N, n, M; T) \quad \text{s.t.} \quad C_{\mathrm{TTS}}(N, n, M; T) \le C
    $}
    \label{dense scaling law}
\end{equation}
Let \( \mathrm{Acc}(N, n, M; T) \) denote the problem-solving rate of model \( M \) on task \( T \), using \( N \) reasoning trials, each with a maximum reasoning length of \( n \).\footnote{For fairness, we do not schedule resources across tasks, but consider a resource upper bound for all the tasks. } 

\begin{figure*} 
    \includegraphics[width=\linewidth]{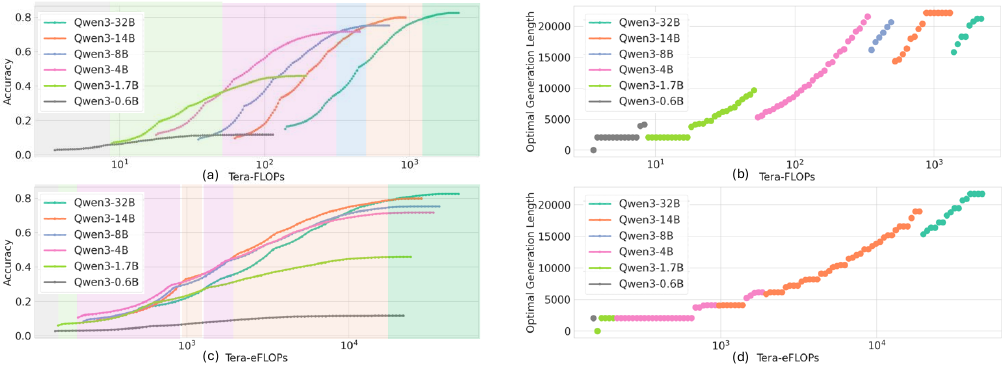} 
    \caption{\textbf{AIME24 Pareto Frontier (\longcot).} Evaluations of Qwen3 series models. By controlling the maximum allowed generation lengths, we control the incurred inference cost in eFLOPs (\textbf{ab} for our scaling law) or FLOPs (\textbf{cd} for previous scaling law) and measure the accuracy (Pass@1) in AIME24. The optimal model is marked with different colors in \textbf{(ac)}. The optimal generation length is presented in \textbf{(bd)}.}
    \label{fig:modelcost} 
\end{figure*} 

In the \textit{Long CoTs} scenario, where $N_T = 1$, we vary $n_T$ to evaluate the model performance under different costs. We present our results in~\Cref{fig:modelcost}. \law highlights two important findings compared to the previous scaling law, which focused on merely FLOPs: 

\begin{itemize}[itemsep=0.5pt, topsep=2pt, leftmargin=*]
    \item \textbf{Efficiency of small models is overestimated.} As shown in~\Cref{fig:sparse_attention_break_down,fig:modelcost} \textbf{(ac)}, smaller models, despite having fewer parameters, are not as efficient as commonly assumed. For example, the 14B model outperforms both the 4B and 8B models even at low accuracy levels (e.g., below $40\%$), and the 0.6B model only lies on the Pareto frontier in regions where accuracy is negligible. In contrast, under previous scaling laws, models of all sizes span a meaningful portion of the Pareto frontier. 
    \item \textbf{Extending CoTs is more effective than enlarging parameters only for models beyond a critical scale (empirically, 14B).} \law reveals that under constrained compute budgets, allocating resources to model scaling yields greater returns than increasing CoT length. As illustrated in~\Cref{fig:modelcost} \textbf{(bd)}, only the 14B and 32B models benefit from generating CoTs longer than 10K tokens; for smaller models (e.g., 1.7B and 4B), switching to a larger model is more advantageous when $L_{\text{out}} < 5\text{K}$. This suggests that, in practice, most of the available compute should be devoted to increasing model size rather than lengthening generations (\Cref{fig:modelcost} \textbf{(d)}). In contrast, previous scaling laws assumed that longer CoTs provided consistent benefits across all model sizes, recommending model scaling only after CoT performance gains had plateaued.    
    
\end{itemize}

\begin{figure*}
    \centering
    \subfloat[Accuracy (eFLOPs)]{
\includegraphics[width=0.32\linewidth]{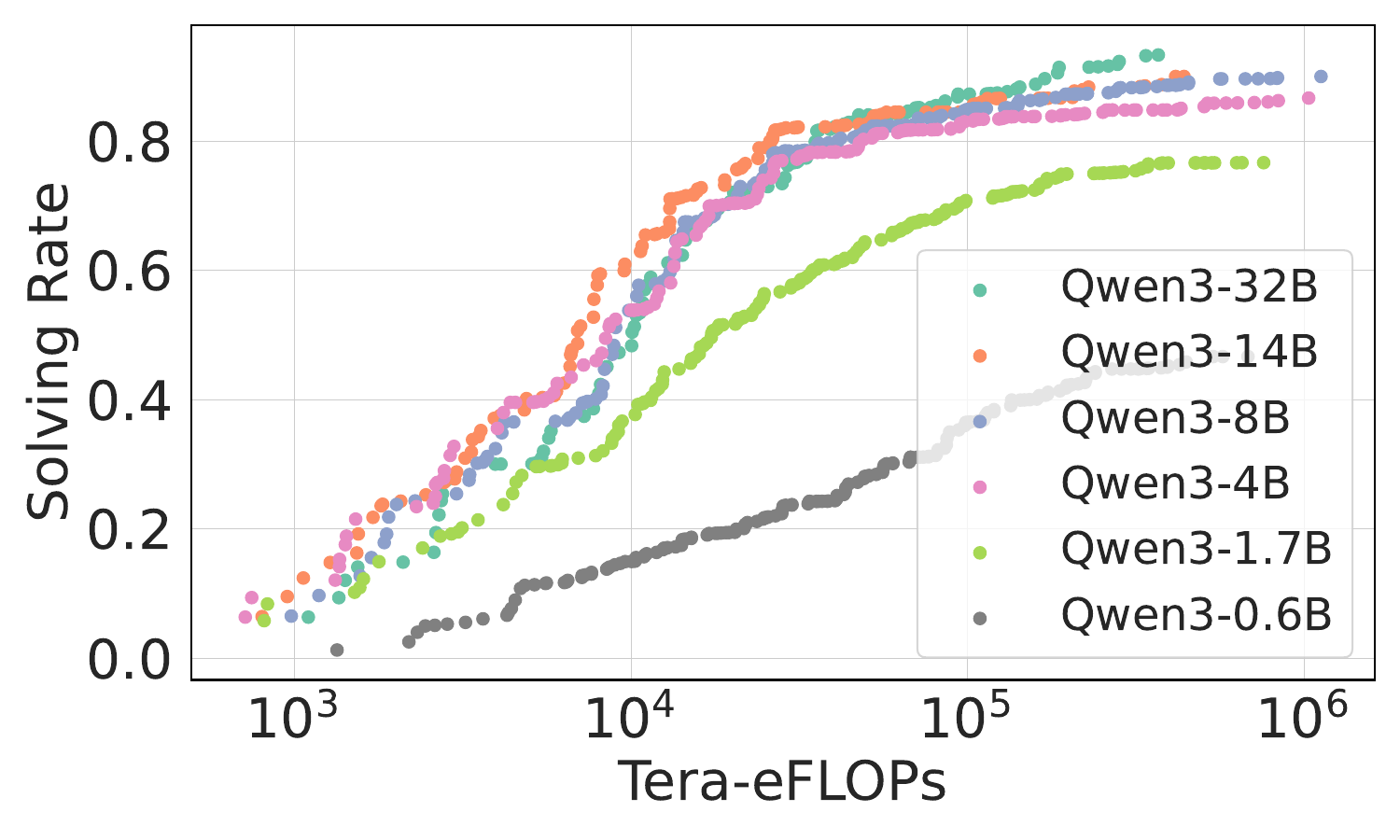}
    \label{fig: bon-acc-eflops}
    }
  \subfloat[Accuracy (FLOPs)]{
\includegraphics[width=0.32\linewidth]{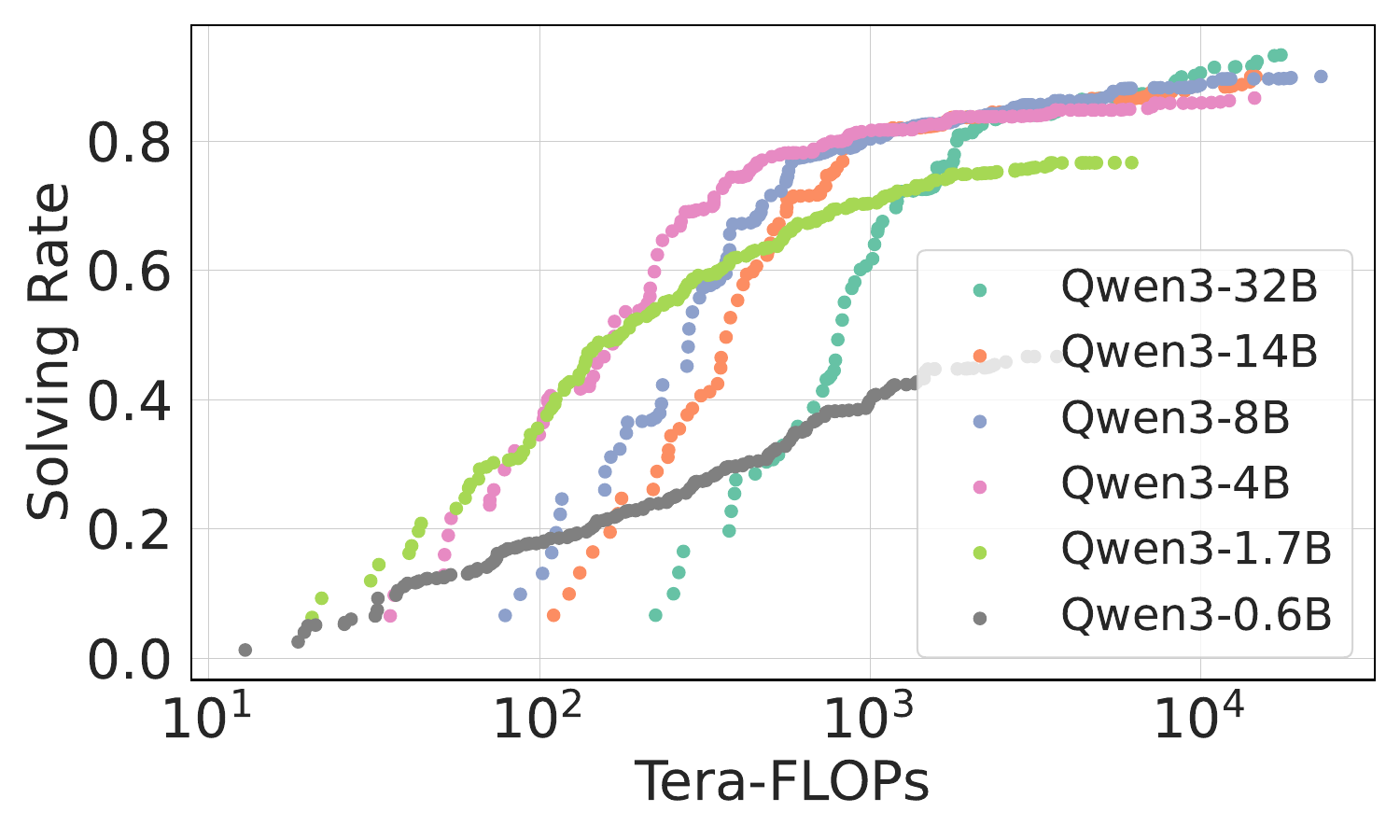}
    \label{fig: bon-acc-flops} 
    }
    \subfloat[Optimal Models]{
\includegraphics[width=0.32\linewidth]{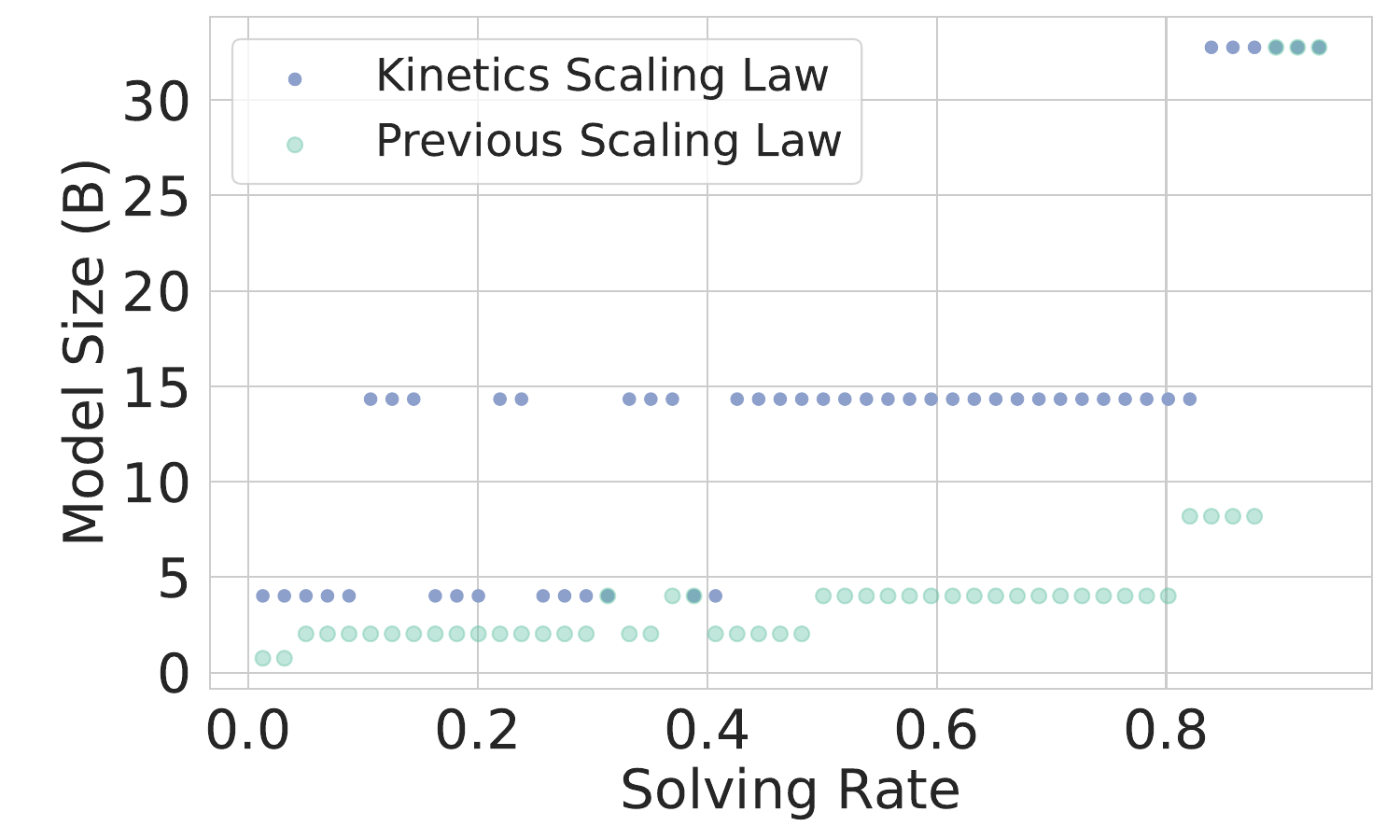}
    \label{fig: bon-model-selection} 
    } 
    \caption{\textbf{AIME24 Pareto Frontier (\bon).} We control the incurred inference cost in eFLOPs (\textbf{a}) or FLOPs (\textbf{b}) and measure the solving rate (coverage) in AIME24 for various models by varying the maximum allowed number of reasoning trials. We use the curve envelopes to project the optimal models in (\textbf{c}). 
    }
\end{figure*}

In the \bon setting, we fix the maximum number of generated tokens at $n_T$, and vary the number of reasoning trials $N$ to evaluate the problem-solving rate (i.e., the probability that at least one trial produces a correct answer). We have similar observations in~\Cref{fig: bon-acc-eflops,fig: bon-acc-flops,fig: bon-model-selection}. Under the previous scaling laws (\Cref{fig: bon-acc-flops}), the most cost-effective strategy to achieve high accuracy is to apply repeated sampling using smaller models. 
\law (\Cref{fig: bon-acc-eflops}) reveals that deploying a 14B model with fewer reasoning trials is more efficient. We also observe a critical size of 14B. For models smaller than 14B, increasing compute is best allocated toward model scaling rather than additional trials. For models at or above 14B, however, further computation is more effectively spent on increasing the number of reasoning trials, up to diminishing returns. 

The above observations are consistent in DeepSeek-R1-Distilled-Qwen series, while the critical model size becomes 7B. Experiments on AIME25 and LiveCodeBench as well as  the analysis of DeepSeek-R1-Distilled-Qwen are presented in~\Cref{densescalinglaw}.
\subsection{Iso-Cost Study}
\label{sec:iso}
We attribute the above divergence between Kinetics and previous scaling laws to two reasons.
\begin{figure*}
    \centering
    \subfloat[KV v.s. Parameters]{
\includegraphics[width=0.31\linewidth]{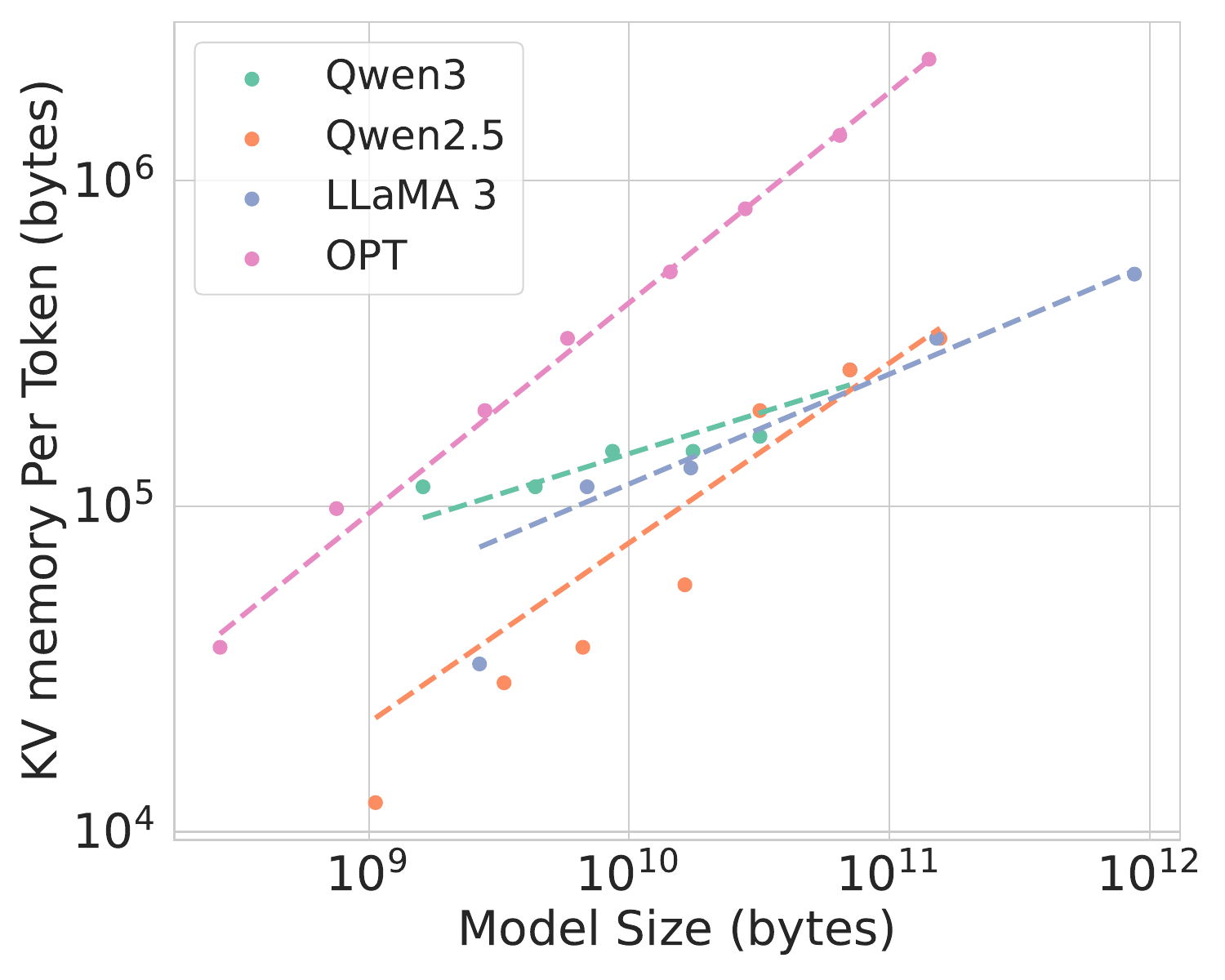}
    \label{fig: kvtrend}
    }
  \subfloat[Iso-eFLOPs]{
\includegraphics[width=0.32\linewidth]{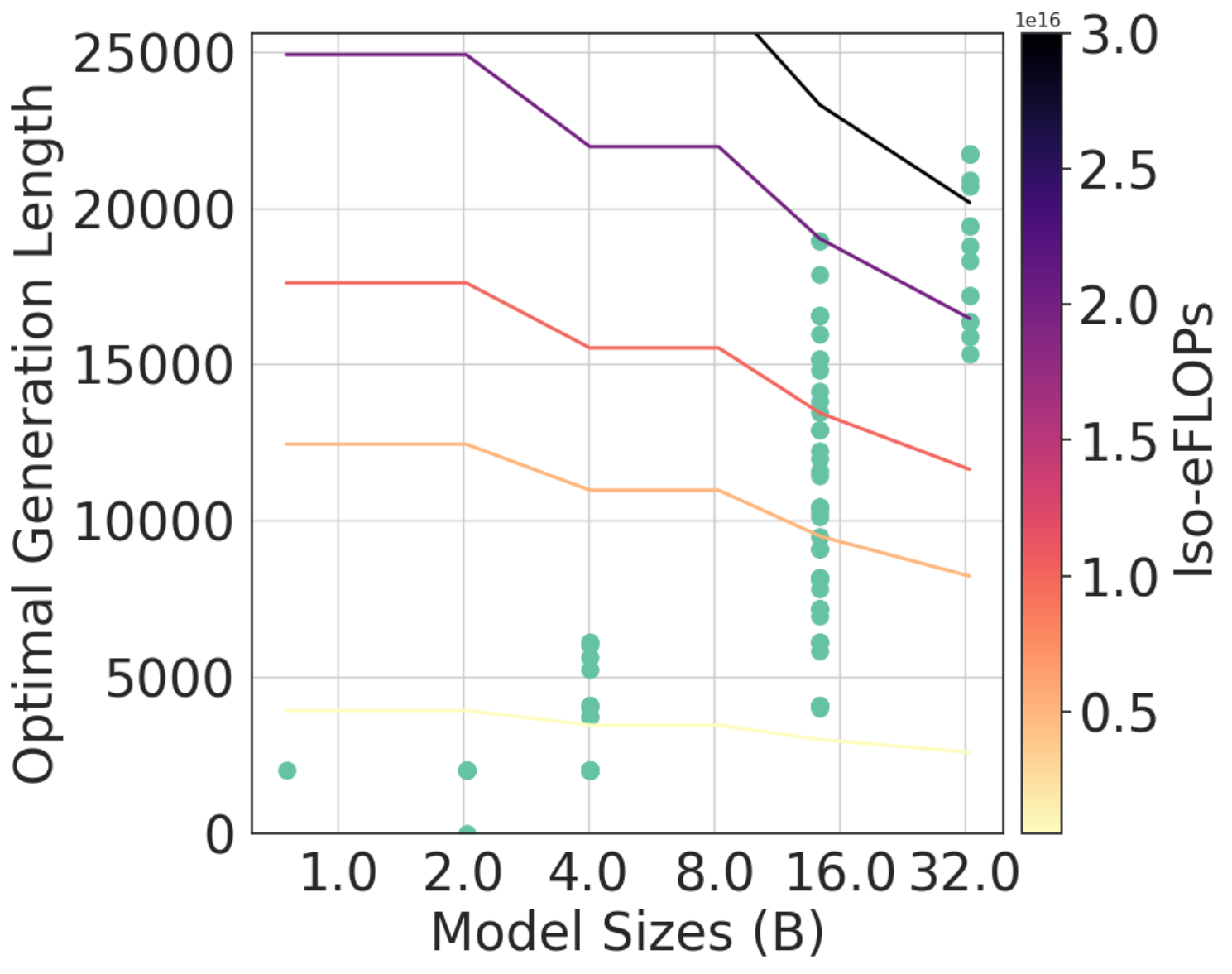} 
    \label{fig: iso-eflops} 
    }
    \subfloat[Iso-FLOPs]{
\includegraphics[width=0.32\linewidth]{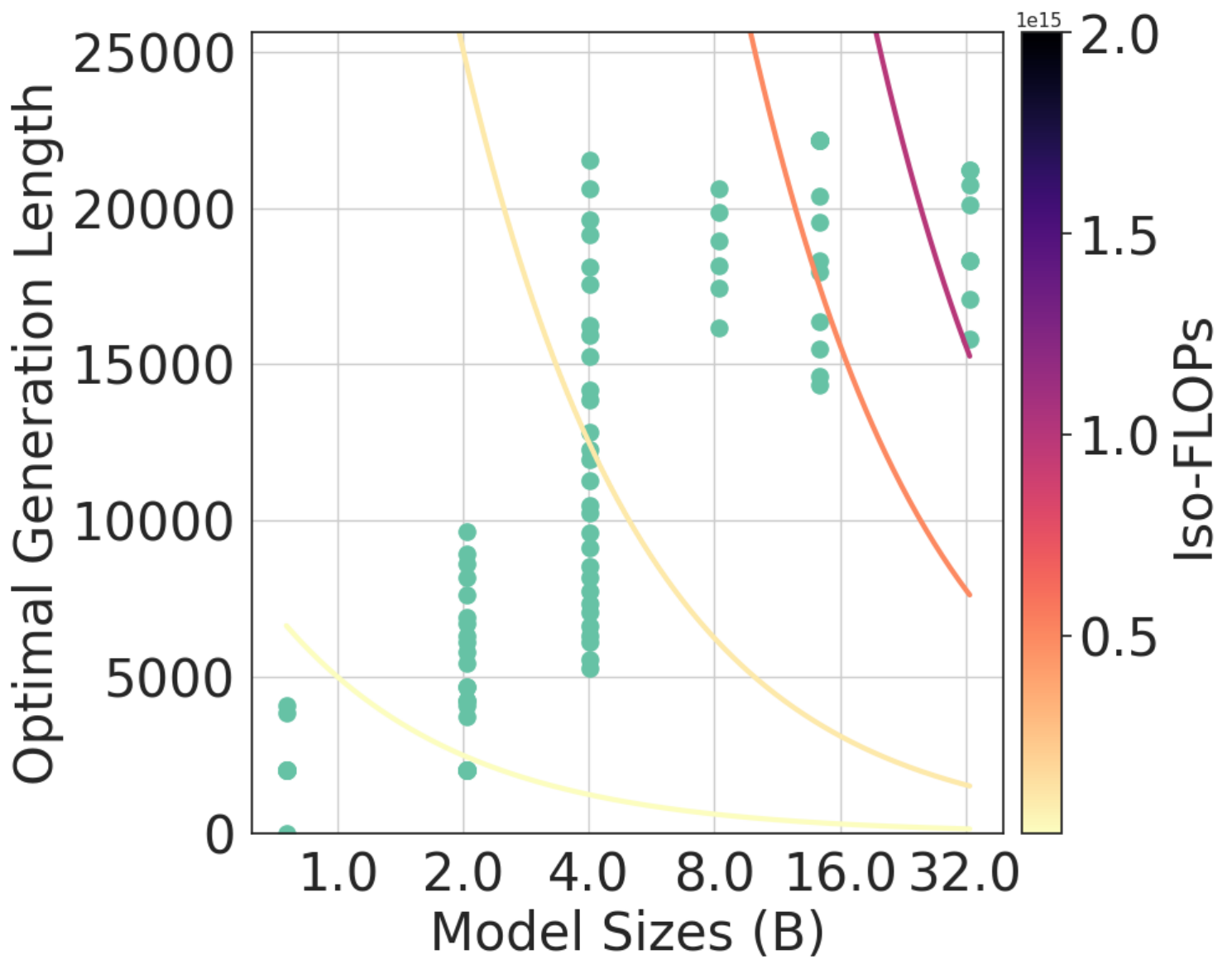} 
    \label{fig: iso-flops} 
    } 
    \caption{\textbf{Explanation of the New Scaling Law.} \textbf{Left:} Analysis across four LLM families reveals a consistent trend of disproportionately slower KV memory growth relative to model size. For the Qwen3 series in particular, doubling model parameters results in only a $1.18\times$ increase in KV cache size. \textbf{Middle and Right:} We compare the Iso-Cost landscapes under the proposed cost model (\textbf{b}) and the traditional model (\textbf{c}).
    }
\end{figure*} 
\begin{itemize}[itemsep=0.5pt, topsep=2pt, leftmargin=*]
\item \textbf{Disproportionation between KV memory size $D$ and model parameters $P$.} Smaller models tend to require significantly more KV cache relative to their parameter size. For example, Qwen3-0.6B demands 3.5GB of KV cache to store 32K tokens, despite the model itself occupying only 1.2GB. In contrast, Qwen3-32B uses just 8GB of KV cache for the same sequence length. Empirically, doubling model parameters results in only a $1.18\times$ increase in KV cache size. As shown in~\Cref{fig: kvtrend}, this phenomenon is consistently observed across model families such as OPT~\citep{zhang2022opt} ($1.55\times$), Qwen2.5~\citep{qwen2.5} ($1.46\times$), and LLaMA3~\citep{grattafiori2024llama} ($1.27\times$).

\item  \textbf{Shift from linear to quadratic cost model.}  Under this revised model, increasing generation length incurs a substantially higher cost than scaling model size; consequently, the tradeoff between model capacity and token budget shifts meaningfully. For instance, under the linear $LP$ model, the cost of generating 8K tokens with a 14B model (which is usually insufficient to solve complex tasks) is treated as equivalent to generating 24K tokens with a 4B model (sufficient to complete most tasks). However, under the $L^2D$ model, the same 14B@8K generation is only comparable in cost to a 4B@9K generation. This tighter bound makes it much harder for smaller models to compensate for their limited capacity through extended generation alone. Thus, only if the gap in model capacities is small enough (e.g., 32B only improves the accuracy by $3\%$ on AIME24 compared to 14B), the benefits of extending generation length might be more effective than directly enlarging model parameters.
\end{itemize}
\Cref{fig: iso-eflops,fig: iso-flops} show an Iso-Cost analysis comparing two cost models. Under \law, the cost grows quadratically with $L_{out}$, while the KV cache scales sub-linearly with model parameters $P$. As a result, when total budget is low, the Iso-eFLOPs contours tend to stretch horizontally, favoring larger model sizes over longer generation lengths. This implies that increasing model size is a more efficient use of resources than generating longer outputs. In contrast, the traditional FLOPs-based model leads to steeply vertical contours, encouraging longer generation before increasing model size.

\section{Test-time Scaling with Sparse Attention}
\label{sec:s3}
Based on our findings in~\Cref{sec: model arch and TTS}, we propose a new scaling paradigm centered on sparse attention. We begin by presenting a simple approach for identifying oracle resource allocation in sparse attention models, which we use to plot the Pareto frontier in~\Cref{sec:sparse scaling laws}. We then analyze the resulting changes in the scaling law and show that sparse attention models with massive tokens generated at test time, no matter sequentially via \longcot or in parallel via \bon, can lead to significantly higher problem-solving rates in~\Cref{sec: sparse law}.

\subsection{Oracle Resource Allocation with Sparse Attention Models}
\label{sec:sparse scaling laws}

\textbf{Problem statement.} Let \( \mathcal{A} \) denote the corresponding sparsity algorithms (e.g., top-\(k\), block top-\(k\) and local. Our goal is to explore the optimal tradeoff among three factors: model $M$, KV budget $B$, and number of trials, and the maximum generation length $(N,n)$. Specifically, 
\begin{equation}
    \colorbox{lavenderpink}{$
    (N, n)_{*}, M_{*}, B_{*} = \arg\max_{(N, n), M, B} \; \mathrm{Acc}(N, n, B, \mathcal{A}, M; T)  \quad \quad
    \text{s.t.} \quad C_{\mathrm{TTS}}(N, n, B, \mathcal{A}, M; T) \le C
    \label{sparse scaling law}
    $}
\end{equation}
The cost function \( C_{\text{TTS}} \) differs from the one in~\Cref{eq: cost model full equations} as it incorporates sparse attention mechanisms (which reduces the quadratic term $L^2D$ back to a linear term $LBD$). This modified cost model is discussed in detail in~\Cref{sparseattentioncost}.

\textbf{Oracle resource allocation:} We present a method to obtain the optimal schedule between generation parameters \( (N, n) \) and the KV budget \( B \) for each task, establishing an upper bound on achievable performance and enabling analysis of the core tradeoff between TTS strategies and sparsity. We begin by solving the subproblem for each individual task \( T \):
\begin{equation}
    \max \quad \mathrm{Acc}(N_T, n_T, B_T, \mathcal{A}, M; T) \quad \text{s.t.} \quad C_{\text{TTS}}(N_T, n_T, B_T, \mathcal{A}, M; T) \le C
    \label{eq:sparse oracle for single question}
\end{equation}
Empirically, we discretize the search space. For instance, in \bon, we discretize the space of $N$ and $B$ by producing a search grid:
\[
G = \{N_{0}, N_{1}, \ldots, N_{i}\} \otimes \{B_{0}, B_{1}, \ldots, B_{j}\}
\]
For each pair \( (N_{a}, B_{b}) \in G \), we compute the corresponding cost \( C_{T,(a,b)} \) and accuracy \( \mathrm{Acc}_{T,(a,b)} \). We use \( (N_{T}, B_{T}) \in G\) which maximizes the accuracy under the cost constraint $C$ as an approximation for~\Cref{eq:sparse oracle for single question}. By combining the optimal configurations \( (N_T, B_T) \) for all tasks \( T \), we obtain a solution to the overall problem in~\Cref{sparse scaling law}. Similar discretizations also applies for \longcot. Thus we find the oracle resource allocation. We present how we obtain the oracle resource allocation in detail in~\Cref{optimalresourceallocation}.

\begin{figure*}
    \centering
    \subfloat[\bon Gen.]{
\includegraphics[width=0.24\linewidth]{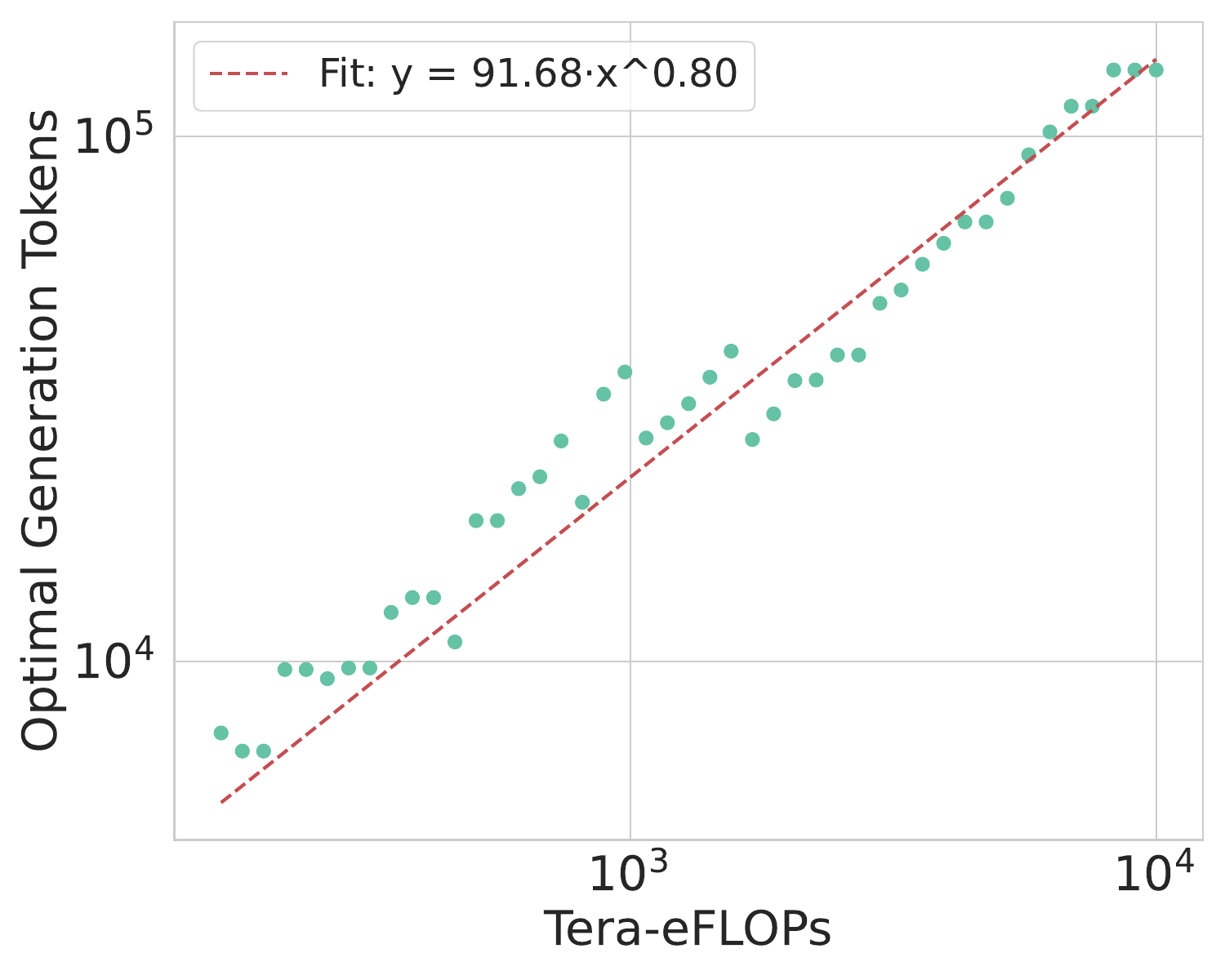}
    \label{fig: bon-8b-kv}
    }
  \subfloat[\bon Budget]{
\includegraphics[width=0.24\linewidth]{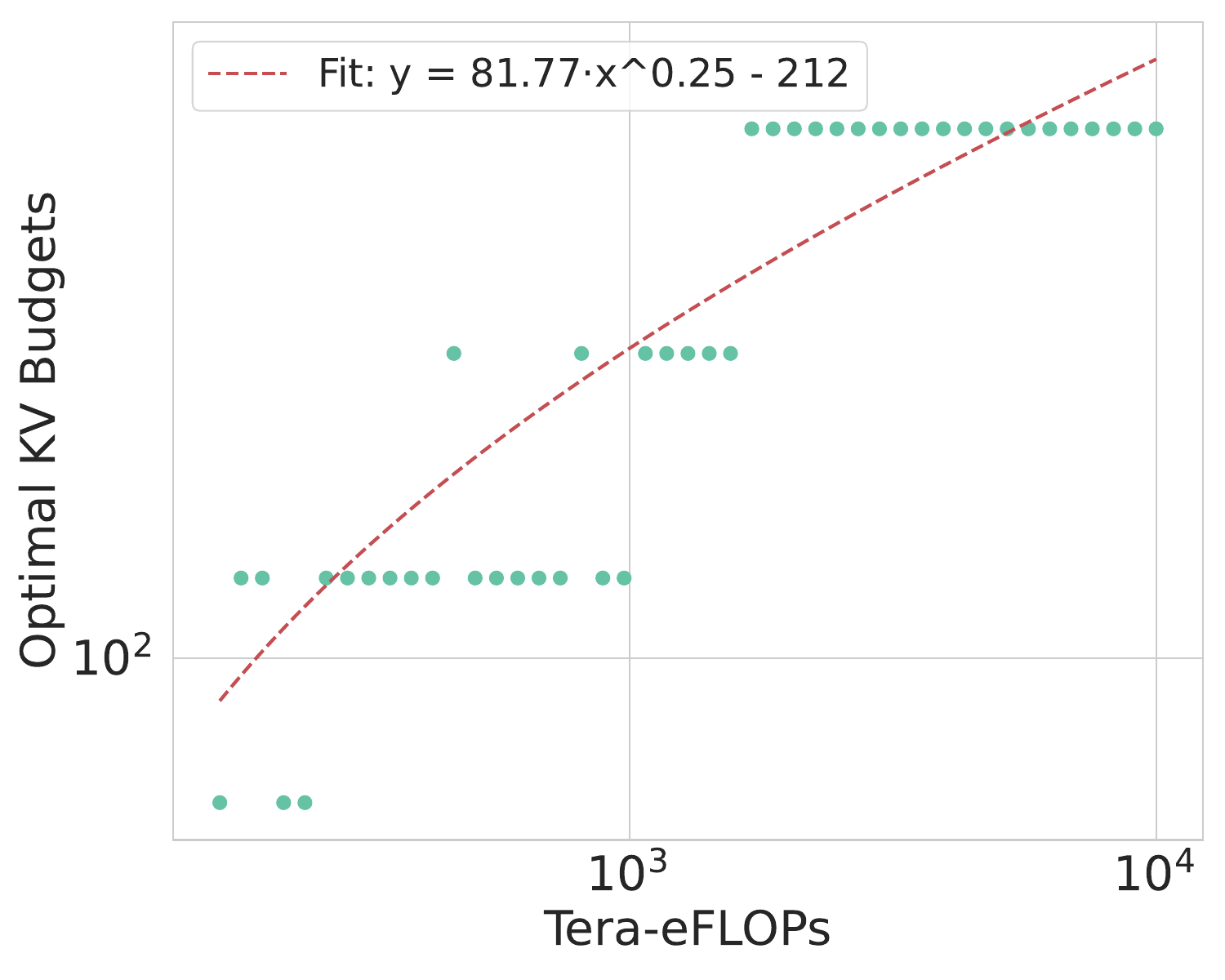}
    \label{fig: bon-8b-tokens} 
    }
    \subfloat[Optimal Model Selection Comparison]{
\includegraphics[width=0.44\linewidth]{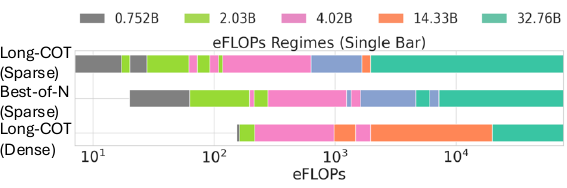}
    \label{fig:best-of-n--sparse-select} 
    } 
    \caption{\textbf{(ab) Tradeoff Between Generated Tokens and KV Budget.}  
We empirically investigate how to balance the tradeoff between generating more tokens and allocating a larger KV cache budget, which may yield more accurate but potentially shorter outputs. Using Qwen3-8B as a representative model, we fit curves to characterize this tradeoff.   
For \bon, we find that for every doubling of the total compute cost, the optimal KV budget increases by a factor of $1.18\times$, while the total number of generated tokens increases by $1.74\times$.  When the KV budget is small, the computational cost is dominated by model parameter-related computation rather than KV cache access. 
We incorporate a model-specific constant (212) into the fitted curve to account for this effect. \textbf{(c) Optimal Model Selection with Sparse Attention.} Compared to the scaling law for the dense models, small models (0.6B, 1.7B, 4B) are more effective with sparse attention. In other words, they occupy more space in the Pareto Frontier (\Cref{fig: bon-total}).
}   
\end{figure*}
\subsection{\sparselaw}
\label{sec: sparse law}
Sparse attention fundamentally reshapes \law and significantly enhances the scalability of TTS. We show \sparselaw with an oracle algorithm $\mathcal{A}=$top-$k$ to demonstrate the extraordinary potential of sparse attention. We present three important findings below. 

\begin{figure*}
    \centering
    \subfloat[\bon Scaling]{
\includegraphics[width=0.32\linewidth]{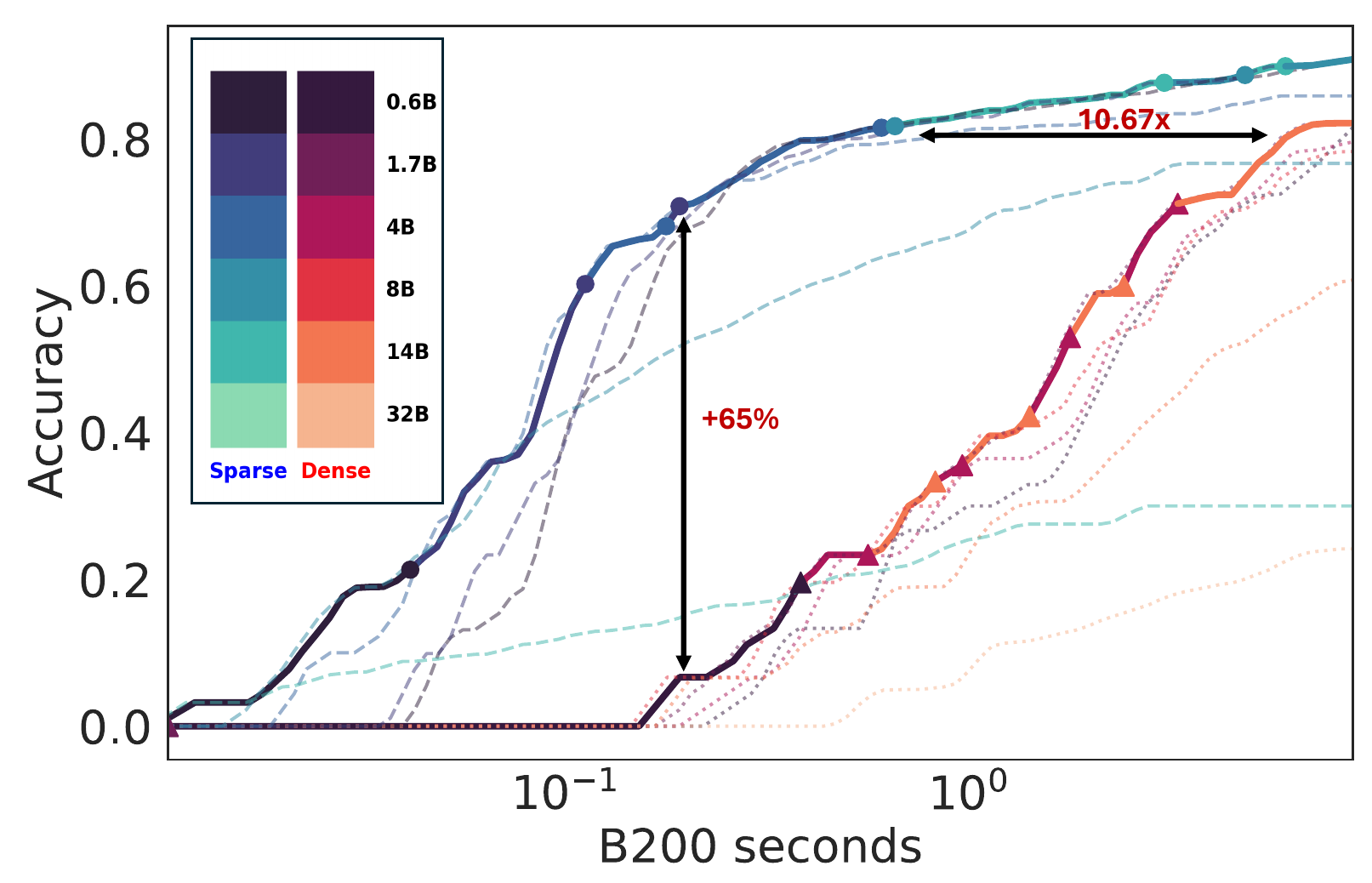}
    \label{fig: bon-total}
    }
    \subfloat[\bon 32B]{
\includegraphics[width=0.32\linewidth]{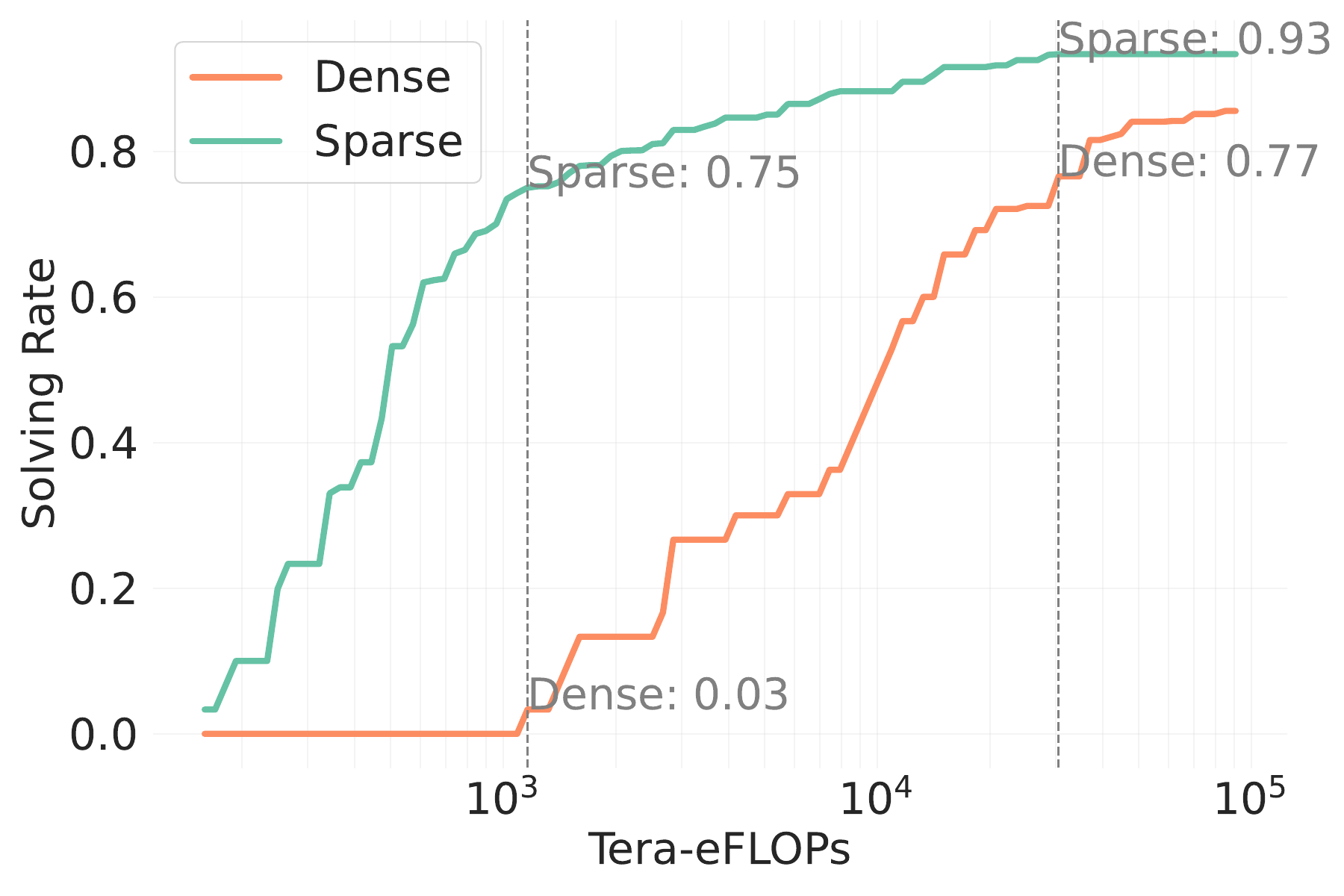}
    \label{fig: bon-32b} 
    }  
     \subfloat[\bon 30B-A3B]{
\includegraphics[width=0.32\linewidth]{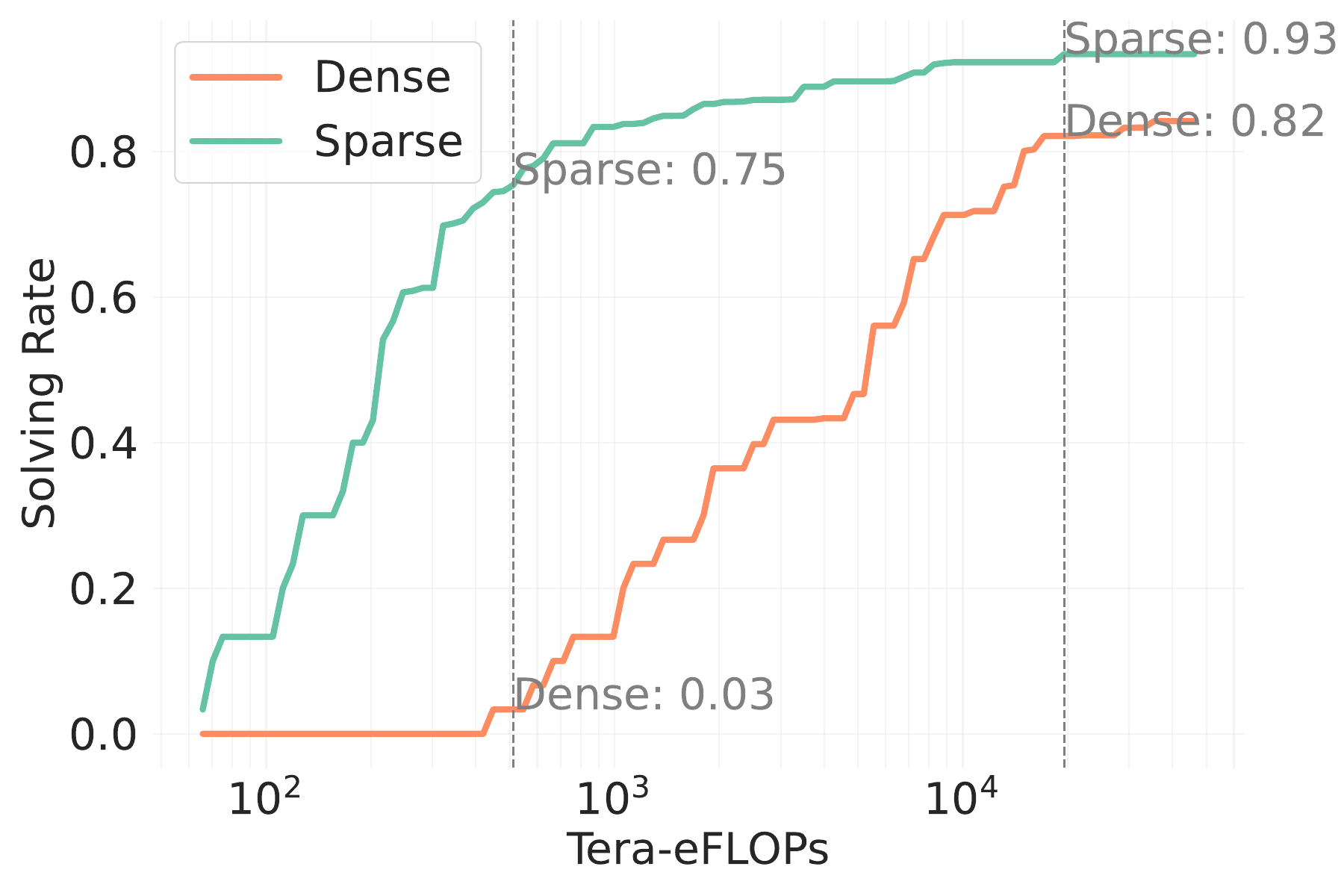}
    \label{fig: bon-moe} 
    }
    \\
     \subfloat[\longcot Scaling]{
\includegraphics[width=0.32\linewidth]{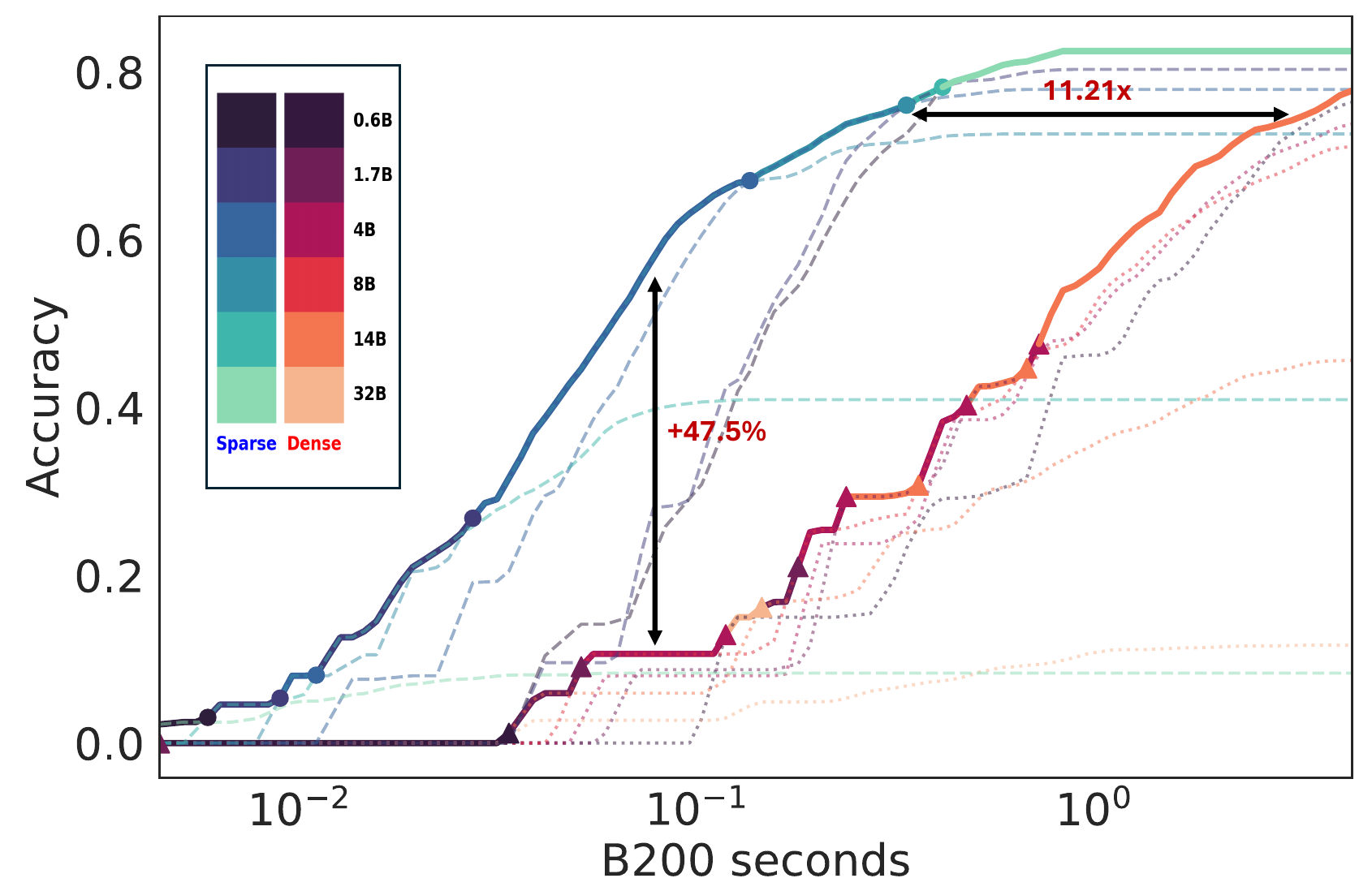}
    \label{fig: cot-total}
    }
    \subfloat[\longcot 32B]{
\includegraphics[width=0.32\linewidth]{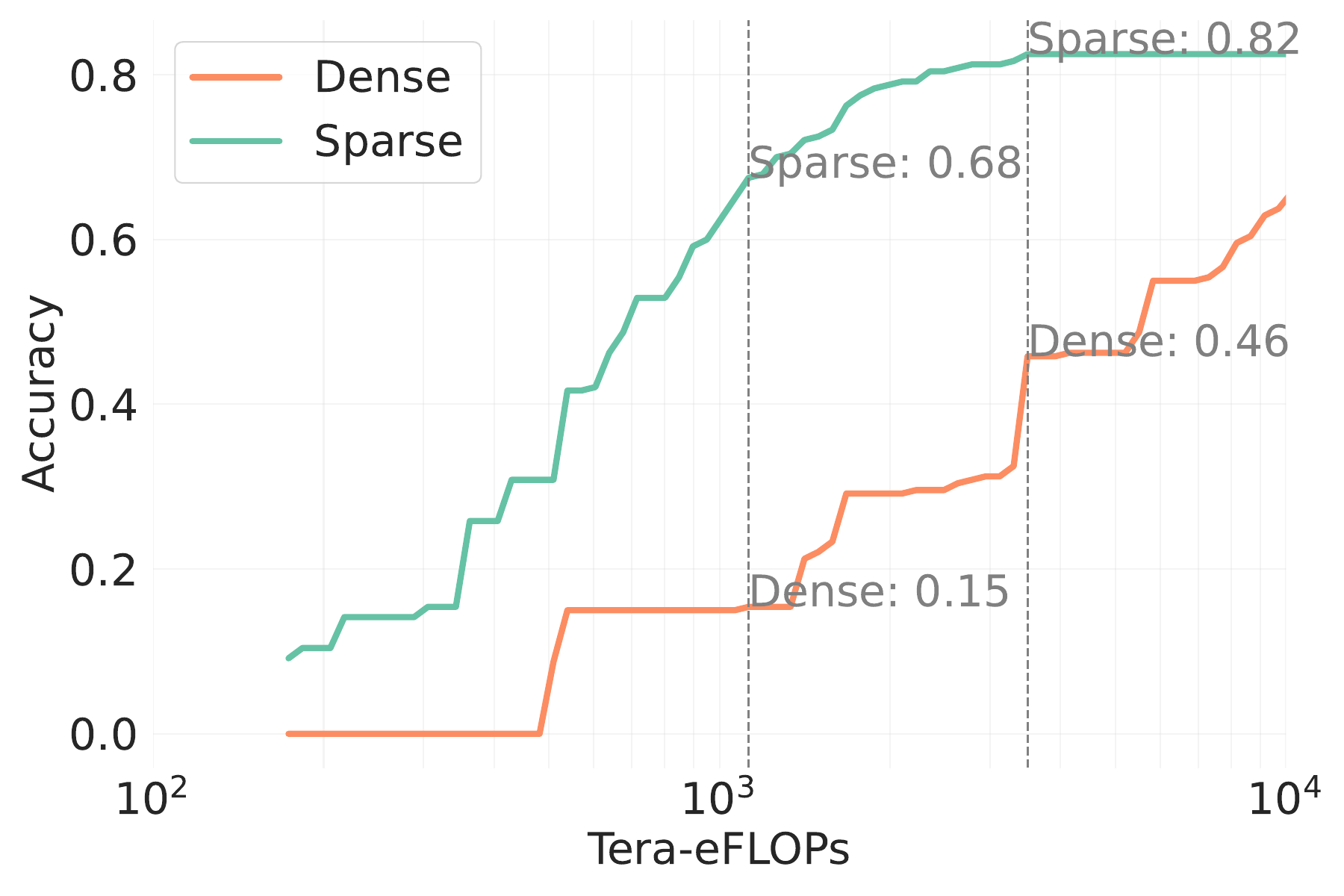}
    \label{fig: cot-32b} 
    } 
     \subfloat[\longcot 30B-A3B]{
\includegraphics[width=0.32\linewidth]{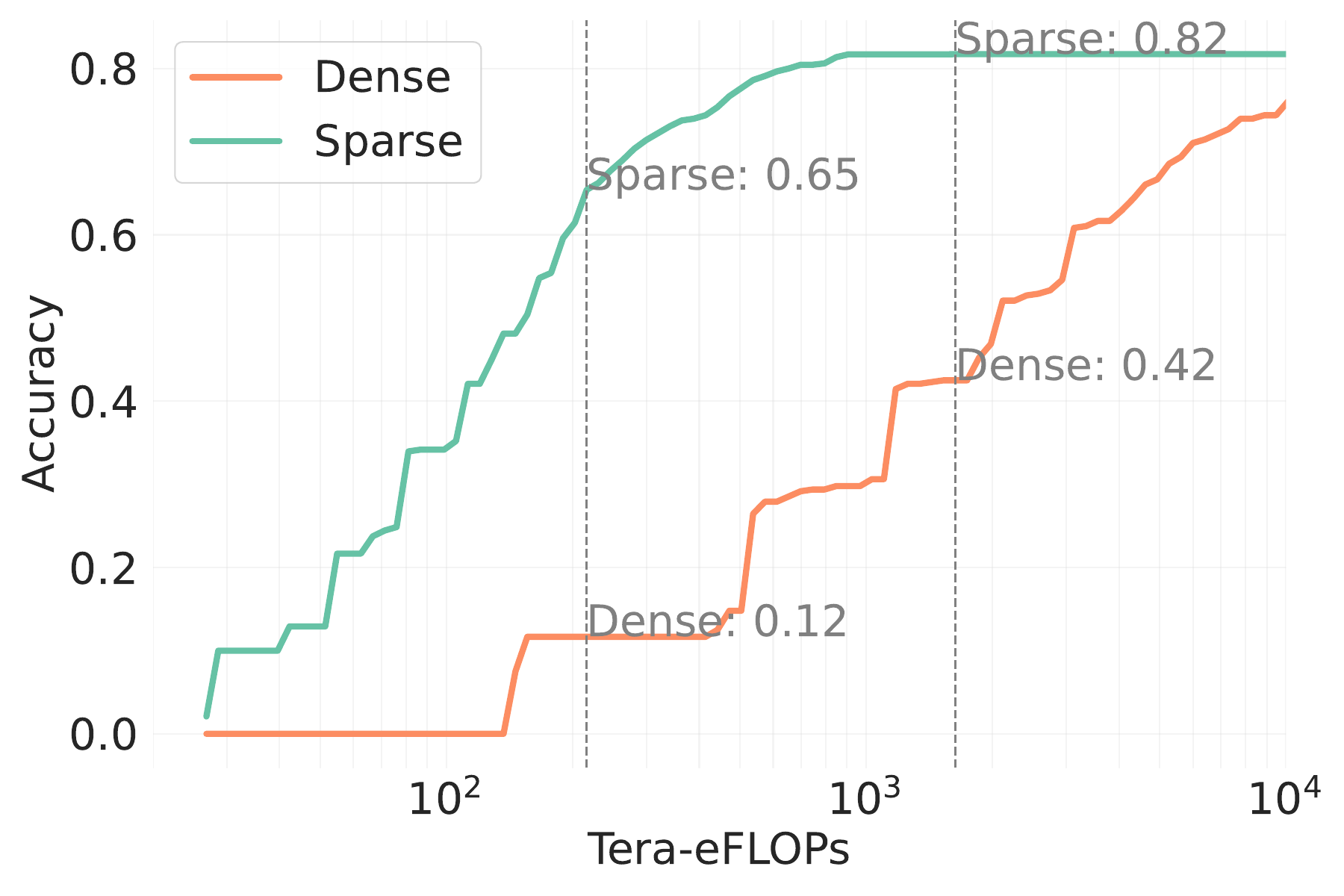}
    \label{fig: cot-moe} 
}\caption{\textbf{Sparse Attention Boosts Test-Time Scaling (AIME24).} In \textbf{(a)(d)}, we show that sparse attention models significantly improve the cost-accuracy trade-off under both inference strategies, ultimately achieving higher problem-solving rates at lower computational budgets. In \textbf{(b)(e)}, we analyze individual model performance (32B) and observe that sparse attention provides notable gains. In low-cost regimes, it can enhance problem-solving rates by \textbf{50--60 percentage points}. Even in high-cost regimes, sparse models maintain an advantage of around \textbf{5 points}, while reaching these performance levels much earlier. In \textbf{(c)(f)}, we show consistent conclusions for MoE models (Qwen3-30B-A3B). We use an oracle algorithm $\mathcal{A}=$top-$k$ here to present an upper bound of sparse attention.
    }
    \vspace{-0.3cm}
\end{figure*} 
\begin{itemize}[itemsep=0.5pt, topsep=2pt, leftmargin=*]
\item \textbf{Sparse attention significantly enhances problem-solving performance.} As shown in~\Cref{fig: bon-total,fig: bon-32b,fig: cot-total,fig: cot-32b}, compared to dense baselines, for both of the inference strategies and models of various sizes, sparse attention models improve problem-solving rates by up to 60 points in the low-cost regime and over 5 points in the high-cost regime. From an efficiency perspective, dense models require over $10\times$ more eFLOPs to match the same solving rate. These findings underscore sparse attention as a key enabler for unlocking the full benefits of test-time scaling. 

\item \textbf{Sparse attention becomes increasingly valuable in high-cost scenarios.} We investigate the tradeoff between KV budget $B$ and generation tokens. For \bon, we analyze how the optimal KV budget and the number of generated tokens scale with cost across $N$ reasoning trials. As shown in~\Cref{fig: bon-8b-kv,fig: bon-8b-tokens}, Our analysis reveals a consistent trend: allocating additional compute toward generating more tokens is generally more effective than expanding the KV cache. In \bon frontier, doubling the cost leads to only a $1.18\times$ increase in KV budget, compared to a $1.74\times$ increase in total generated tokens.

\item \textbf{Sparse attention reshapes \law.} As shown in \Cref{fig:best-of-n--sparse-select}, applying sparse attention significantly improves the efficiency of smaller models (0.6B, 1.7B, 4B), allowing them to re-emerge on the Pareto frontier across a broader range. Sparse attention reduces attention cost from a quadratic cost term ($L^2D$) to a linear one ($LBD$), making it negligible or comparable when compared to the cost of computing with model parameters ($LP$). 
\end{itemize}
Further results for AIME25 and LiveCodeBench are presented in~\Cref{sparsescalinglaw}, where we also ablate the performance of sparse attention on tasks with different difficulties. 

\textbf{Discussion: MoE models.} The emerging MoEs reduce the computation cost by a factor of $10\times$ to $20\times$~\citep{dai2024deepseekmoe,yang2025qwen3technicalreport,llama4modelcard}, further exacerbating the bottleneck in attention. We present the advantages of sparse scaling on Qwen3-30B-A3B in~\Cref{fig: bon-moe,fig: cot-moe}.

\section{Experimental Validation}

\label{sec:experiments}

\begin{figure*}
    \centering
     \subfloat[\bon Scaling (block top-$k$)]{
\includegraphics[width=0.32\linewidth]{neurips25/figures/AIME24-blocktopk-trial.pdf}
    \label{fig: bon-bs-frontier} 
    }
     \subfloat[\longcot Scaling (block top-$k$)]{
\includegraphics[width=0.315\linewidth]{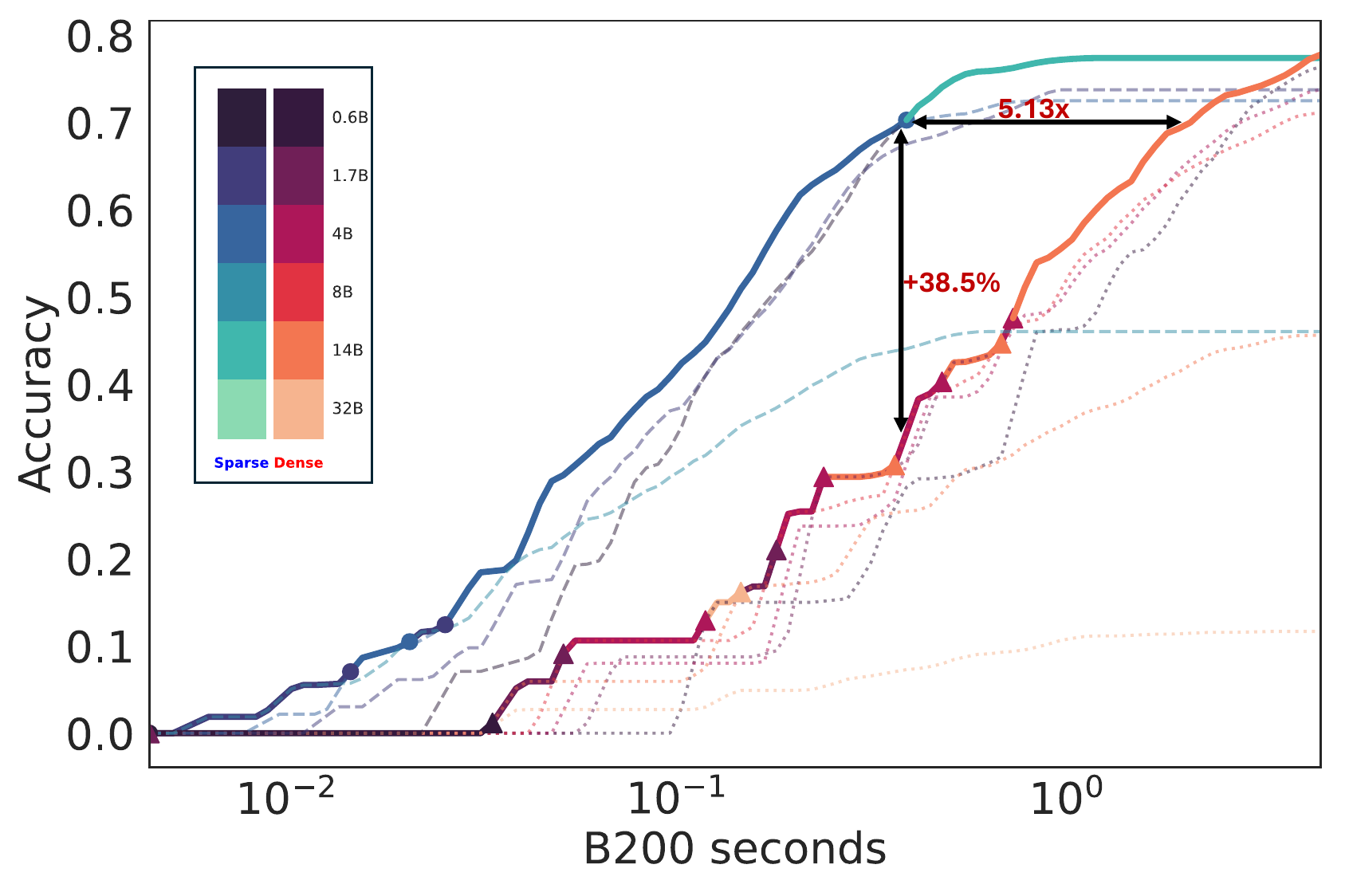}
    \label{fig: cot-bs-frontier} 
    }
     \subfloat[Sparse Algorithm Comparison]{
\includegraphics[width=0.32\linewidth]{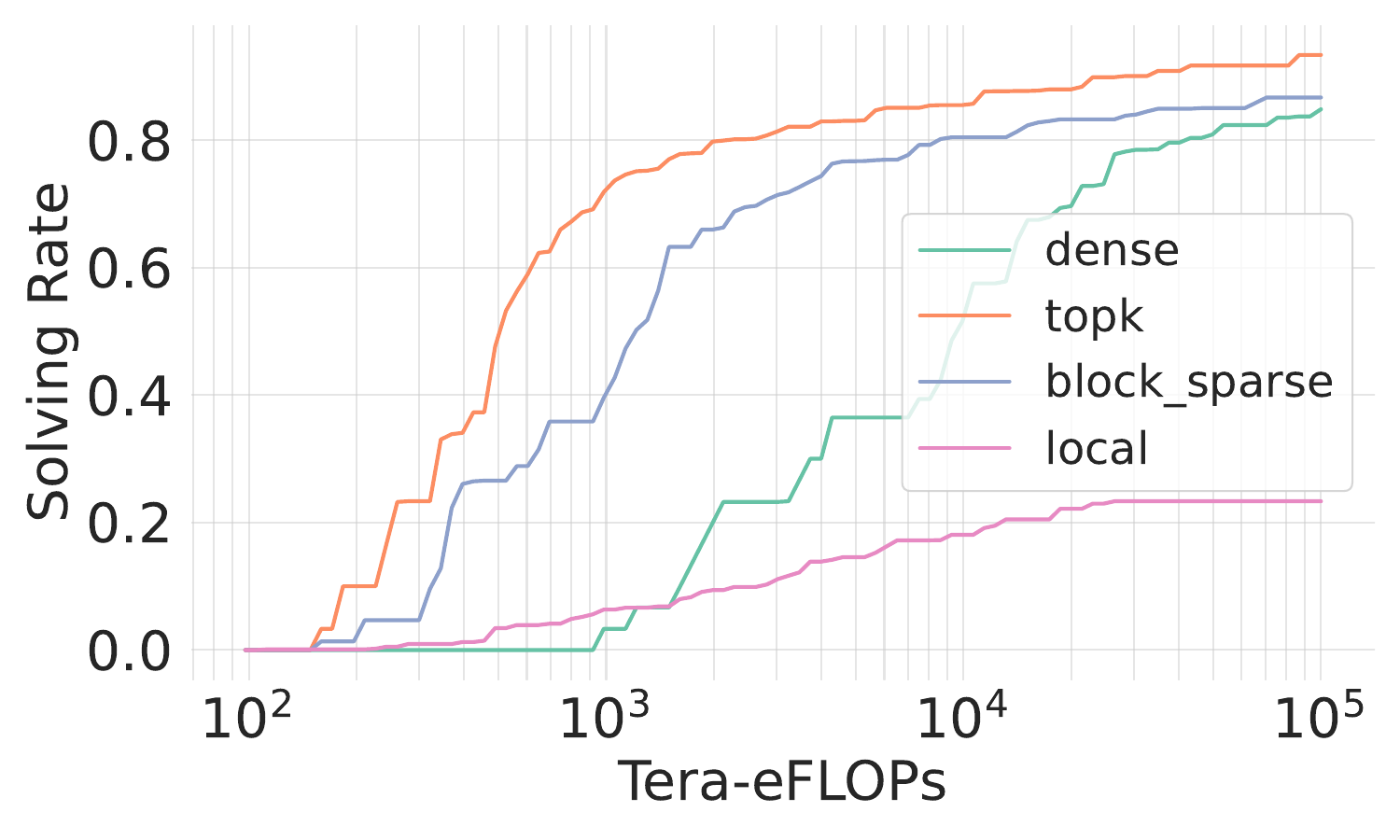}
    \label{fig: sparse comparison}
    }
    \caption{\textbf{Block top-$k$ attention}.  In \textbf{(a)} and \textbf{(b)}, we illustrate the optimality of block top-$k$ sparse attention in terms of TTS on AIME24 dataset. Although upper bounded by the oracle top-k attention performance, block top-$k$ achieves a good trade-off between effectiveness and tractability. Although easy to implement, the performance of local attention is poor \textbf{(c)}.
    }
    \vspace{-0.6cm}
\end{figure*}
In this section, we demonstrate the practicality of \sparselaw through block top-$k$ attention. We first show the scaling of block top-$k$ attention, which is even comparable to the oracle top-$k$ attention. Then we report empirical improvements in task throughput (number of tasks performed per unit time) using our block-sparse implementation. In addition, we conduct ablation studies with alternative sparsification strategies, such as local attention, to highlight the importance of the KV selection mechanism.

\subsection{Block Top-$k$ Attention}
\begin{wrapfigure}{r}{0.7\textwidth}
  \centering \includegraphics[width=0.95\linewidth]{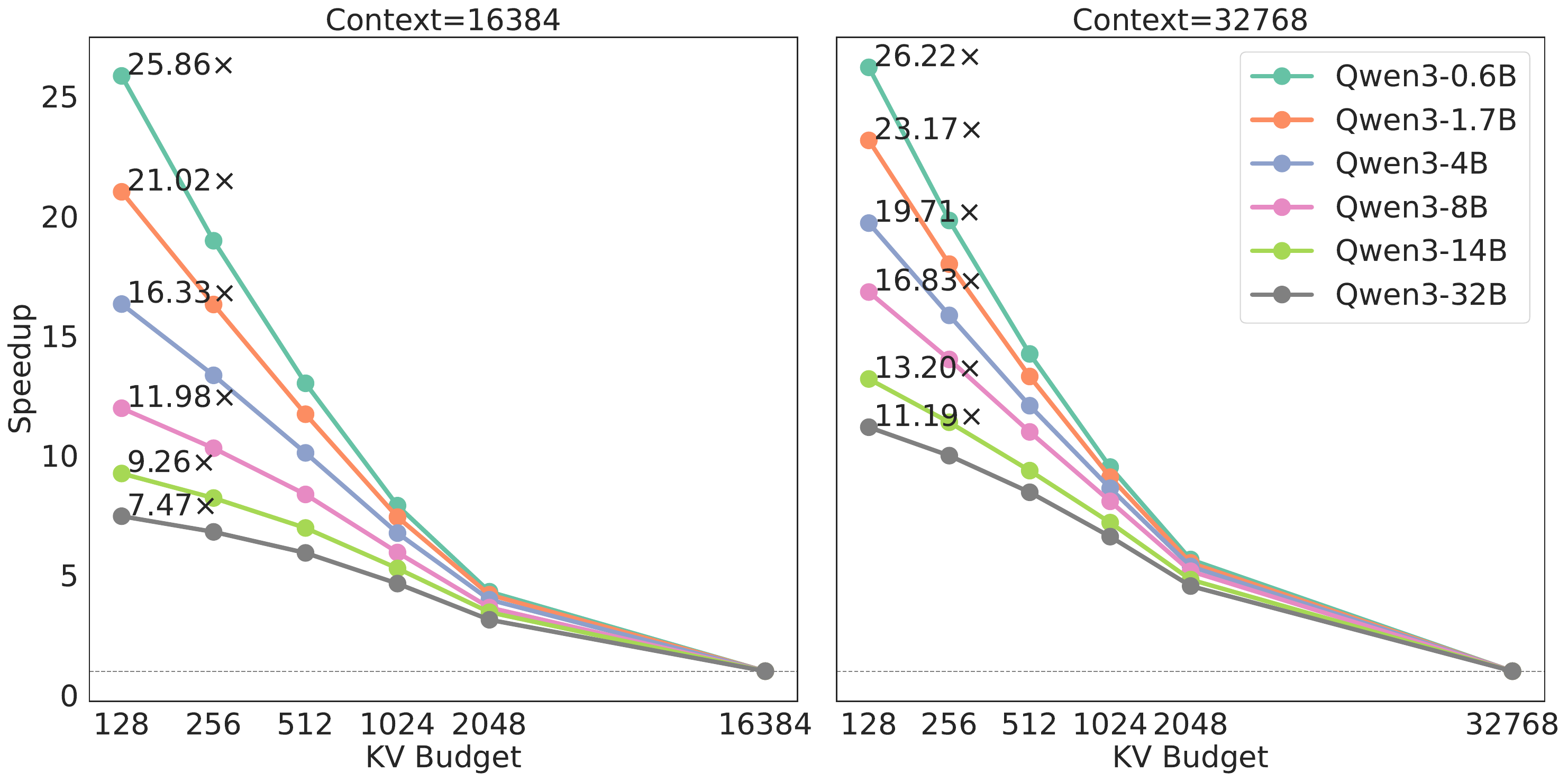}
  \caption{Throughput improvement with block top-$k$ attention.}
  \label{fig:sparse_attention}
  \vspace{-1em} 
\end{wrapfigure}
While top-$k$ attention offers attractive theoretical scaling, it is computationally intractable in practice. Instead, we adopt block top-$k$ attention for two key reasons. First, it exploits temporal locality in attention patterns~\citep{sun2024triforcelosslessaccelerationlong} to retrieve semantically related key-value (KV) blocks, reaching high accuracy. Second, its localized retrieval is hardware-efficient and integrates seamlessly with paged attention~\citep{kwon2023efficientmemorymanagementlarge}, enabling high-throughput decoding. Block top-$k$ attention is proved efficiently implementable in massive prior work~\citep{tang2024quest,sun2024shadowkv,xu2025xattention,zaheer2020bigbird,yuan2025native}. In practice, we compute a representative vector for each KV block by averaging its key vectors, and use these to score the relevance of blocks to each query. Importance scores are shared across query heads within a group, following the Grouped Query Attention (GQA) scheme. The definition of block top-$k$ attention is introduced in detail in~\Cref{topkandblocktopk}.

\textbf{Performance and Scaling.}  \textbf{First,} as shown in~\Cref{fig: bon-bs-frontier,fig: cot-bs-frontier}, block top-$k$ attention demonstrates comparable scaling to the oracle top-$k$ attention, improving accuracy by 45 points in the low-cost regime and achieving equivalent accuracy while using $8.58\times$ fewer resources compared to dense attention. More accuracy evaluations across various benchmarks (including with MoE models) are presented in~\Cref{sparsescalinglaw}. \textbf{Second}, we compare block top-$k$ with local attention in~\Cref{fig: sparse comparison}. Although local attention is more efficient due to its static sparsity pattern, it performs significantly worse. Its poor test-time scaling prevents it from outperforming dense attention except in very low-accuracy regimes.

\subsection{Implementation and Empirical Results}

\textbf{Implementation}. To demonstrate the practical efficiency gain of sparse attention, we build our attention backend on Flashinfer~\citep{ye2025flashinfer} and torch compile\footnote{\url{https://docs.pytorch.org/tutorials/intermediate/torch_compile_tutorial_.html}}. Alongside the paged KV cache, we introduce an auxiliary data structure to store block-level average key vectors. The KV block size is chosen such that the memory load from the block-average vectors and the selected top-$k$ KV blocks remains balanced. This design enables sub-quadratic KV loading cost as the number of reasoning tokens increases. Rather than constructing a full end-to-end serving system, we estimate the overall model execution time using per-layer latency and throughput measurements~\citep{he2024fastdecode}. 

\textbf{Results.} We illustrate the benefit of block top-$k$ attention across different model sizes on H200 nodes (8 GPUs per node) with  batch size and context length of (4096, 16384) and (2048, 32768). Here we assume uniform workload of tasks with similar context lengths and generation lengths. As shown in~\Cref{fig:sparse_attention}, block top-$k$ attention substantially improves inference throughput, particularly for smaller models. For instance, the Qwen3-0.6B model achieves a $25.9\sim26.2\times$ increase in throughput. This improvement reflects the growing inefficiency of dense attention at longer contexts, which disproportionately affects smaller models. 

The substantial improvement in throughput highlights the potential for corresponding gains in \textit{task-level throughput}, when appropriately co-designed with inference systems and test-time strategies. We leave this direction for future work.

\section{Conclusion and Discussion}
\label{sec:conclusion}
This work introduces the \textit{Kinetics Scaling Law} based on the insight that attention costs rather than parameter counts are the dominant factor in TTS. We demonstrate that sparse attention fundamentally reshapes the scaling landscape, enabling longer generations and significantly higher accuracy. We envision the \law as a foundational tool for guiding end-to-end design across LLM serving, agent frameworks, and reinforcement learning environments. \sparselaw may signal a new paradigm, enabling continued progress even beyond pretraining plateaus. While our analysis centers on NVIDIA GPUs, the underlying principle that scaling memory bandwidth is more challenging and costly than scaling FLOPs applies broadly across hardware platforms. Ultimately, this study highlights the need for co-designing model architectures, test-time inference techniques, and hardware infrastructure as a critical step toward enabling the next wave of scalable LLM deployment.

\section*{Limitations, Future Scope, and Broader Impact}
\label{limitationsandimpact}
\paragraph{Limitations.}  
Our experiments primarily focus on \textbf{Qwen3}~\citep{yang2025qwen3technicalreport} and \textbf{DeepSeek-R1-Distilled-Qwen}~\citep{guo2025deepseek}, two state-of-the-art pretrained reasoning model series, evaluated from the inference perspective. However, the effects of training and post-training strategies are not fully explored and may influence the performance gaps and robustness to sparse attention mechanisms. In addition, our cost analysis assumes a cloud-based serving environment, where computational resources are typically sufficient and large batch sizes are feasible. In contrast, local deployment scenarios, such as those using ollama,\footnote{\url{https://github.com/ollama/ollama}} often face limited VRAM where access to model parameters can dominate inference costs. Smaller models may be more appropriate in such settings, and our findings may not fully extend to these use cases.

\captionsetup[sub]{justification=centering}
\begin{figure*}[h]
    \centering
    \subfloat[gen length vs $N_{opt}$ correlation \\ (top-$k$ attention)]{
\includegraphics[width=0.4\linewidth]{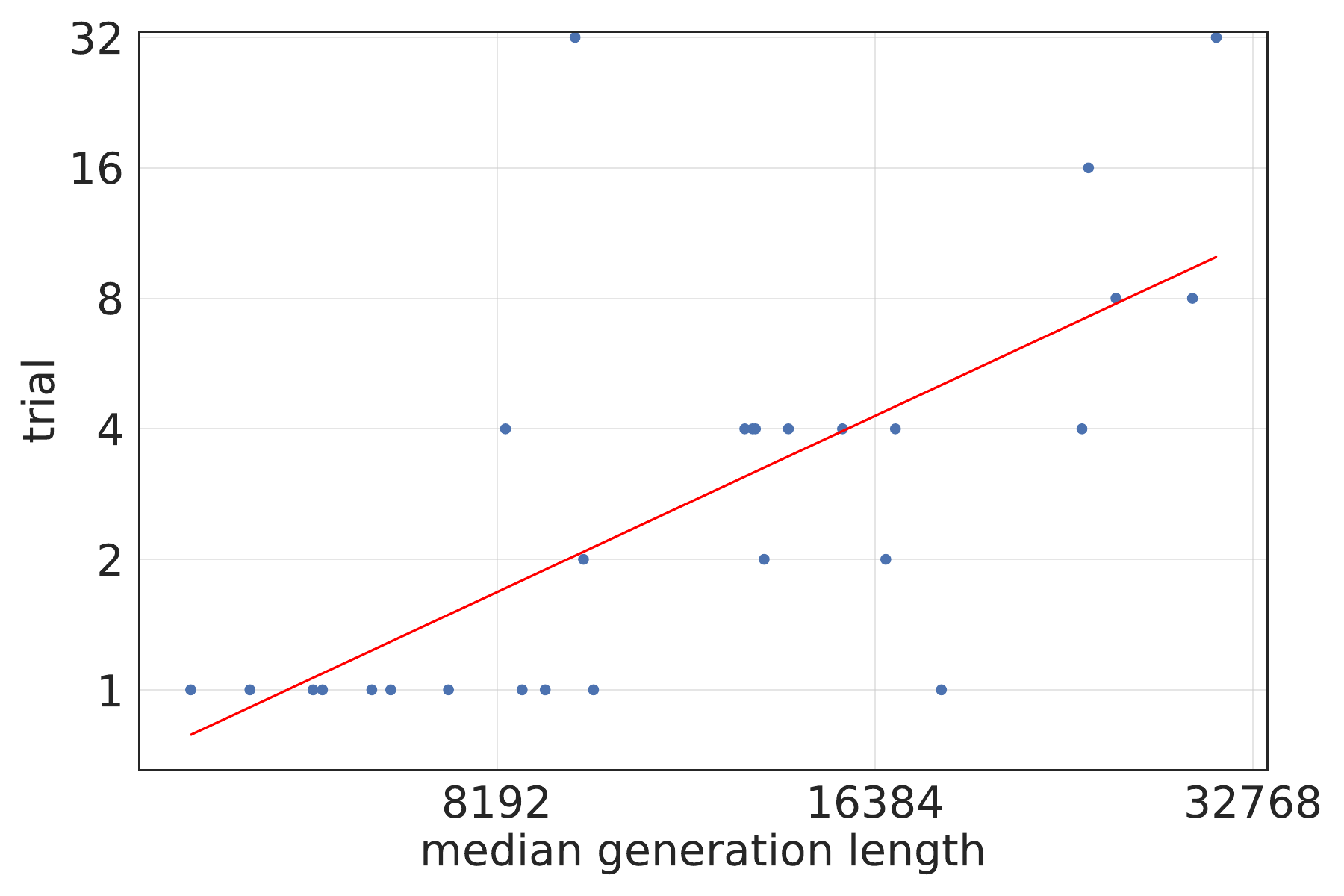}
    \label{fig: bon-topk-correlation}
    }
  \subfloat[gen length vs $N_{opt}$ correlation \\ (block top-$k$ attention)]{
\includegraphics[width=0.4\linewidth]{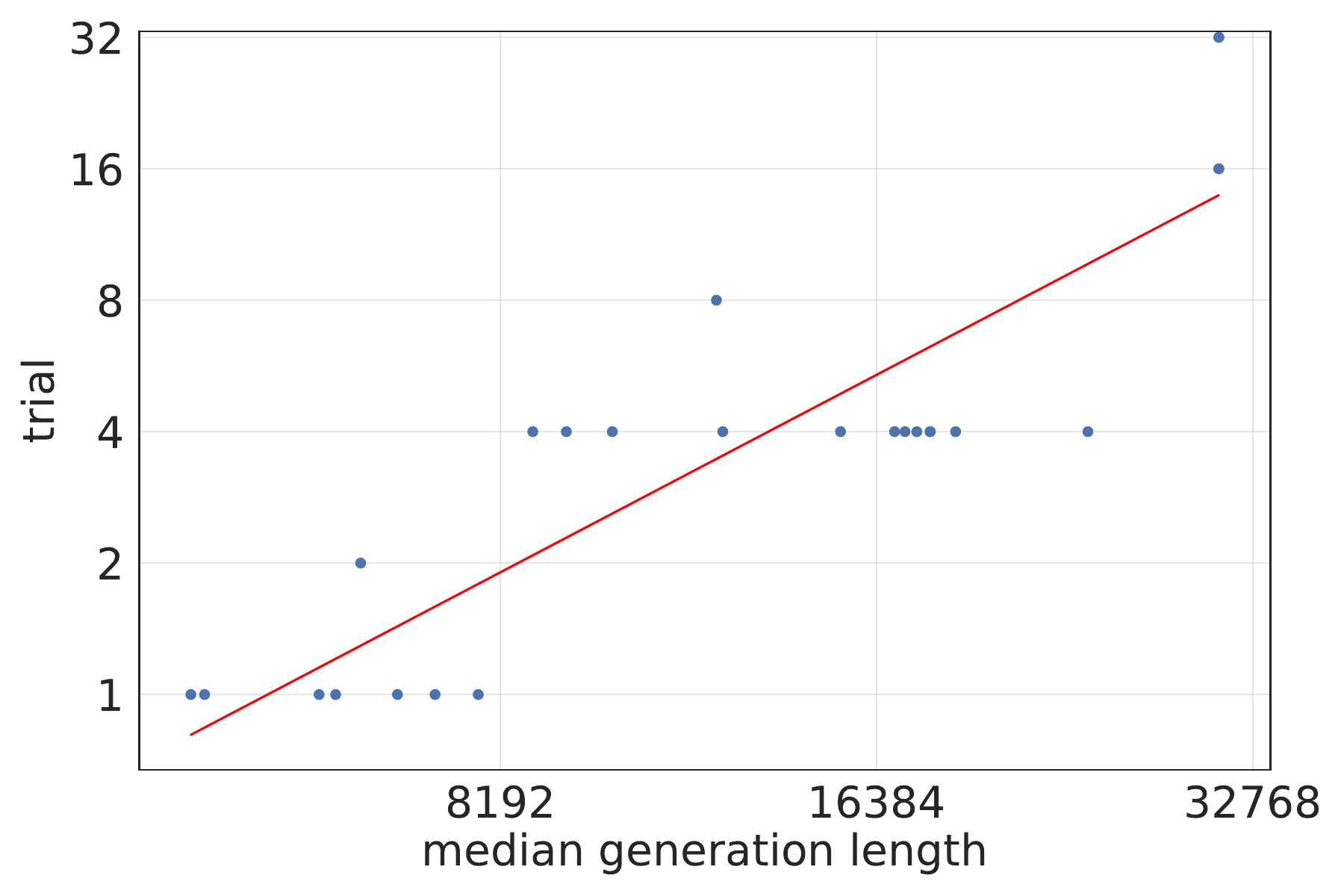}
    \label{fig: bon-blocktopk-correlation} 
    }
\caption{\textbf{Correlation between Generation Length and Number of Trials.} Longer generations correlate strongly with the optimal number of trials ($N_{opt})$, serving as a proxy for problem difficulty. \textbf{(a)} shows this trend for top-$k$ and \textbf{(b)} block top-$k$ attention on the AIME24 dataset using the Qwen3-8B model.}

\label{fig: correlation}
\end{figure*}

\paragraph{Future Scope.}
Our sparse scaling law offers valuable insights for enriching the applications of sparse attention algorithms and the design space of test-time scaling strategies. On one hand, except for top-$k$, currently we only discuss a simple variant, i.e., block top-$k$, and have already demonstrated strong scalability. There exist more advanced sparse attention algorithms~\citep{tang2024quest,chen2024magicpig,yuan2025native,lin2025twilight} which may push test-time scaling to a much higher boundary. On the other hand, test-time scaling algorithms are proposed to adaptively allocate computation to tasks, or even to tokens~\citep{arora2502training,mohtashami2023cotformer,ma2025cot,ma2025reasoning}. Extending them towards to new resource allocation problems in sparse attention is critical to reach the limit of \sparselaw. For instance, since generation length strongly correlates with the optimal number of trials under sparse attention (as shown in~\Cref{fig: correlation}), it can be used as a dynamic signal to adjust the number of trials and KV budget. Moreover, sparse attention drastically reduces inference cost, enabling more reasoning trials and longer generations. This unlocks greater flexibility in configuring TTS strategies within a fixed resource budget.

\paragraph{Broader Impact.}  
This work aims to contribute to the understanding of efficiency and scalability challenges in the test-time scaling era, spanning model architecture, system-level implementation, and hardware design. We highlight the central role of sparsity in addressing these challenges. In addition, by shifting the focus from token-level metrics to \textit{task-level throughput}, we open up broader opportunities for algorithm–system co-design and emphasize the real-world utility of generative AI for society.  Our study is algorithmic in nature and does not target specific applications. While large language models can be misused in harmful ways, this work does not introduce new capabilities or risks beyond those already present in existing systems. Test-time scaling  can consume a substantial amount of energy, raising concerns about the environmental sustainability of widespread deployment. By promoting sparse attention, our work hopes to help to reduce the carbon footprint and energy consumption of inference systems and support the broader goal of sustainable AI.

\section*{Acknowledgements}
We would like to thank Yong Wu, Xinyu Yang and Harry Dong for providing us constructive feedback on our paper and computing resources of NVIDIA. This work was partially supported by Google Research Award, Amazon Research Award, Intel, Li Auto, Moffett AI, and CMU CyLab Seed funding. This material is also based upon work supported by the National Science Foundation under Grant No. 2326610. Any opinions, findings, and conclusions or recommendations expressed are those of the authors and do not necessarily reflect the views of the National Science Foundation.
\clearpage
\clearpage
\newpage
\bibliographystyle{assets/plainnat}
\bibliography{paper}

\begin{thebibliography}{110}
\providecommand{\natexlab}[1]{#1}
\providecommand{\url}[1]{\texttt{#1}}
\expandafter\ifx\csname urlstyle\endcsname\relax
  \providecommand{\doi}[1]{doi: #1}\else
  \providecommand{\doi}{doi: \begingroup \urlstyle{rm}\Url}\fi

\bibitem[AI@Meta(2025)]{llama4modelcard}
AI@Meta.
\newblock Llama 4 model card.
\newblock 2025.
\newblock \url{https://github.com/meta-llama/llama-models/blob/main/models/llama4/MODEL_CARD.md}.

\bibitem[Arora and Zanette()]{arora2502training}
Daman Arora and Andrea Zanette.
\newblock Training language models to reason efficiently, 2025.
\newblock \emph{URL https://arxiv. org/abs/2502.04463}.

\bibitem[Beeching et~al.()Beeching, Tunstall, and Rush]{beeching2024scalingtesttimecompute}
Edward Beeching, Lewis Tunstall, and Sasha Rush.
\newblock Scaling test-time compute with open models.
\newblock \url{https://huggingface.co/spaces/HuggingFaceH4/blogpost-scaling-test-time-compute}.

\bibitem[Beltagy et~al.(2020)Beltagy, Peters, and Cohan]{beltagy2020longformer}
Iz~Beltagy, Matthew~E Peters, and Arman Cohan.
\newblock Longformer: The long-document transformer.
\newblock \emph{arXiv preprint arXiv:2004.05150}, 2020.

\bibitem[Brown et~al.(2024)Brown, Juravsky, Ehrlich, Clark, Le, R{\'e}, and Mirhoseini]{brown2024large}
Bradley Brown, Jordan Juravsky, Ryan Ehrlich, Ronald Clark, Quoc~V Le, Christopher R{\'e}, and Azalia Mirhoseini.
\newblock Large language monkeys: Scaling inference compute with repeated sampling.
\newblock \emph{arXiv preprint arXiv:2407.21787}, 2024.

\bibitem[Cai et~al.(2024)Cai, Tian, Wang, and Chen]{cai2024lococo}
Ruisi Cai, Yuandong Tian, Zhangyang Wang, and Beidi Chen.
\newblock Lococo: Dropping in convolutions for long context compression.
\newblock \emph{arXiv preprint arXiv:2406.05317}, 2024.

\bibitem[Chen et~al.(2021)Chen, Dao, Winsor, Song, Rudra, and R{\'e}]{chen2021scatterbrain}
Beidi Chen, Tri Dao, Eric Winsor, Zhao Song, Atri Rudra, and Christopher R{\'e}.
\newblock Scatterbrain: Unifying sparse and low-rank attention.
\newblock \emph{Advances in Neural Information Processing Systems}, 34:\penalty0 17413--17426, 2021.

\bibitem[Chen et~al.(2023)Chen, Borgeaud, Irving, Lespiau, Sifre, and Jumper]{chen2023accelerating}
Charlie Chen, Sebastian Borgeaud, Geoffrey Irving, Jean-Baptiste Lespiau, Laurent Sifre, and John Jumper.
\newblock Accelerating large language model decoding with speculative sampling.
\newblock \emph{arXiv preprint arXiv:2302.01318}, 2023.

\bibitem[Chen et~al.(2025{\natexlab{a}})Chen, Cusumano-Towner, Huval, Petrenko, Hamburger, Koltun, and Kr{\"a}henb{\"u}hl]{chen2025reinforcement}
Kevin Chen, Marco Cusumano-Towner, Brody Huval, Aleksei Petrenko, Jackson Hamburger, Vladlen Koltun, and Philipp Kr{\"a}henb{\"u}hl.
\newblock Reinforcement learning for long-horizon interactive llm agents.
\newblock \emph{arXiv preprint arXiv:2502.01600}, 2025{\natexlab{a}}.

\bibitem[Chen et~al.(2025{\natexlab{b}})Chen, Hui, Cui, Yang, Liu, Sun, Lin, and Liu]{ParScale}
Mouxiang Chen, Binyuan Hui, Zeyu Cui, Jiaxi Yang, Dayiheng Liu, Jianling Sun, Junyang Lin, and Zhongxin Liu.
\newblock Parallel scaling law for language models.
\newblock \emph{arXiv preprint arXiv:2505.10475}, 2025{\natexlab{b}}.
\newblock \url{https://arxiv.org/abs/2505.10475}.

\bibitem[Chen et~al.(2024)Chen, Sadhukhan, Ye, Zhou, Zhang, Nolte, Tian, Douze, Bottou, Jia, et~al.]{chen2024magicpig}
Zhuoming Chen, Ranajoy Sadhukhan, Zihao Ye, Yang Zhou, Jianyu Zhang, Niklas Nolte, Yuandong Tian, Matthijs Douze, Leon Bottou, Zhihao Jia, et~al.
\newblock Magicpig: Lsh sampling for efficient llm generation.
\newblock \emph{arXiv preprint arXiv:2410.16179}, 2024.

\bibitem[Child et~al.(2019)Child, Gray, Radford, and Sutskever]{child2019generating}
Rewon Child, Scott Gray, Alec Radford, and Ilya Sutskever.
\newblock Generating long sequences with sparse transformers.
\newblock \emph{arXiv preprint arXiv:1904.10509}, 2019.

\bibitem[Choromanski et~al.(2020)Choromanski, Likhosherstov, Dohan, Song, Gane, Sarlos, Hawkins, Davis, Mohiuddin, Kaiser, et~al.]{choromanski2020rethinking}
Krzysztof Choromanski, Valerii Likhosherstov, David Dohan, Xingyou Song, Andreea Gane, Tamas Sarlos, Peter Hawkins, Jared Davis, Afroz Mohiuddin, Lukasz Kaiser, et~al.
\newblock Rethinking attention with performers.
\newblock \emph{arXiv preprint arXiv:2009.14794}, 2020.

\bibitem[Dai et~al.(2024)Dai, Deng, Zhao, Xu, Gao, Chen, Li, Zeng, Yu, Wu, et~al.]{dai2024deepseekmoe}
Damai Dai, Chengqi Deng, Chenggang Zhao, RX~Xu, Huazuo Gao, Deli Chen, Jiashi Li, Wangding Zeng, Xingkai Yu, Yu~Wu, et~al.
\newblock Deepseekmoe: Towards ultimate expert specialization in mixture-of-experts language models.
\newblock \emph{arXiv preprint arXiv:2401.06066}, 2024.

\bibitem[Dao(2023)]{flash_attn2}
Tri Dao.
\newblock Flashattention-2: Faster attention with better parallelism and work partitioning.
\newblock \emph{CoRR}, abs/2307.08691, 2023.
\newblock \doi{10.48550/ARXIV.2307.08691}.
\newblock \url{https://doi.org/10.48550/arXiv.2307.08691}.

\bibitem[Dao et~al.(2021)Dao, Chen, Liang, Yang, Song, Rudra, and R{\'e}]{dao2021pixelated}
Tri Dao, Beidi Chen, Kaizhao Liang, Jiaming Yang, Zhao Song, Atri Rudra, and Christopher R{\'e}.
\newblock Pixelated butterfly: Simple and efficient sparse training for neural network models.
\newblock In \emph{International Conference on Learning Representations (ICLR)}, 2021.
\newblock \url{https://arxiv.org/abs/2112.00029}.

\bibitem[Dao et~al.(2022)Dao, Fu, Ermon, Rudra, and R{\'{e}}]{flash_attn}
Tri Dao, Daniel~Y. Fu, Stefano Ermon, Atri Rudra, and Christopher R{\'{e}}.
\newblock Flashattention: Fast and memory-efficient exact attention with io-awareness.
\newblock In Sanmi Koyejo, S.~Mohamed, A.~Agarwal, Danielle Belgrave, K.~Cho, and A.~Oh, editors, \emph{Advances in Neural Information Processing Systems 35: Annual Conference on Neural Information Processing Systems 2022, NeurIPS 2022, New Orleans, LA, USA, November 28 - December 9, 2022}, 2022.

\bibitem[Daras et~al.(2020)Daras, Kitaev, Odena, and Dimakis]{daras2020smyrf}
Giannis Daras, Nikita Kitaev, Augustus Odena, and Alexandros~G Dimakis.
\newblock Smyrf: Efficient attention using asymmetric clustering.
\newblock \emph{arXiv preprint arXiv:2010.05315}, 2020.

\bibitem[DeepSeek-AI(2025)]{deepseekinfra}
DeepSeek-AI.
\newblock Deepseek open infra index.
\newblock 2025.
\newblock \url{https://github.com/deepseek-ai/open-infra-index/blob/main/202502OpenSourceWeek/day_6_one_more_thing_deepseekV3R1_inference_system_overview.md}.

\bibitem[Dettmers et~al.(2022)Dettmers, Lewis, Belkada, and Zettlemoyer]{dettmers2022gpt3}
Tim Dettmers, Mike Lewis, Younes Belkada, and Luke Zettlemoyer.
\newblock Gpt3. int8 (): 8-bit matrix multiplication for transformers at scale.
\newblock \emph{Advances in neural information processing systems}, 35:\penalty0 30318--30332, 2022.

\bibitem[Driess et~al.(2023)Driess, Nguyen, Xia, et~al.]{driess2023palme}
Danny Driess, Minh Nguyen, Fei Xia, et~al.
\newblock Palm-e: An embodied multimodal language model.
\newblock \emph{arXiv preprint arXiv:2303.03378}, 2023.

\bibitem[Du et~al.(2021)Du, Huang, Dai, Tong, Lepikhin, Xu, Chen, Wu, and Dean]{du2021glam}
Nan Du, Yanping Huang, Andrew~M. Dai, Simon Tong, Dmitry Lepikhin, Yuanzhong Xu, Dehao Chen, Yonghui Wu, and Jeff Dean.
\newblock Glam: Efficient scaling of language models with mixture-of-experts.
\newblock \emph{arXiv preprint arXiv:2112.06905}, 2021.

\bibitem[Fedus et~al.(2022)Fedus, Zoph, and Shazeer]{fedus2022switch}
William Fedus, Barret Zoph, and Noam Shazeer.
\newblock Switch transformers: Scaling to trillion parameter models with simple and efficient sparsity.
\newblock \emph{Journal of Machine Learning Research}, 23\penalty0 (1):\penalty0 5232--5270, 2022.

\bibitem[Feng et~al.(2023)Feng, Wan, Wen, McAleer, Wen, Zhang, and Wang]{feng2023alphazero}
Xidong Feng, Ziyu Wan, Muning Wen, Stephen~Marcus McAleer, Ying Wen, Weinan Zhang, and Jun Wang.
\newblock Alphazero-like tree-search can guide large language model decoding and training.
\newblock \emph{arXiv preprint arXiv:2309.17179}, 2023.

\bibitem[Frantar and Alistarh(2023)]{frantar2023sparsegpt}
Elias Frantar and Dan Alistarh.
\newblock Sparsegpt: Massive language models can be accurately pruned in one-shot.
\newblock In \emph{International Conference on Machine Learning}, pages 10323--10337. PMLR, 2023.

\bibitem[Frantar et~al.(2022)Frantar, Ashkboos, Hoefler, and Alistarh]{frantar2022gptq}
Elias Frantar, Saleh Ashkboos, Torsten Hoefler, and Dan Alistarh.
\newblock Gptq: Accurate post-training quantization for generative pre-trained transformers.
\newblock \emph{arXiv preprint arXiv:2210.17323}, 2022.

\bibitem[Fu et~al.(2024)Fu, Chen, Zhu, Fu, Dai, Qiao, and Zhang]{fu2024efficiently}
Yichao Fu, Junda Chen, Siqi Zhu, Zheyu Fu, Zhongdongming Dai, Aurick Qiao, and Hao Zhang.
\newblock Efficiently serving llm reasoning programs with certaindex.
\newblock \emph{arXiv preprint arXiv:2412.20993}, 2024.

\bibitem[Grattafiori et~al.(2024)Grattafiori, Dubey, Jauhri, Pandey, Kadian, Al-Dahle, Letman, Mathur, Schelten, Vaughan, et~al.]{grattafiori2024llama}
Aaron Grattafiori, Abhimanyu Dubey, Abhinav Jauhri, Abhinav Pandey, Abhishek Kadian, Ahmad Al-Dahle, Aiesha Letman, Akhil Mathur, Alan Schelten, Alex Vaughan, et~al.
\newblock The llama 3 herd of models.
\newblock \emph{arXiv preprint arXiv:2407.21783}, 2024.

\bibitem[Gu and Dao(2023)]{mamba}
Albert Gu and Tri Dao.
\newblock Mamba: Linear-time sequence modeling with selective state spaces.
\newblock \emph{CoRR}, abs/2312.00752, 2023.
\newblock \doi{10.48550/ARXIV.2312.00752}.
\newblock \url{https://doi.org/10.48550/arXiv.2312.00752}.

\bibitem[Gu et~al.(2022)Gu, Goel, and R{\'{e}}]{s4}
Albert Gu, Karan Goel, and Christopher R{\'{e}}.
\newblock Efficiently modeling long sequences with structured state spaces.
\newblock In \emph{The Tenth International Conference on Learning Representations, {ICLR} 2022, Virtual Event, April 25-29, 2022}. OpenReview.net, 2022.
\newblock \url{https://openreview.net/forum?id=uYLFoz1vlAC}.

\bibitem[Guo et~al.(2025)Guo, Yang, Zhang, Song, Zhang, Xu, Zhu, Ma, Wang, Bi, et~al.]{guo2025deepseek}
Daya Guo, Dejian Yang, Haowei Zhang, Junxiao Song, Ruoyu Zhang, Runxin Xu, Qihao Zhu, Shirong Ma, Peiyi Wang, Xiao Bi, et~al.
\newblock Deepseek-r1: Incentivizing reasoning capability in llms via reinforcement learning.
\newblock \emph{arXiv preprint arXiv:2501.12948}, 2025.

\bibitem[Hao et~al.(2024)Hao, Sukhbaatar, Su, Li, Hu, Weston, and Tian]{hao2024training}
Shibo Hao, Sainbayar Sukhbaatar, DiJia Su, Xian Li, Zhiting Hu, Jason Weston, and Yuandong Tian.
\newblock Training large language models to reason in a continuous latent space.
\newblock \emph{arXiv preprint arXiv:2412.06769}, 2024.

\bibitem[He and Zhai(2024)]{he2024fastdecode}
Jiaao He and Jidong Zhai.
\newblock Fastdecode: High-throughput gpu-efficient llm serving using heterogeneous pipelines.
\newblock \emph{arXiv preprint arXiv:2403.11421}, 2024.

\bibitem[Hoefler et~al.(2021)Hoefler, Alistarh, Ben-Nun, Dryden, and Peste]{hoefler2021sparsity}
Torsten Hoefler, Dan Alistarh, Tal Ben-Nun, Nikoli Dryden, and Alexandra Peste.
\newblock Sparsity in deep learning: Pruning and growth for efficient inference and training in neural networks.
\newblock \emph{Journal of Machine Learning Research}, 22\penalty0 (241):\penalty0 1--124, 2021.

\bibitem[Hoffmann et~al.(2022)Hoffmann, Borgeaud, Mensch, Buchatskaya, Cai, Rutherford, Casas, Hendricks, Welbl, Clark, et~al.]{hoffmann2022training}
Jordan Hoffmann, Sebastian Borgeaud, Arthur Mensch, Elena Buchatskaya, Trevor Cai, Eliza Rutherford, Diego de~Las Casas, Lisa~Anne Hendricks, Johannes Welbl, Aidan Clark, et~al.
\newblock Training compute-optimal large language models.
\newblock \emph{arXiv preprint arXiv:2203.15556}, 2022.

\bibitem[Hooper et~al.(2024)Hooper, Kim, Mohammadzadeh, Mahoney, Shao, Keutzer, and Gholami]{NEURIPS2024_028fcbcf}
Coleman Hooper, Sehoon Kim, Hiva Mohammadzadeh, Michael~W. Mahoney, Yakun~Sophia Shao, Kurt Keutzer, and Amir Gholami.
\newblock Kvquant: Towards 10 million context length llm inference with kv cache quantization.
\newblock In A.~Globerson, L.~Mackey, D.~Belgrave, A.~Fan, U.~Paquet, J.~Tomczak, and C.~Zhang, editors, \emph{Advances in Neural Information Processing Systems}, volume~37, pages 1270--1303. Curran Associates, Inc., 2024.
\newblock \url{https://proceedings.neurips.cc/paper_files/paper/2024/file/028fcbcf85435d39a40c4d61b42c99a4-Paper-Conference.pdf}.

\bibitem[Hu et~al.(2025)Hu, Huang, Wang, Li, Hu, Liu, Chen, Xie, and Shan]{hu2025efficient}
Junhao Hu, Wenrui Huang, Weidong Wang, Zhenwen Li, Tiancheng Hu, Zhixia Liu, Xusheng Chen, Tao Xie, and Yizhou Shan.
\newblock Efficient long-decoding inference with reasoning-aware attention sparsity.
\newblock \emph{arXiv preprint arXiv:2502.11147}, 2025.

\bibitem[Huang et~al.(2022)Huang, Fei, and Finn]{huang2022language}
Wenlong Huang, Fei Fei, and Chelsea Finn.
\newblock Language models as zero-shot planners: Extracting actionable knowledge for embodied agents.
\newblock \emph{arXiv preprint arXiv:2201.07207}, 2022.

\bibitem[Jaech et~al.(2024)Jaech, Kalai, Lerer, Richardson, El-Kishky, Low, Helyar, Madry, Beutel, Carney, et~al.]{jaech2024openai}
Aaron Jaech, Adam Kalai, Adam Lerer, Adam Richardson, Ahmed El-Kishky, Aiden Low, Alec Helyar, Aleksander Madry, Alex Beutel, Alex Carney, et~al.
\newblock Openai o1 system card.
\newblock \emph{arXiv preprint arXiv:2412.16720}, 2024.

\bibitem[Jain et~al.(2024)Jain, Han, Gu, Li, Yan, Zhang, Wang, Solar-Lezama, Sen, and Stoica]{jain2024livecodebench}
Naman Jain, King Han, Alex Gu, Wen-Ding Li, Fanjia Yan, Tianjun Zhang, Sida Wang, Armando Solar-Lezama, Koushik Sen, and Ion Stoica.
\newblock Livecodebench: Holistic and contamination free evaluation of large language models for code.
\newblock \emph{arXiv preprint arXiv:2403.07974}, 2024.

\bibitem[Jiang et~al.(2024)Jiang, Sablayrolles, Roux, Mensch, Savary, Bamford, Chaplot, Casas, Hanna, Bressand, et~al.]{jiang2024mixtral}
Albert~Q Jiang, Alexandre Sablayrolles, Antoine Roux, Arthur Mensch, Blanche Savary, Chris Bamford, Devendra~Singh Chaplot, Diego de~las Casas, Emma~Bou Hanna, Florian Bressand, et~al.
\newblock Mixtral of experts.
\newblock \emph{arXiv preprint arXiv:2401.04088}, 2024.

\bibitem[Juravsky et~al.(2024)Juravsky, Brown, Ehrlich, Fu, R{\'e}, and Mirhoseini]{juravsky2024hydragen}
Jordan Juravsky, Bradley Brown, Ryan Ehrlich, Daniel~Y Fu, Christopher R{\'e}, and Azalia Mirhoseini.
\newblock Hydragen: High-throughput llm inference with shared prefixes.
\newblock \emph{arXiv preprint arXiv:2402.05099}, 2024.

\bibitem[Kaplan et~al.(2020)Kaplan, McCandlish, Henighan, Brown, Chess, Child, Gray, Radford, Wu, and Amodei]{kaplan2020scaling}
Jared Kaplan, Sam McCandlish, Tom Henighan, Tom~B Brown, Benjamin Chess, Rewon Child, Scott Gray, Alec Radford, Jeffrey Wu, and Dario Amodei.
\newblock Scaling laws for neural language models.
\newblock \emph{arXiv preprint arXiv:2001.08361}, 2020.

\bibitem[Katharopoulos et~al.(2020)Katharopoulos, Vyas, Pappas, and Fleuret]{linear_attention}
Angelos Katharopoulos, Apoorv Vyas, Nikolaos Pappas, and Fran{\c{c}}ois Fleuret.
\newblock Transformers are rnns: Fast autoregressive transformers with linear attention.
\newblock In \emph{Proceedings of the 37th International Conference on Machine Learning, {ICML} 2020, 13-18 July 2020, Virtual Event}, volume 119 of \emph{Proceedings of Machine Learning Research}, pages 5156--5165. {PMLR}, 2020.
\newblock \url{http://proceedings.mlr.press/v119/katharopoulos20a.html}.

\bibitem[Kitaev et~al.(2020)Kitaev, Kaiser, and Levskaya]{kitaev2020reformer}
Nikita Kitaev, {\L}ukasz Kaiser, and Anselm Levskaya.
\newblock Reformer: The efficient transformer.
\newblock In \emph{The International Conference on Machine Learning ({ICML})}, 2020.

\bibitem[Kumar et~al.(2024)Kumar, Ankner, Spector, Bordelon, Muennighoff, Paul, Pehlevan, R{\'e}, and Raghunathan]{kumar2024scaling}
Tanishq Kumar, Zachary Ankner, Benjamin~F Spector, Blake Bordelon, Niklas Muennighoff, Mansheej Paul, Cengiz Pehlevan, Christopher R{\'e}, and Aditi Raghunathan.
\newblock Scaling laws for precision.
\newblock \emph{arXiv preprint arXiv:2411.04330}, 2024.

\bibitem[Kwon et~al.(2023)Kwon, Li, Zhuang, Sheng, Zheng, Yu, Gonzalez, Zhang, and Stoica]{kwon2023efficientmemorymanagementlarge}
Woosuk Kwon, Zhuohan Li, Siyuan Zhuang, Ying Sheng, Lianmin Zheng, Cody~Hao Yu, Joseph~E. Gonzalez, Hao Zhang, and Ion Stoica.
\newblock Efficient memory management for large language model serving with pagedattention, 2023.
\newblock \url{https://arxiv.org/abs/2309.06180}.

\bibitem[Leviathan et~al.(2023)Leviathan, Kalman, and Matias]{leviathan2023fast}
Yaniv Leviathan, Matan Kalman, and Yossi Matias.
\newblock Fast inference from transformers via speculative decoding.
\newblock In \emph{International Conference on Machine Learning}, pages 19274--19286. PMLR, 2023.

\bibitem[Li et~al.(2024)Li, Huang, Yang, Venkitesh, Locatelli, Ye, Cai, Lewis, and Chen]{li2024snapkvllmknowslooking}
Yuhong Li, Yingbing Huang, Bowen Yang, Bharat Venkitesh, Acyr Locatelli, Hanchen Ye, Tianle Cai, Patrick Lewis, and Deming Chen.
\newblock Snapkv: Llm knows what you are looking for before generation, 2024.
\newblock \url{https://arxiv.org/abs/2404.14469}.

\bibitem[Lieber et~al.(2024)Lieber, Lenz, Bata, Cohen, Osin, Dalmedigos, Safahi, Meirom, Belinkov, Shalev-Shwartz, Abend, Alon, Asida, Bergman, Glozman, Gokhman, Manevich, Ratner, Rozen, Shwartz, Zusman, and Shoham]{lieber2024jambahybridtransformermambalanguage}
Opher Lieber, Barak Lenz, Hofit Bata, Gal Cohen, Jhonathan Osin, Itay Dalmedigos, Erez Safahi, Shaked Meirom, Yonatan Belinkov, Shai Shalev-Shwartz, Omri Abend, Raz Alon, Tomer Asida, Amir Bergman, Roman Glozman, Michael Gokhman, Avashalom Manevich, Nir Ratner, Noam Rozen, Erez Shwartz, Mor Zusman, and Yoav Shoham.
\newblock Jamba: A hybrid transformer-mamba language model, 2024.
\newblock \url{https://arxiv.org/abs/2403.19887}.

\bibitem[Lin et~al.(2025)Lin, Tang, Yang, Wang, Tang, Tian, Stoica, Han, and Gao]{lin2025twilight}
Chaofan Lin, Jiaming Tang, Shuo Yang, Hanshuo Wang, Tian Tang, Boyu Tian, Ion Stoica, Song Han, and Mingyu Gao.
\newblock Twilight: Adaptive attention sparsity with hierarchical top-$ p $ pruning.
\newblock \emph{arXiv preprint arXiv:2502.02770}, 2025.

\bibitem[Lin et~al.(2024{\natexlab{a}})Lin, Tang, Tang, Yang, Chen, Wang, Xiao, Dang, Gan, and Han]{lin2024awq}
Ji~Lin, Jiaming Tang, Haotian Tang, Shang Yang, Wei-Ming Chen, Wei-Chen Wang, Guangxuan Xiao, Xingyu Dang, Chuang Gan, and Song Han.
\newblock Awq: Activation-aware weight quantization for on-device llm compression and acceleration.
\newblock \emph{Proceedings of Machine Learning and Systems}, 6:\penalty0 87--100, 2024{\natexlab{a}}.

\bibitem[Lin et~al.(2024{\natexlab{b}})Lin, Tang, Yang, Zhang, Xiao, Gan, and Han]{lin2024qserve}
Yujun Lin, Haotian Tang, Shang Yang, Zhekai Zhang, Guangxuan Xiao, Chuang Gan, and Song Han.
\newblock Qserve: W4a8kv4 quantization and system co-design for efficient llm serving.
\newblock \emph{arXiv preprint arXiv:2405.04532}, 2024{\natexlab{b}}.

\bibitem[Liu et~al.(2024{\natexlab{a}})Liu, Feng, Wang, Wang, Liu, Zhao, Dengr, Ruan, Dai, Guo, et~al.]{liu2024deepseek}
Aixin Liu, Bei Feng, Bin Wang, Bingxuan Wang, Bo~Liu, Chenggang Zhao, Chengqi Dengr, Chong Ruan, Damai Dai, Daya Guo, et~al.
\newblock Deepseek-v2: A strong, economical, and efficient mixture-of-experts language model.
\newblock \emph{arXiv preprint arXiv:2405.04434}, 2024{\natexlab{a}}.

\bibitem[Liu et~al.(2024{\natexlab{b}})Liu, Chen, Lu, Jiang, Han, Zhang, Chen, Zhang, Ding, Zhang, et~al.]{liu2024retrievalattention}
Di~Liu, Meng Chen, Baotong Lu, Huiqiang Jiang, Zhenhua Han, Qianxi Zhang, Qi~Chen, Chengruidong Zhang, Bailu Ding, Kai Zhang, et~al.
\newblock Retrievalattention: Accelerating long-context llm inference via vector retrieval.
\newblock \emph{arXiv preprint arXiv:2409.10516}, 2024{\natexlab{b}}.

\bibitem[Liu et~al.(2023)Liu, Wang, Dao, Zhou, Yuan, Song, Shrivastava, Zhang, Tian, Re, et~al.]{liu2023deja}
Zichang Liu, Jue Wang, Tri Dao, Tianyi Zhou, Binhang Yuan, Zhao Song, Anshumali Shrivastava, Ce~Zhang, Yuandong Tian, Christopher Re, et~al.
\newblock Deja vu: Contextual sparsity for efficient llms at inference time.
\newblock In \emph{International Conference on Machine Learning}, pages 22137--22176. PMLR, 2023.

\bibitem[Liu et~al.(2024{\natexlab{c}})Liu, Yuan, Jin, Zhong, Xu, Braverman, Chen, and Hu]{liu2024kivi}
Zirui Liu, Jiayi Yuan, Hongye Jin, Shaochen Zhong, Zhaozhuo Xu, Vladimir Braverman, Beidi Chen, and Xia Hu.
\newblock Kivi: A tuning-free asymmetric 2bit quantization for kv cache.
\newblock \emph{arXiv preprint arXiv:2402.02750}, 2024{\natexlab{c}}.

\bibitem[Ma et~al.(2025{\natexlab{a}})Ma, He, Snell, Griggs, Min, and Zaharia]{ma2025reasoning}
Wenjie Ma, Jingxuan He, Charlie Snell, Tyler Griggs, Sewon Min, and Matei Zaharia.
\newblock Reasoning models can be effective without thinking.
\newblock \emph{arXiv preprint arXiv:2504.09858}, 2025{\natexlab{a}}.

\bibitem[Ma et~al.(2025{\natexlab{b}})Ma, Wan, Yu, Fang, and Wang]{ma2025cot}
Xinyin Ma, Guangnian Wan, Runpeng Yu, Gongfan Fang, and Xinchao Wang.
\newblock Cot-valve: Length-compressible chain-of-thought tuning.
\newblock \emph{arXiv preprint arXiv:2502.09601}, 2025{\natexlab{b}}.

\bibitem[MAA(2024)]{aime24}
MAA.
\newblock American invitational mathematics examination 2024, 2024.
\newblock \url{https://artofproblemsolving.com/wiki/index.php/American_Invitational_Mathematics_Examination?srsltid=AfmBOoqiDCiaGTLQrsRTKsZui8RFnjOZqM4qIqY3yGB3sBaqOaxwf_Xt}.

\bibitem[MAA(2025)]{aime25}
MAA.
\newblock American invitational mathematics examination 2025, 2025.
\newblock \url{https://artofproblemsolving.com/wiki/index.php/American_Invitational_Mathematics_Examination?srsltid=AfmBOoqiDCiaGTLQrsRTKsZui8RFnjOZqM4qIqY3yGB3sBaqOaxwf_Xt}.

\bibitem[Mazar{\'e} et~al.(2025)Mazar{\'e}, Szilvasy, Lomeli, Massa, Murray, J{\'e}gou, and Douze]{mazare2025inference}
Pierre-Emmanuel Mazar{\'e}, Gergely Szilvasy, Maria Lomeli, Francisco Massa, Naila Murray, Herv{\'e} J{\'e}gou, and Matthijs Douze.
\newblock Inference-time sparse attention with asymmetric indexing.
\newblock \emph{arXiv preprint arXiv:2502.08246}, 2025.

\bibitem[Miao et~al.(2023)Miao, Oliaro, Zhang, Cheng, Wang, Zhang, Wong, Zhu, Yang, Shi, et~al.]{miao2023specinfer}
Xupeng Miao, Gabriele Oliaro, Zhihao Zhang, Xinhao Cheng, Zeyu Wang, Zhengxin Zhang, Rae Ying~Yee Wong, Alan Zhu, Lijie Yang, Xiaoxiang Shi, et~al.
\newblock Specinfer: Accelerating generative large language model serving with tree-based speculative inference and verification.
\newblock \emph{arXiv preprint arXiv:2305.09781}, 2023.

\bibitem[Mishra et~al.(2021)Mishra, Latorre, Pool, Stosic, Stosic, Venkatesh, Yu, and Micikevicius]{mishra2021accelerating}
Asit Mishra, Jorge~Albericio Latorre, Jeff Pool, Darko Stosic, Dusan Stosic, Ganesh Venkatesh, Chong Yu, and Paulius Micikevicius.
\newblock Accelerating sparse deep neural networks.
\newblock \emph{arXiv preprint arXiv:2104.08378}, 2021.

\bibitem[Mohtashami et~al.(2023)Mohtashami, Pagliardini, and Jaggi]{mohtashami2023cotformer}
Amirkeivan Mohtashami, Matteo Pagliardini, and Martin Jaggi.
\newblock Cotformer: A chain-of-thought driven architecture with budget-adaptive computation cost at inference.
\newblock \emph{arXiv preprint arXiv:2310.10845}, 2023.

\bibitem[Molchanov et~al.(2017)Molchanov, Ashukha, and Vetrov]{molchanov2017variational}
Dmitry Molchanov, Arsenii Ashukha, and Dmitry Vetrov.
\newblock Variational dropout sparsifies deep neural networks.
\newblock In \emph{Proceedings of the 34th International Conference on Machine Learning}, pages 2498--2507. PMLR, 2017.

\bibitem[Muennighoff et~al.(2025)Muennighoff, Yang, Shi, Li, Fei-Fei, Hajishirzi, Zettlemoyer, Liang, Cand{\`e}s, and Hashimoto]{muennighoff2025s1}
Niklas Muennighoff, Zitong Yang, Weijia Shi, Xiang~Lisa Li, Li~Fei-Fei, Hannaneh Hajishirzi, Luke Zettlemoyer, Percy Liang, Emmanuel Cand{\`e}s, and Tatsunori Hashimoto.
\newblock s1: Simple test-time scaling.
\newblock \emph{arXiv preprint arXiv:2501.19393}, 2025.

\bibitem[Nakano et~al.(2021)Nakano, Hilton, Wu, et~al.]{nakano2021webgpt}
Reiichiro Nakano, Jacob Hilton, Jeffrey Wu, et~al.
\newblock Webgpt: Browser-assisted question-answering with human feedback.
\newblock \emph{arXiv preprint arXiv:2112.09332}, 2021.

\bibitem[Nawrot et~al.(2025)Nawrot, Li, Huang, Ruder, Marchisio, and Ponti]{nawrot2025sparse}
Piotr Nawrot, Robert Li, Renjie Huang, Sebastian Ruder, Kelly Marchisio, and Edoardo~M Ponti.
\newblock The sparse frontier: Sparse attention trade-offs in transformer llms.
\newblock \emph{arXiv preprint arXiv:2504.17768}, 2025.

\bibitem[Ning et~al.(2023)Ning, Lin, Zhou, Wang, Yang, and Wang]{ning2023skeleton}
Xuefei Ning, Zinan Lin, Zixuan Zhou, Zifu Wang, Huazhong Yang, and Yu~Wang.
\newblock Skeleton-of-thought: Prompting llms for efficient parallel generation.
\newblock \emph{arXiv preprint arXiv:2307.15337}, 2023.

\bibitem[NovaSky-Team(2025)]{sky}
NovaSky-Team.
\newblock Sky-t1: Train your own o1 preview model within \$450.
\newblock https://novasky-ai.github.io/posts/sky-t1, 2025.
\newblock Accessed: 2025-01-09.

\bibitem[Paliotta et~al.(2025)Paliotta, Wang, Pagliardini, Li, Bick, Kolter, Gu, Fleuret, and Dao]{paliotta2025thinkingslowfastscaling}
Daniele Paliotta, Junxiong Wang, Matteo Pagliardini, Kevin~Y. Li, Aviv Bick, J.~Zico Kolter, Albert Gu, François Fleuret, and Tri Dao.
\newblock Thinking slow, fast: Scaling inference compute with distilled reasoners, 2025.
\newblock \url{https://arxiv.org/abs/2502.20339}.

\bibitem[Pope et~al.(2023)Pope, Douglas, Chowdhery, Devlin, Bradbury, Heek, Xiao, Agrawal, and Dean]{pope2023efficiently}
Reiner Pope, Sholto Douglas, Aakanksha Chowdhery, Jacob Devlin, James Bradbury, Jonathan Heek, Kefan Xiao, Shivani Agrawal, and Jeff Dean.
\newblock Efficiently scaling transformer inference.
\newblock \emph{Proceedings of Machine Learning and Systems}, 5:\penalty0 606--624, 2023.

\bibitem[Qwen-Team(2025)]{qwq32b}
Qwen-Team.
\newblock Qwq-32b: Embracing the power of reinforcement learning, March 2025.
\newblock \url{https://qwenlm.github.io/blog/qwq-32b/}.

\bibitem[Sadhukhan et~al.(2024)Sadhukhan, Chen, Chen, Tiwari, Lai, Shi, Yen, May, Chen, and Chen]{sadhukhan2024magicdec}
Ranajoy Sadhukhan, Jian Chen, Zhuoming Chen, Vashisth Tiwari, Ruihang Lai, Jinyuan Shi, Ian En-Hsu Yen, Avner May, Tianqi Chen, and Beidi Chen.
\newblock Magicdec: Breaking the latency-throughput tradeoff for long context generation with speculative decoding.
\newblock \emph{arXiv preprint arXiv:2408.11049}, 2024.

\bibitem[Shazeer et~al.(2017)Shazeer, Mirhoseini, Maziarz, Davis, Le, Hinton, and Dean]{shazeer2017outrageously}
Noam Shazeer, Azalia Mirhoseini, Krzysztof Maziarz, Andy Davis, Quoc Le, Geoffrey Hinton, and Jeff Dean.
\newblock Outrageously large neural networks: The sparsely-gated mixture-of-experts layer.
\newblock \emph{arXiv preprint arXiv:1701.06538}, 2017.

\bibitem[Sheng et~al.(2023)Sheng, Zheng, Yuan, Li, Ryabinin, Chen, Liang, R{\'e}, Stoica, and Zhang]{sheng2023flexgen}
Ying Sheng, Lianmin Zheng, Binhang Yuan, Zhuohan Li, Max Ryabinin, Beidi Chen, Percy Liang, Christopher R{\'e}, Ion Stoica, and Ce~Zhang.
\newblock Flexgen: High-throughput generative inference of large language models with a single gpu.
\newblock In \emph{International Conference on Machine Learning}, pages 31094--31116. PMLR, 2023.

\bibitem[Snell et~al.(2024)Snell, Lee, Xu, and Kumar]{snell2024scaling}
Charlie Snell, Jaehoon Lee, Kelvin Xu, and Aviral Kumar.
\newblock Scaling llm test-time compute optimally can be more effective than scaling model parameters.
\newblock \emph{arXiv preprint arXiv:2408.03314}, 2024.

\bibitem[Snowflake-Team(2024)]{snowflake_arctic_2024}
Snowflake-Team.
\newblock Snowflake arctic.
\newblock \url{https://github.com/Snowflake-Labs/snowflake-arctic}, 2024.
\newblock Apache 2.0 License.

\bibitem[Sun et~al.(2024{\natexlab{a}})Sun, Chang, Bao, Zheng, Zheng, Liu, Dong, Chi, and Chen]{sun2024shadowkv}
Hanshi Sun, Li-Wen Chang, Wenlei Bao, Size Zheng, Ningxin Zheng, Xin Liu, Harry Dong, Yuejie Chi, and Beidi Chen.
\newblock Shadowkv: Kv cache in shadows for high-throughput long-context llm inference.
\newblock \emph{arXiv preprint arXiv:2410.21465}, 2024{\natexlab{a}}.

\bibitem[Sun et~al.(2024{\natexlab{b}})Sun, Chen, Yang, Tian, and Chen]{sun2024triforcelosslessaccelerationlong}
Hanshi Sun, Zhuoming Chen, Xinyu Yang, Yuandong Tian, and Beidi Chen.
\newblock Triforce: Lossless acceleration of long sequence generation with hierarchical speculative decoding, 2024{\natexlab{b}}.
\newblock \url{https://arxiv.org/abs/2404.11912}.

\bibitem[Sun et~al.(2024{\natexlab{c}})Sun, Haider, Zhang, Yang, Qiu, Yin, Wang, Bartlett, and Zanette]{sun2024fast}
Hanshi Sun, Momin Haider, Ruiqi Zhang, Huitao Yang, Jiahao Qiu, Ming Yin, Mengdi Wang, Peter Bartlett, and Andrea Zanette.
\newblock Fast best-of-n decoding via speculative rejection.
\newblock \emph{arXiv preprint arXiv:2410.20290}, 2024{\natexlab{c}}.

\bibitem[Sutskever(2024)]{Ilya}
Ilya Sutskever.
\newblock Sequence to sequence learning with neural networks: what a decade.
\newblock In \emph{NeurIPS (Keynote talk)}, 2024.
\newblock \url{https://www.youtube.com/watch?v=1yvBqasHLZs}.

\bibitem[Svirschevski et~al.(2024)Svirschevski, May, Chen, Chen, Jia, and Ryabinin]{svirschevski2024specexec}
Ruslan Svirschevski, Avner May, Zhuoming Chen, Beidi Chen, Zhihao Jia, and Max Ryabinin.
\newblock Specexec: Massively parallel speculative decoding for interactive llm inference on consumer devices.
\newblock \emph{Advances in Neural Information Processing Systems}, 37:\penalty0 16342--16368, 2024.

\bibitem[Tack et~al.(2025)Tack, Lanchantin, Yu, Cohen, Kulikov, Lan, Hao, Tian, Weston, and Li]{tack2025llm}
Jihoon Tack, Jack Lanchantin, Jane Yu, Andrew Cohen, Ilia Kulikov, Janice Lan, Shibo Hao, Yuandong Tian, Jason Weston, and Xian Li.
\newblock Llm pretraining with continuous concepts.
\newblock \emph{arXiv preprint arXiv:2502.08524}, 2025.

\bibitem[Tang et~al.(2024)Tang, Zhao, Zhu, Xiao, Kasikci, and Han]{tang2024quest}
Jiaming Tang, Yilong Zhao, Kan Zhu, Guangxuan Xiao, Baris Kasikci, and Song Han.
\newblock Quest: Query-aware sparsity for efficient long-context llm inference.
\newblock \emph{arXiv preprint arXiv:2406.10774}, 2024.

\bibitem[Tibshirani(1996)]{tibshirani1996regression}
Robert Tibshirani.
\newblock Regression shrinkage and selection via the lasso.
\newblock \emph{Journal of the Royal Statistical Society: Series B (Methodological)}, 58\penalty0 (1):\penalty0 267--288, 1996.

\bibitem[Tirumala and Wong(2024)]{tirumala2024nvidia}
Ajay Tirumala and Raymond Wong.
\newblock Nvidia blackwell platform: Advancing generative ai and accelerated computing.
\newblock In \emph{2024 IEEE Hot Chips 36 Symposium (HCS)}, pages 1--33. IEEE Computer Society, 2024.

\bibitem[Wang et~al.(2025)Wang, Li, Paliotta, Ritter, Rush, and Dao]{wang2025m1scalabletesttimecompute}
Junxiong Wang, Wen-Ding Li, Daniele Paliotta, Daniel Ritter, Alexander~M. Rush, and Tri Dao.
\newblock M1: Towards scalable test-time compute with mamba reasoning models, 2025.
\newblock \url{https://arxiv.org/abs/2504.10449}.

\bibitem[Wang et~al.(2022)Wang, Wei, Schuurmans, Le, Chi, Narang, Chowdhery, and Zhou]{wang2022self}
Xuezhi Wang, Jason Wei, Dale Schuurmans, Quoc Le, Ed~Chi, Sharan Narang, Aakanksha Chowdhery, and Denny Zhou.
\newblock Self-consistency improves chain of thought reasoning in language models.
\newblock \emph{arXiv preprint arXiv:2203.11171}, 2022.

\bibitem[Wei et~al.(2022)Wei, Wang, Schuurmans, Bosma, Xia, Chi, Le, Zhou, et~al.]{wei2022chain}
Jason Wei, Xuezhi Wang, Dale Schuurmans, Maarten Bosma, Fei Xia, Ed~Chi, Quoc~V Le, Denny Zhou, et~al.
\newblock Chain-of-thought prompting elicits reasoning in large language models.
\newblock \emph{Advances in neural information processing systems}, 35:\penalty0 24824--24837, 2022.

\bibitem[Wu et~al.(2024)Wu, Sun, Li, Welleck, and Yang]{wu2024inference}
Yangzhen Wu, Zhiqing Sun, Shanda Li, Sean Welleck, and Yiming Yang.
\newblock Inference scaling laws: An empirical analysis of compute-optimal inference for problem-solving with language models.
\newblock \emph{arXiv preprint arXiv:2408.00724}, 2024.

\bibitem[Xiao et~al.(2024)Xiao, Tian, Chen, Han, and Lewis]{xiao2024efficientstreaminglanguagemodels}
Guangxuan Xiao, Yuandong Tian, Beidi Chen, Song Han, and Mike Lewis.
\newblock Efficient streaming language models with attention sinks, 2024.
\newblock \url{https://arxiv.org/abs/2309.17453}.

\bibitem[Xu et~al.(2025)Xu, Xiao, Huang, Guo, and Han]{xu2025xattention}
Ruyi Xu, Guangxuan Xiao, Haofeng Huang, Junxian Guo, and Song Han.
\newblock Xattention: Block sparse attention with antidiagonal scoring.
\newblock \emph{arXiv preprint arXiv:2503.16428}, 2025.

\bibitem[Yang et~al.(2024{\natexlab{a}})Yang, Yang, Hui, Zheng, Yu, Zhou, Li, Li, Liu, Huang, Dong, Wei, Lin, Tang, Wang, Yang, Tu, Zhang, Ma, Xu, Zhou, Bai, He, Lin, Dang, Lu, Chen, Yang, Li, Xue, Ni, Zhang, Wang, Peng, Men, Gao, Lin, Wang, Bai, Tan, Zhu, Li, Liu, Ge, Deng, Zhou, Ren, Zhang, Wei, Ren, Fan, Yao, Zhang, Wan, Chu, Liu, Cui, Zhang, and Fan]{qwen2}
An~Yang, Baosong Yang, Binyuan Hui, Bo~Zheng, Bowen Yu, Chang Zhou, Chengpeng Li, Chengyuan Li, Dayiheng Liu, Fei Huang, Guanting Dong, Haoran Wei, Huan Lin, Jialong Tang, Jialin Wang, Jian Yang, Jianhong Tu, Jianwei Zhang, Jianxin Ma, Jin Xu, Jingren Zhou, Jinze Bai, Jinzheng He, Junyang Lin, Kai Dang, Keming Lu, Keqin Chen, Kexin Yang, Mei Li, Mingfeng Xue, Na~Ni, Pei Zhang, Peng Wang, Ru~Peng, Rui Men, Ruize Gao, Runji Lin, Shijie Wang, Shuai Bai, Sinan Tan, Tianhang Zhu, Tianhao Li, Tianyu Liu, Wenbin Ge, Xiaodong Deng, Xiaohuan Zhou, Xingzhang Ren, Xinyu Zhang, Xipin Wei, Xuancheng Ren, Yang Fan, Yang Yao, Yichang Zhang, Yu~Wan, Yunfei Chu, Yuqiong Liu, Zeyu Cui, Zhenru Zhang, and Zhihao Fan.
\newblock Qwen2 technical report.
\newblock \emph{arXiv preprint arXiv:2407.10671}, 2024{\natexlab{a}}.

\bibitem[Yang et~al.(2024{\natexlab{b}})Yang, Yang, Zhang, Hui, Zheng, Yu, Li, Liu, Huang, Wei, Lin, Yang, Tu, Zhang, Yang, Yang, Zhou, Lin, Dang, Lu, Bao, Yang, Yu, Li, Xue, Zhang, Zhu, Men, Lin, Li, Xia, Ren, Ren, Fan, Su, Zhang, Wan, Liu, Cui, Zhang, and Qiu]{qwen2.5}
An~Yang, Baosong Yang, Beichen Zhang, Binyuan Hui, Bo~Zheng, Bowen Yu, Chengyuan Li, Dayiheng Liu, Fei Huang, Haoran Wei, Huan Lin, Jian Yang, Jianhong Tu, Jianwei Zhang, Jianxin Yang, Jiaxi Yang, Jingren Zhou, Junyang Lin, Kai Dang, Keming Lu, Keqin Bao, Kexin Yang, Le~Yu, Mei Li, Mingfeng Xue, Pei Zhang, Qin Zhu, Rui Men, Runji Lin, Tianhao Li, Tingyu Xia, Xingzhang Ren, Xuancheng Ren, Yang Fan, Yang Su, Yichang Zhang, Yu~Wan, Yuqiong Liu, Zeyu Cui, Zhenru Zhang, and Zihan Qiu.
\newblock Qwen2.5 technical report.
\newblock \emph{arXiv preprint arXiv:2412.15115}, 2024{\natexlab{b}}.

\bibitem[Yang et~al.(2025)Yang, Li, Yang, Zhang, Hui, Zheng, Yu, Gao, Huang, Lv, Zheng, Liu, Zhou, Huang, Hu, Ge, Wei, Lin, Tang, Yang, Tu, Zhang, Yang, Yang, Zhou, Zhou, Lin, Dang, Bao, Yang, Yu, Deng, Li, Xue, Li, Zhang, Wang, Zhu, Men, Gao, Liu, Luo, Li, Tang, Yin, Ren, Wang, Zhang, Ren, Fan, Su, Zhang, Zhang, Wan, Liu, Wang, Cui, Zhang, Zhou, and Qiu]{yang2025qwen3technicalreport}
An~Yang, Anfeng Li, Baosong Yang, Beichen Zhang, Binyuan Hui, Bo~Zheng, Bowen Yu, Chang Gao, Chengen Huang, Chenxu Lv, Chujie Zheng, Dayiheng Liu, Fan Zhou, Fei Huang, Feng Hu, Hao Ge, Haoran Wei, Huan Lin, Jialong Tang, Jian Yang, Jianhong Tu, Jianwei Zhang, Jianxin Yang, Jiaxi Yang, Jing Zhou, Jingren Zhou, Junyang Lin, Kai Dang, Keqin Bao, Kexin Yang, Le~Yu, Lianghao Deng, Mei Li, Mingfeng Xue, Mingze Li, Pei Zhang, Peng Wang, Qin Zhu, Rui Men, Ruize Gao, Shixuan Liu, Shuang Luo, Tianhao Li, Tianyi Tang, Wenbiao Yin, Xingzhang Ren, Xinyu Wang, Xinyu Zhang, Xuancheng Ren, Yang Fan, Yang Su, Yichang Zhang, Yinger Zhang, Yu~Wan, Yuqiong Liu, Zekun Wang, Zeyu Cui, Zhenru Zhang, Zhipeng Zhou, and Zihan Qiu.
\newblock Qwen3 technical report, 2025.
\newblock \url{https://arxiv.org/abs/2505.09388}.

\bibitem[Yao et~al.(2023{\natexlab{a}})Yao, Zhao, Yu, et~al.]{yao2023react}
Shinn Yao, Jiaming Zhao, Dian Yu, et~al.
\newblock React: Synergizing reasoning and acting in language models.
\newblock \emph{Advances in Neural Information Processing Systems (NeurIPS)}, 2023{\natexlab{a}}.

\bibitem[Yao et~al.(2023{\natexlab{b}})Yao, Yu, Zhao, Shafran, Griffiths, Cao, and Narasimhan]{yao2023tree}
Shunyu Yao, Dian Yu, Jeffrey Zhao, Izhak Shafran, Tom Griffiths, Yuan Cao, and Karthik Narasimhan.
\newblock Tree of thoughts: Deliberate problem solving with large language models.
\newblock \emph{Advances in neural information processing systems}, 36:\penalty0 11809--11822, 2023{\natexlab{b}}.

\bibitem[Ye et~al.(2025)Ye, Chen, Lai, Lin, Zhang, Wang, Chen, Kasikci, Grover, Krishnamurthy, and Ceze]{ye2025flashinfer}
Zihao Ye, Lequn Chen, Ruihang Lai, Wuwei Lin, Yineng Zhang, Stephanie Wang, Tianqi Chen, Baris Kasikci, Vinod Grover, Arvind Krishnamurthy, and Luis Ceze.
\newblock Flashinfer: Efficient and customizable attention engine for llm inference serving.
\newblock \emph{arXiv preprint arXiv:2501.01005}, 2025.
\newblock \url{https://arxiv.org/abs/2501.01005}.

\bibitem[Yu et~al.(2022)Yu, Jeong, Kim, Kim, and Chun]{280922}
Gyeong-In Yu, Joo~Seong Jeong, Geon-Woo Kim, Soojeong Kim, and Byung-Gon Chun.
\newblock Orca: A distributed serving system for {Transformer-Based} generative models.
\newblock In \emph{16th USENIX Symposium on Operating Systems Design and Implementation (OSDI 22)}, pages 521--538, Carlsbad, CA, July 2022. USENIX Association.
\newblock ISBN 978-1-939133-28-1.
\newblock \url{https://www.usenix.org/conference/osdi22/presentation/yu}.

\bibitem[Yuan et~al.(2025)Yuan, Gao, Dai, Luo, Zhao, Zhang, Xie, Wei, Wang, Xiao, et~al.]{yuan2025native}
Jingyang Yuan, Huazuo Gao, Damai Dai, Junyu Luo, Liang Zhao, Zhengyan Zhang, Zhenda Xie, YX~Wei, Lean Wang, Zhiping Xiao, et~al.
\newblock Native sparse attention: Hardware-aligned and natively trainable sparse attention.
\newblock \emph{arXiv preprint arXiv:2502.11089}, 2025.

\bibitem[Yuan et~al.(2024{\natexlab{a}})Yuan, Shang, Zhou, Dong, Zhou, Xue, Wu, Li, Gu, Lee, Yan, Chen, Sun, and Keutzer]{yuan2024llminferenceunveiledsurvey}
Zhihang Yuan, Yuzhang Shang, Yang Zhou, Zhen Dong, Zhe Zhou, Chenhao Xue, Bingzhe Wu, Zhikai Li, Qingyi Gu, Yong~Jae Lee, Yan Yan, Beidi Chen, Guangyu Sun, and Kurt Keutzer.
\newblock Llm inference unveiled: Survey and roofline model insights, 2024{\natexlab{a}}.
\newblock \url{https://arxiv.org/abs/2402.16363}.

\bibitem[Yuan et~al.(2024{\natexlab{b}})Yuan, Shang, Zhou, Dong, Zhou, Xue, Wu, Li, Gu, Lee, et~al.]{yuan2024llm}
Zhihang Yuan, Yuzhang Shang, Yang Zhou, Zhen Dong, Zhe Zhou, Chenhao Xue, Bingzhe Wu, Zhikai Li, Qingyi Gu, Yong~Jae Lee, et~al.
\newblock Llm inference unveiled: Survey and roofline model insights.
\newblock \emph{arXiv preprint arXiv:2402.16363}, 2024{\natexlab{b}}.

\bibitem[Zaheer et~al.(2020)Zaheer, Guruganesh, Dubey, Ainslie, Alberti, Ontanon, Pham, Ravula, Wang, Yang, et~al.]{zaheer2020bigbird}
Manzil Zaheer, Guru Guruganesh, Kumar~Avinava Dubey, Joshua Ainslie, Chris Alberti, Santiago Ontanon, Philip Pham, Anirudh Ravula, Qifan Wang, Li~Yang, et~al.
\newblock Big bird: Transformers for longer sequences.
\newblock \emph{Advances in Neural Information Processing Systems}, 33, 2020.

\bibitem[Zandieh et~al.(2023)Zandieh, Han, Daliri, and Karbasi]{zandieh2023kdeformer}
Amir Zandieh, Insu Han, Majid Daliri, and Amin Karbasi.
\newblock Kdeformer: Accelerating transformers via kernel density estimation.
\newblock In \emph{International Conference on Machine Learning}, pages 40605--40623. PMLR, 2023.

\bibitem[Zhang et~al.(2022)Zhang, Roller, Goyal, Artetxe, Chen, Chen, Dewan, Diab, Li, Lin, et~al.]{zhang2022opt}
Susan Zhang, Stephen Roller, Naman Goyal, Mikel Artetxe, Moya Chen, Shuohui Chen, Christopher Dewan, Mona Diab, Xian Li, Xi~Victoria Lin, et~al.
\newblock Opt: Open pre-trained transformer language models.
\newblock \emph{arXiv preprint arXiv:2205.01068}, 2022.

\bibitem[Zhang et~al.(2023)Zhang, Sheng, Zhou, Chen, Zheng, Cai, Song, Tian, R{\'e}, Barrett, et~al.]{zhang2023h2o}
Zhenyu Zhang, Ying Sheng, Tianyi Zhou, Tianlong Chen, Lianmin Zheng, Ruisi Cai, Zhao Song, Yuandong Tian, Christopher R{\'e}, Clark Barrett, et~al.
\newblock H2o: Heavy-hitter oracle for efficient generative inference of large language models.
\newblock \emph{Advances in Neural Information Processing Systems}, 36:\penalty0 34661--34710, 2023.

\bibitem[Zheng et~al.(2024)Zheng, Yin, Xie, Sun, Huang, Yu, Cao, Kozyrakis, Stoica, Gonzalez, et~al.]{zheng2024sglang}
Lianmin Zheng, Liangsheng Yin, Zhiqiang Xie, Chuyue~Livia Sun, Jeff Huang, Cody~Hao Yu, Shiyi Cao, Christos Kozyrakis, Ion Stoica, Joseph~E Gonzalez, et~al.
\newblock Sglang: Efficient execution of structured language model programs.
\newblock \emph{Advances in Neural Information Processing Systems}, 37:\penalty0 62557--62583, 2024.

\bibitem[Zhu et~al.(2024)Zhu, Zhao, Zhao, Zuo, Gu, Xie, Gao, Xu, Tang, Ye, et~al.]{zhu2024nanoflow}
Kan Zhu, Yilong Zhao, Liangyu Zhao, Gefei Zuo, Yile Gu, Dedong Xie, Yufei Gao, Qinyu Xu, Tian Tang, Zihao Ye, et~al.
\newblock Nanoflow: Towards optimal large language model serving throughput.
\newblock \emph{arXiv preprint arXiv:2408.12757}, 2024.

\end{thebibliography}
\clearpage
\newpage
\beginappendix

    
    
    
    

\section*{Table of Contents}
\vspace{-0.5em}
\begin{appendixtocfont}
\begin{itemize}[left=1.5em, label=--, itemsep=0.5em, topsep=0.2em]
    \item \textbf{\hyperref[costmodel]{\nameref{costmodel}}} \dotfill \pageref{costmodel}
    \begin{itemize}[left=2.5em, label=•]
        \item \hyperref[maxvsadditive]{\nameref{maxvsadditive}} \dotfill \pageref{maxvsadditive}
        \item \hyperref[sparseattentioncost]{\nameref{sparseattentioncost}} \dotfill \pageref{sparseattentioncost}
    \end{itemize}
    
    \item \textbf{\hyperref[densescalinglaw]{\nameref{densescalinglaw}}} \dotfill \pageref{densescalinglaw}
    \begin{itemize}[left=2.5em, label=•]
        \item \hyperref[densebenchmarks]{\nameref{densebenchmarks}} \dotfill \pageref{densebenchmarks}
        \item \hyperref[denseaddmodels]{\nameref{denseaddmodels}} \dotfill \pageref{denseaddmodels}
    \end{itemize}
    
    \item \textbf{\hyperref[sparsescalinglaw]{\nameref{sparsescalinglaw}}} \dotfill \pageref{sparsescalinglaw}
    \begin{itemize}[left=2.5em, label=•]
        \item \hyperref[sparsebenchmarks]{\nameref{sparsebenchmarks}} \dotfill \pageref{sparsebenchmarks}
        \item \hyperref[sparseanalysis]{\nameref{sparseanalysis}} \dotfill \pageref{sparseanalysis}
    \end{itemize}
    
    \item \textbf{\hyperref[expdetails]{\nameref{expdetails}}} \dotfill \pageref{expdetails}
    \begin{itemize}[left=2.5em, label=•]
        \item \hyperref[costaccuracysolving]{\nameref{costaccuracysolving}} \dotfill \pageref{costaccuracysolving}
        \item \hyperref[optimalresourceallocation]{\nameref{optimalresourceallocation}} \dotfill \pageref{optimalresourceallocation}
        \item \hyperref[topkandblocktopk]{\nameref{topkandblocktopk}} \dotfill \pageref{topkandblocktopk}
    \end{itemize}
    
    \item \textbf{\hyperref[relatedwork]{\nameref{relatedwork}}} \dotfill \pageref{relatedwork}
\end{itemize}
\end{appendixtocfont}

\section{Cost Model}
\label{costmodel}
In this section, we delve into the cost models used in the \law. We show empirically that adopting a max cost model does not alter the scaling behavior and outline methods for calculating the cost of sparse attention models.
\begin{figure*}[h]
    \centering
    \includegraphics[width=\linewidth]{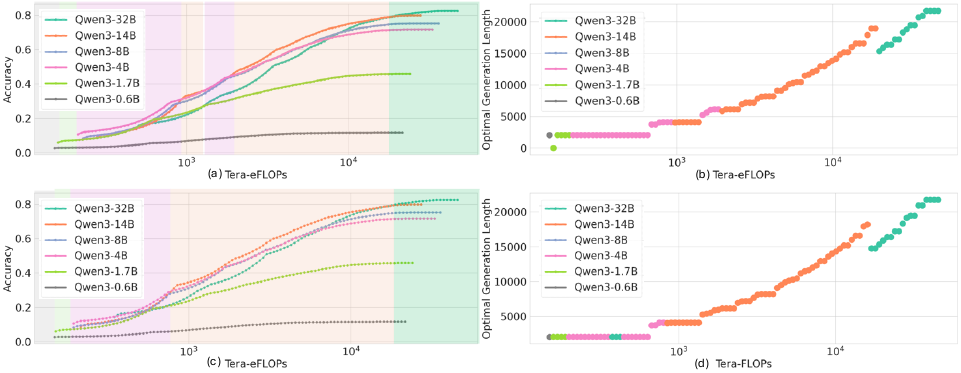} 
    \caption{\textbf{AIME24 Pareto Frontier (\longcot) with Max Cost Models.} \textbf{(a)(b)} is the original plot with the additive cost model. \textbf{(c)(d)} is the corresponding plot using max cost models. Compared to the original plots, the overall trend is similar except that larger models span a slightly broader region on the Pareto frontier. For example, the 14B model now consistently outperforms the 4B model with a noticeable gap around accuracy 0.3 and maintains dominance thereafter. In contrast, under the additive cost model in~\Cref{fig:modelcost}\textbf{(a)}, the two models alternate in performance until accuracy exceeds 0.4. This suggests that, when evaluated using a max cost model, larger models appear slightly more efficient relative to their performance under additive cost models.}
    \label{rooflinemodel}
\end{figure*}
\subsection{Max Cost Model v.s. Additive Cost Model}
\label{maxvsadditive}
Max cost model is widely used in performance modeling~\citep{yuan2024llm}. 
It assumes that computation and memory operations can be fully overlapped with each other and only considers the bottleneck operation for cost measurement.
\[
C_{\text{max-cost}} = \max(C_{\text{comp}},\, C_{\text{mem}} \times I)
\]
where \(C_{\text{comp}}\) denotes the compute cost, \(C_{\text{mem}}\) the memory cost per access, and \(I\) the memory intensity. 

In this section, we analyze the \law using the max cost model. For clarity, we refer to the cost model \(C_{\text{comp}} + C_{\text{mem}} \times I\), which is used in the main paper, as \textbf{the additive cost model}.

\begin{figure*}
    \centering
    \subfloat[\longcot]{
\includegraphics[width=0.46\linewidth]{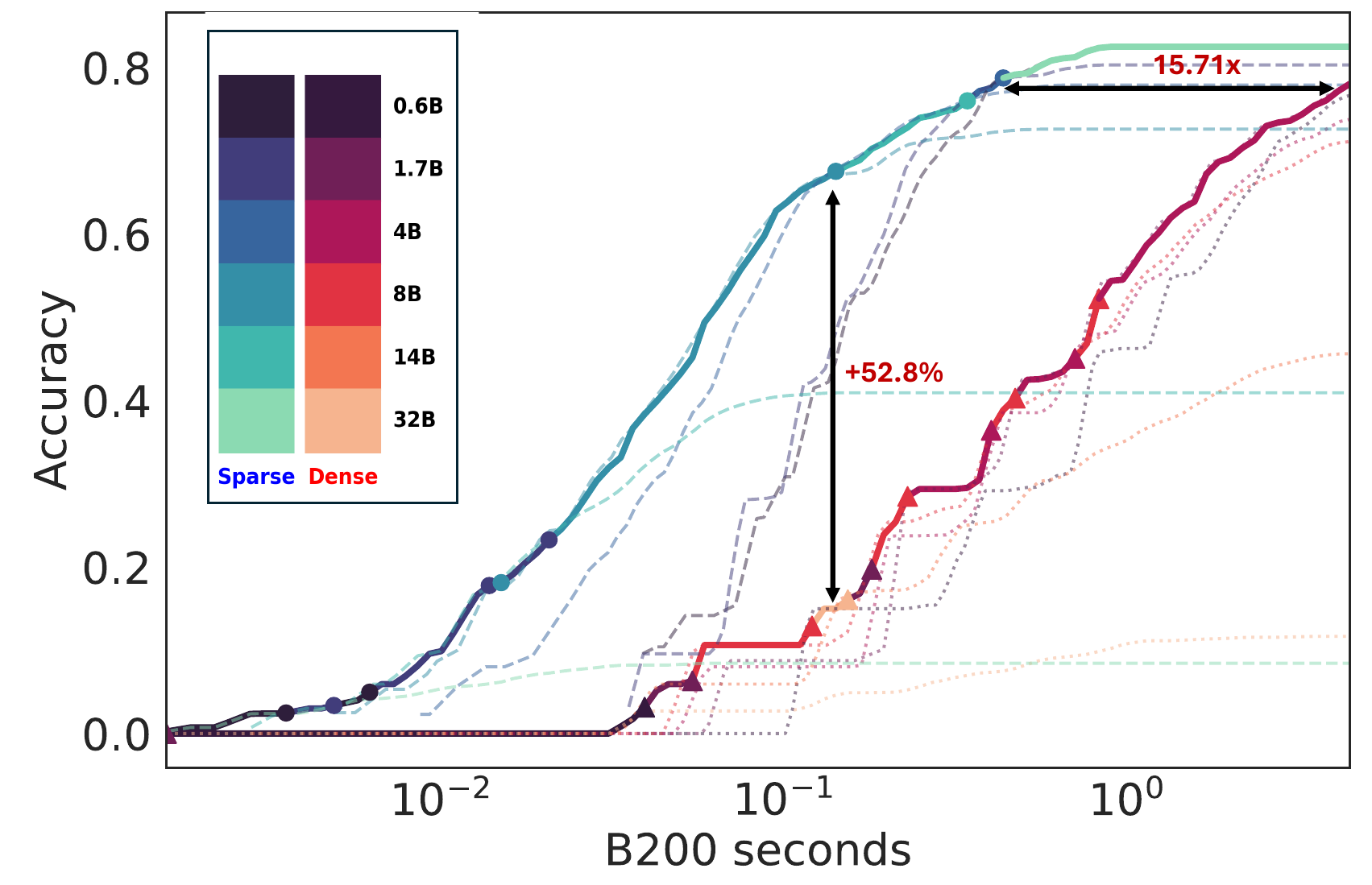}
    \label{fig: topk-cot-roofline}
    }
  \subfloat[\bon]{
\includegraphics[width=0.46\linewidth]{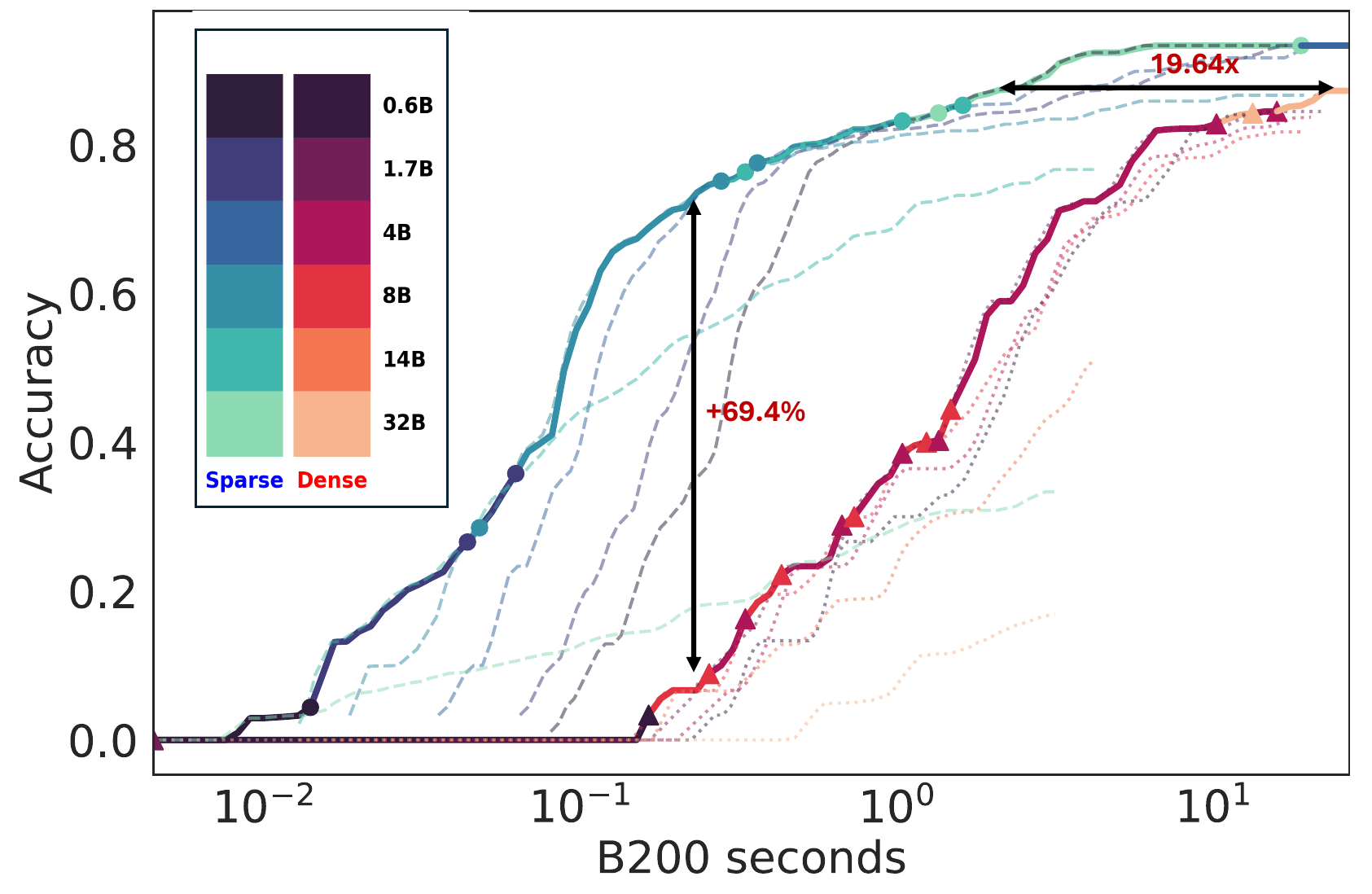}
    \label{fig: topk-bon-roofline} 
    } 
    \caption{\textbf{Sparse attention scales significantly better under max cost models.} We re-plot~\Cref{fig: cot-total,fig: bon-total} using max cost models. Compared to the original plots, the performance and efficiency gaps between sparse attention models and dense models become more pronounced. In \longcot, the accuracy and efficiency gaps increase from $47.5$ points and $11.21\times$ to $52.8$ points and $15.71\times$, respectively. In \bon, the gaps widen from $65$ points and $10.67\times$ to $69.4$ points and $19.64\times$.}
\end{figure*}

We draw two conclusions from empirical results \textbf{under the max cost model}:
\begin{figure*}
    \centering
\includegraphics[width=0.4\linewidth]{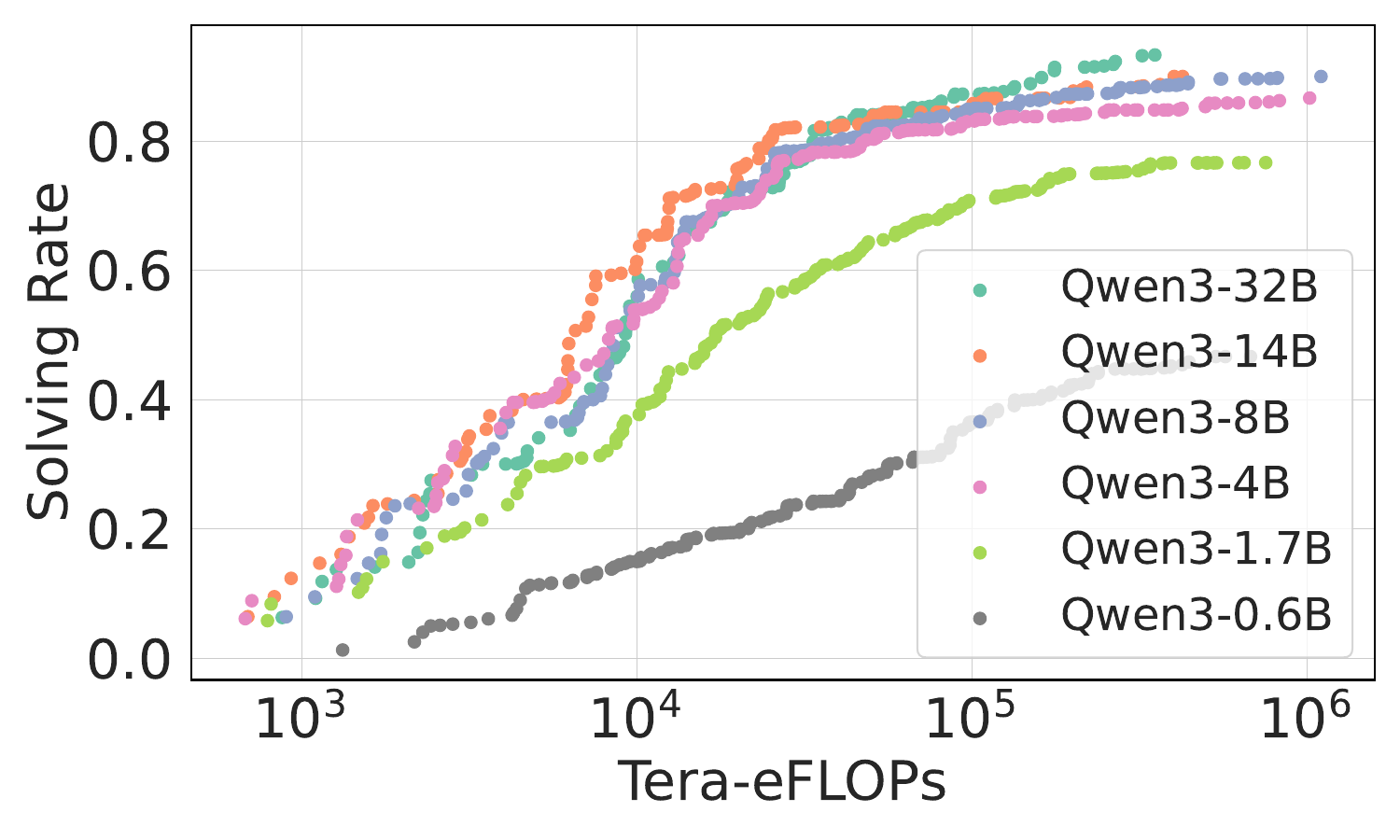}
    \caption{\textbf{AIME24 Pareto Frontier (\bon) with Max Cost Models.} We re-plot~\Cref{fig: bon-acc-eflops} using max cost models. The Pareto Frontier is very similar under different cost models.}
    \label{fig: bon-roofline}
\end{figure*}
\begin{itemize}[itemsep=0.5pt, topsep=2pt, leftmargin=*]
    \item \textbf{\law for dense models still holds.} We re-plot~\Cref{fig:modelcost}\textbf{(a)(b)} and~\Cref{fig: bon-acc-eflops} under the measurement of max cost models in~\Cref{rooflinemodel,fig: bon-roofline}.  We find except that in \longcot scenarios, large models become slightly more effective in low-cost regime (with accuracy$\sim$0.3), the overall trends are very close to the plots with additive cost models.
    \item  \textbf{Sparse attention solves problems more cost-effectively.} We re-plot~\Cref{fig: cot-total,fig: bon-total} in~\Cref{fig: topk-cot-roofline,fig: topk-bon-roofline}. Under the max cost models, in \longcot, the accuracy and efficiency gaps increase from $47.5$ points and $11.21\times$ to $52.8$ points and $15.71\times$, respectively. In \bon, the gaps widen from $65$ points and $10.67\times$ to $69.4$ points and $19.64\times$. These results indicate that under the max cost model, our claim that sparse attention can enhance problem-solving performance is strengthen. Compared to dense attention models, sparse attention models tend to have more balanced memory and compute costs. Thus omitting one of them via a max cost model will favor sparse attention models.
\end{itemize}

\subsection{Details about Sparse Attention Cost Model}
\label{sparseattentioncost}
Sparse attention models follow different cost functions due to the sparsification of KV memory access. In this paper, we focus on algorithms that impose a uniform KV budget (denoted as \( B \)) per attention head for each decoded token. We consider $L_{in} \geq B$ for the sake of simplicity. Under this setting, the cost model for sparse attention is given by:

\begin{equation}
  C_{\text{sparse}} = \underbrace{2NPL_{\text{out}} + 2rNDBL_{\text{out}}}_{\text{compute}} + \underbrace{ 2INDBL_{\text{out}}}_{\text{memory}}.
  \label{eq: sparse}
\end{equation}

In practical implementations, we must also account for the overhead associated with retrieving or searching KV memory, denoted as \( C_{\text{search}} \), which depends on the specific sparse attention algorithm \( \mathcal{A} \). For example, in block top-\(k\) selection, the search cost is:
\begin{equation}
  C_{\text{search}} = \underbrace{\frac{2NL_{\text{in}}DL_{\text{out}} + rNDL_{\text{out}}^2}{2\text{Block-Size}}}_{\text{compute}} + \underbrace{\frac{2IL_{\text{in}}DL_{\text{out}} + INDL_{\text{out}}^2}{2\text{Block-Size}}}_{\text{memory}}.  
  \label{eq: search}
\end{equation}

In our work, we choose the Block-Size in such a way that $C_{\text{sparse}}$ and $C_{\text{search}}$ are roughly balanced, so that the sparse attention cost increases sub-linearly with generation length.

For local attention and oracle top-$k$ attention, we assume no search overhead, i.e., \( C_{\text{search}} = 0 \).

Many sparse attention algorithms skip the first layer~\citep{tang2024quest,chen2024magicpig,zhang2023h2o}, resulting in only a minor increase in total cost. For the Qwen3 series, this additional overhead is bounded by \(3.57\%\) for the 0.6B model and by \(1.56\%\) for the 32B model.

\section{\law}
\label{densescalinglaw}
In this section, we further verify \law for dense models proposed in~\Cref{sec:analysis} with extended experimental results of different benchmarks and model series.
\subsection{Additional Benchmarks}
\label{densebenchmarks}
\begin{figure*}
    \centering \includegraphics[width=\linewidth]{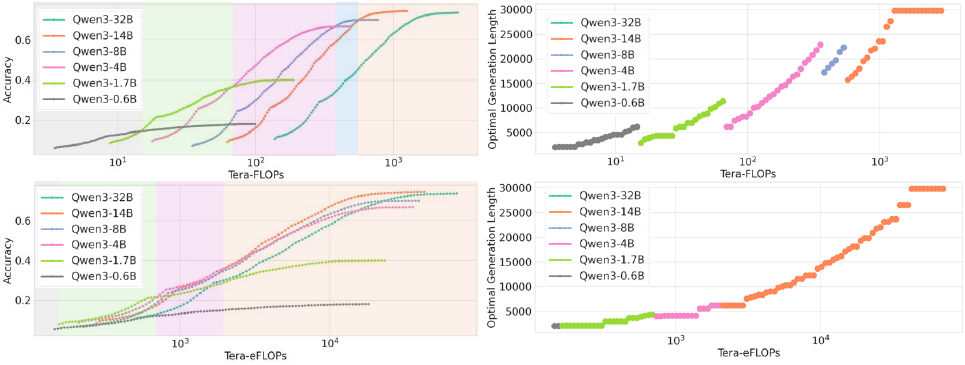}
    \caption{\textbf{AIME25 Pareto Frontier (\longcot).} We conduct the same experiments as~\Cref{fig:modelcost}.}
    \label{fig:aime25qwen3}
\end{figure*}

\begin{figure*}
    \centering
    \subfloat[Accuracy (eFLOPs)]{
\includegraphics[width=0.32\linewidth]{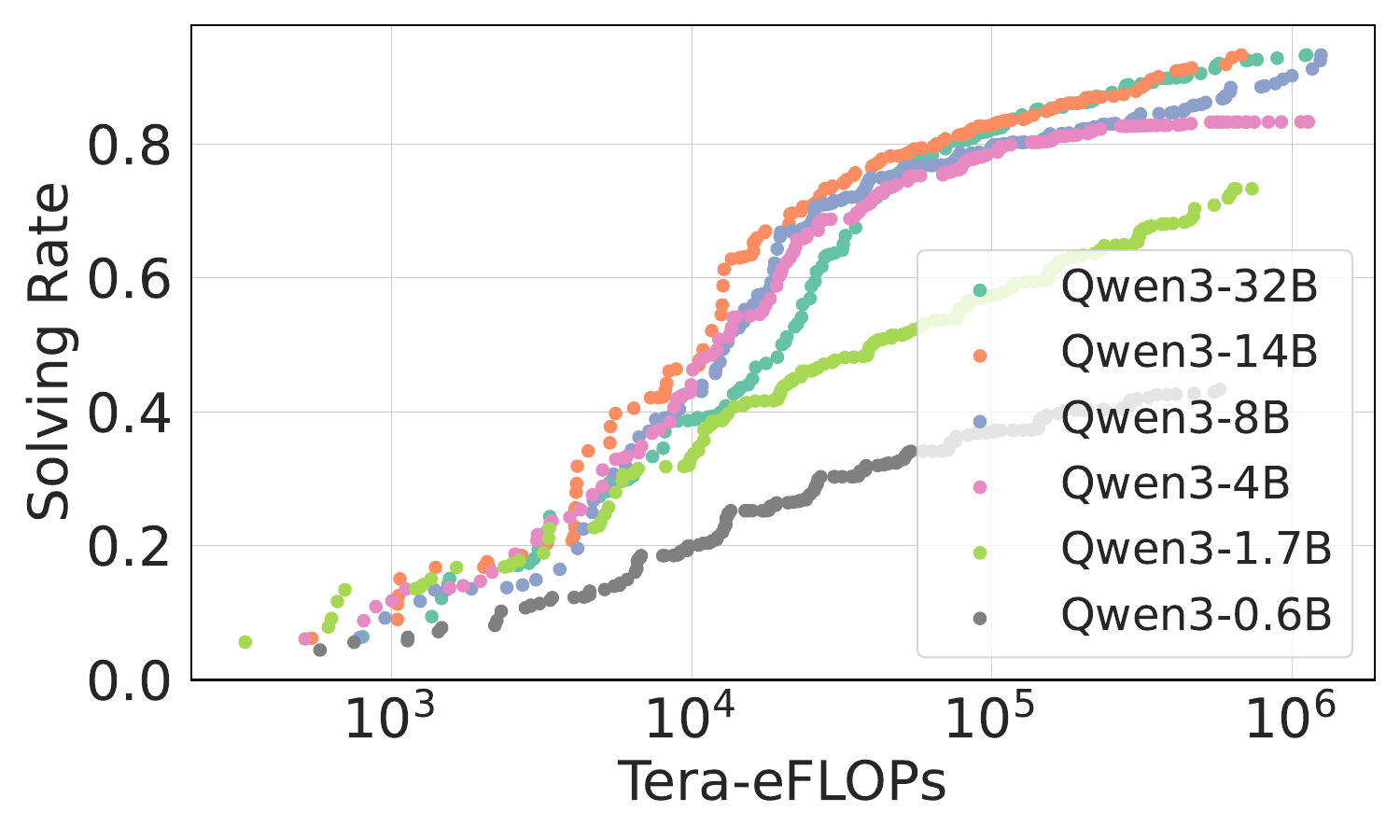}
    \label{fig: bon-acc-eflops-aime25}
    }
  \subfloat[Accuracy (FLOPs)]{
\includegraphics[width=0.32\linewidth]{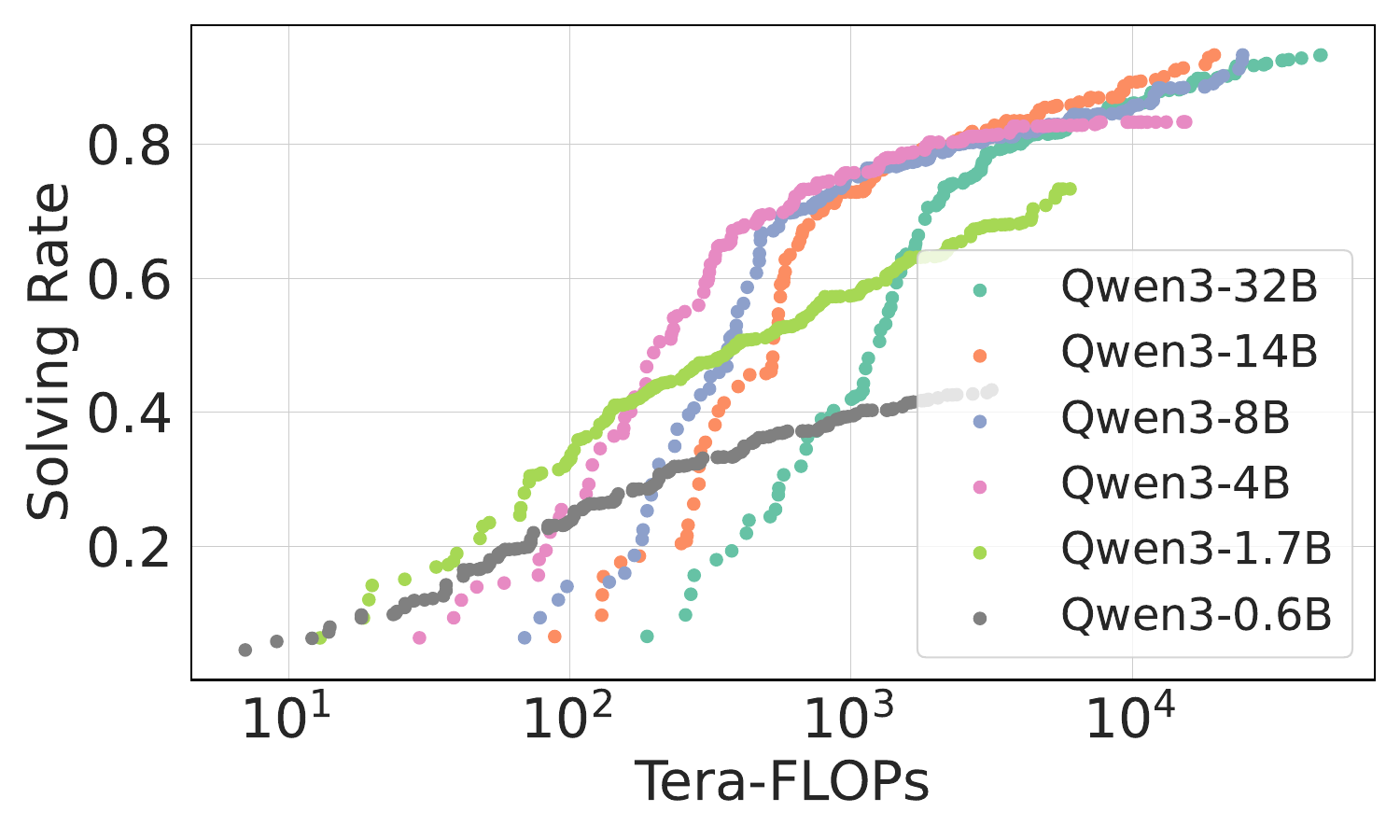}
    \label{fig: bon-acc-flops-aime25} 
    }
    \subfloat[Optimal Models]{
\includegraphics[width=0.32\linewidth]{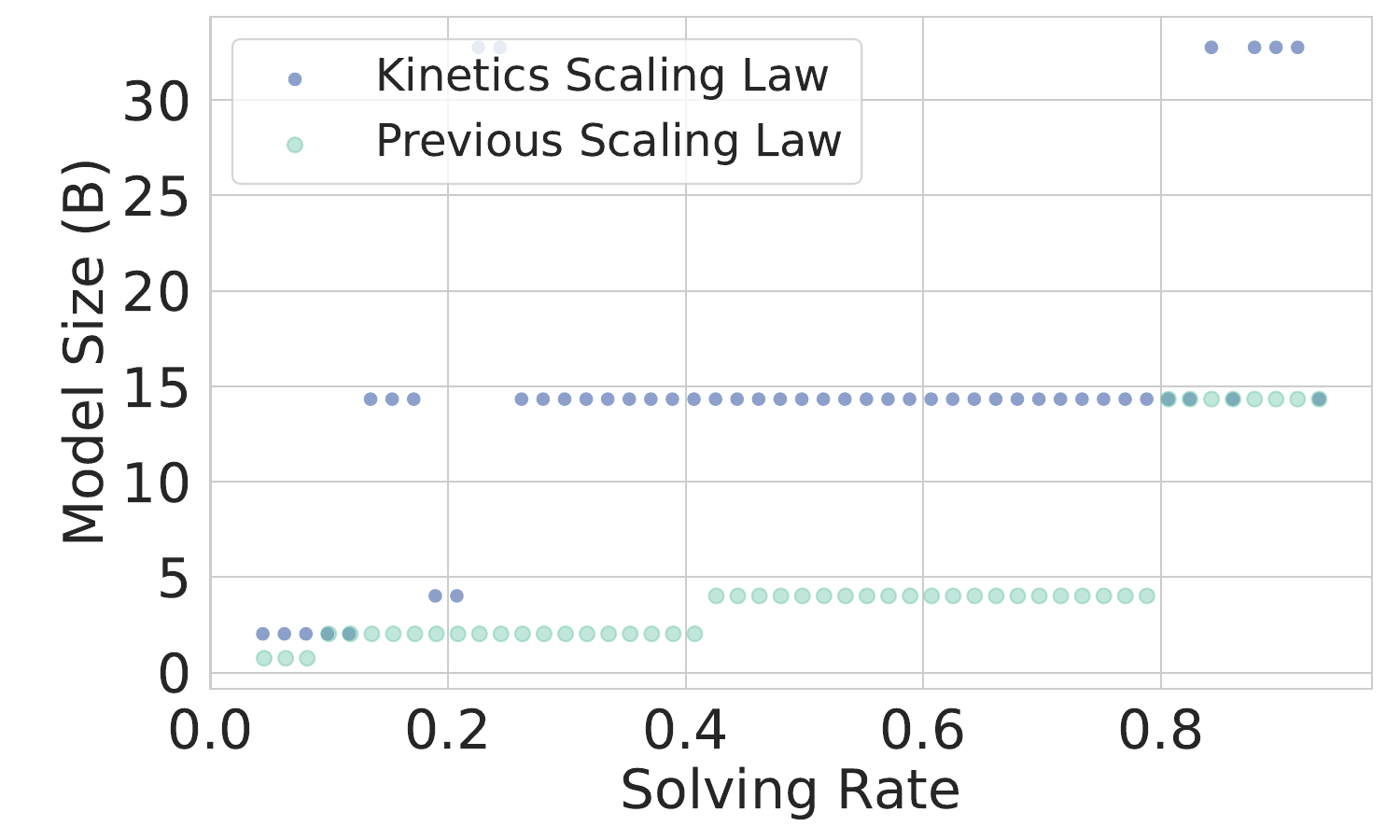}
    \label{fig: bon-model-selection-aime25} 
    } 
    \caption{\textbf{AIME25 Pareto Frontier (\bon).} We conduct the same experiments as~\Cref{fig: bon-acc-eflops,fig: bon-model-selection,fig: bon-acc-flops}. 
    }
\end{figure*}

\begin{figure*}
    \centering
    \includegraphics[width=\linewidth]{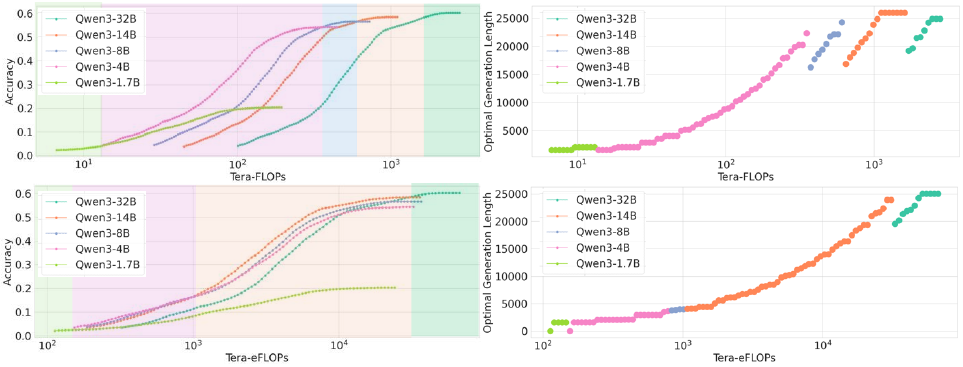}
    \caption{\textbf{LiveCodeBench Pareto Frontier (\longcot).} We conduct the same experiments as~\Cref{fig:modelcost}.}
    \label{fig:lcbqwen3}
\end{figure*}

\begin{figure*}
    \centering
    \subfloat[Accuracy (eFLOPs)]{
\includegraphics[width=0.32\linewidth]{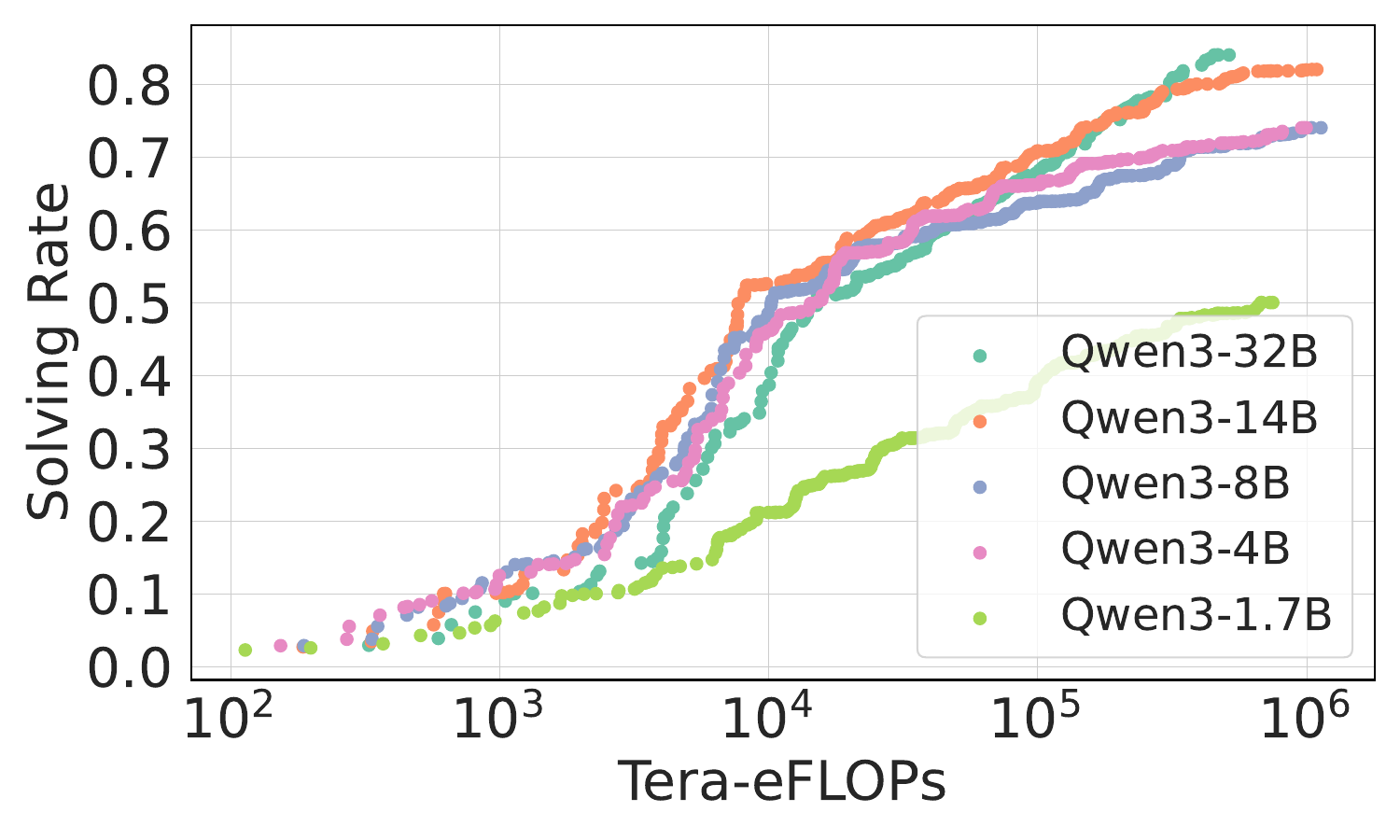}
    \label{fig: bon-acc-eflops-lcb}
    }
  \subfloat[Accuracy (FLOPs)]{
\includegraphics[width=0.32\linewidth]{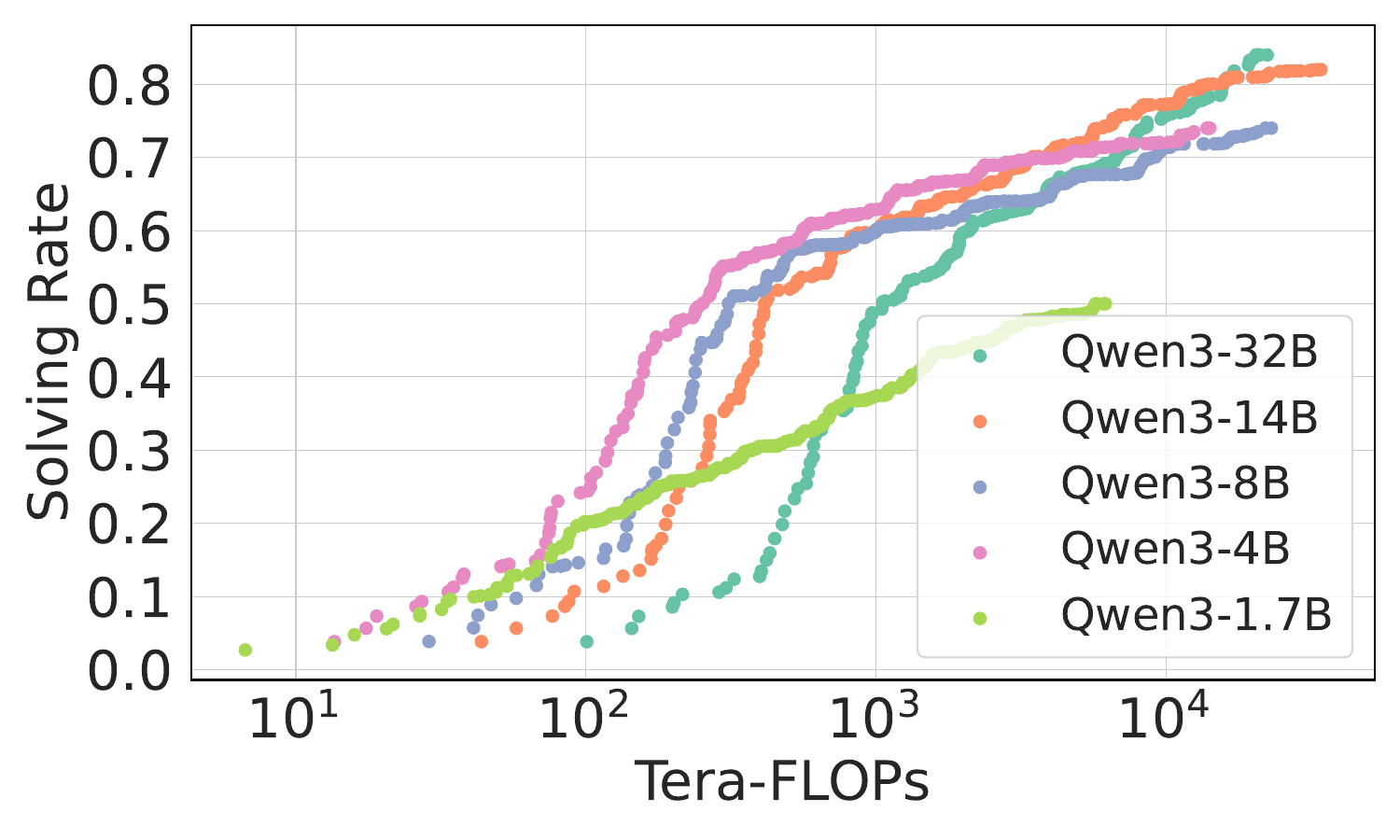}
    \label{fig: bon-acc-flops-lcb} 
    }
    \subfloat[Optimal Models]{
\includegraphics[width=0.32\linewidth]{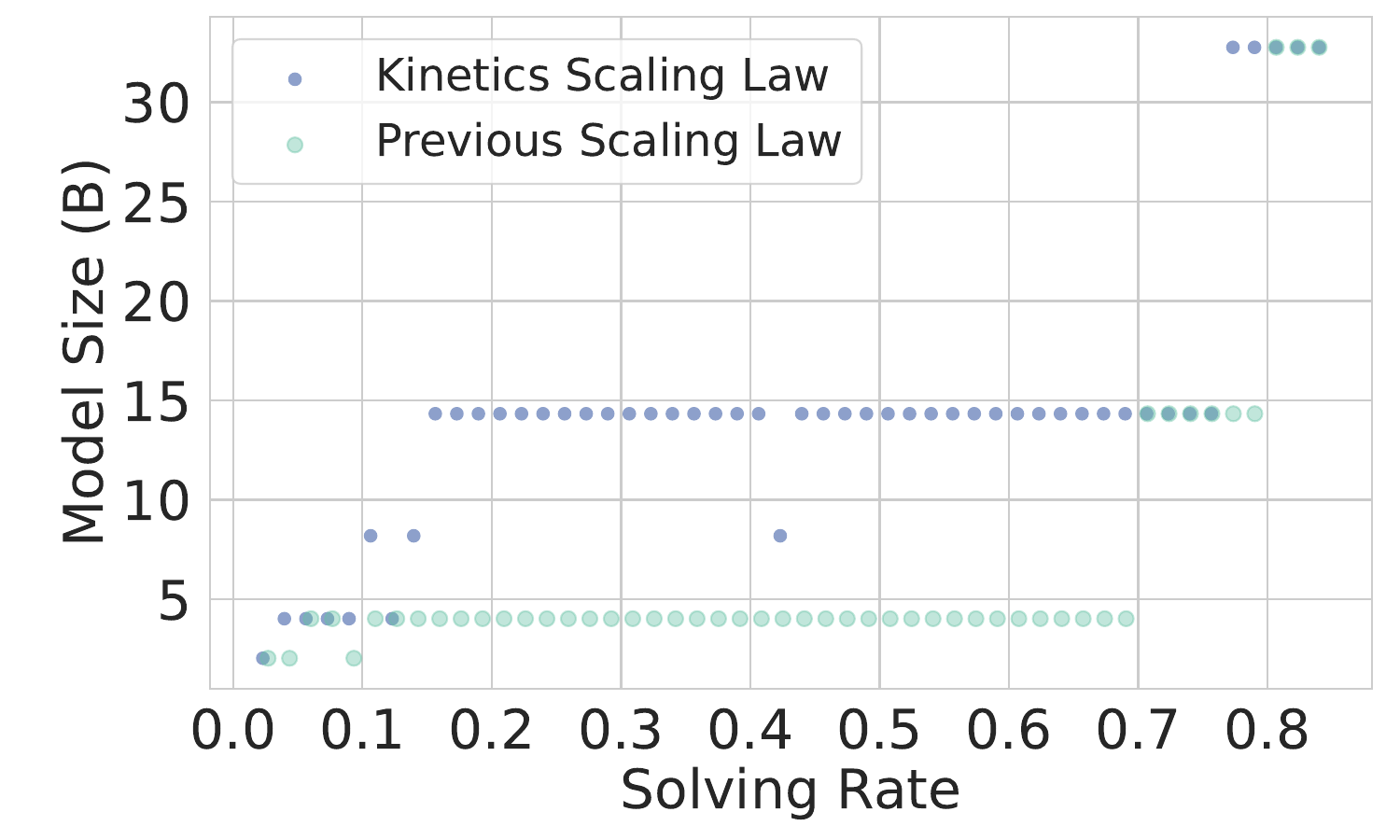}
    \label{fig: bon-model-selection-lcb} 
    } 
    \caption{\textbf{LiveCodeBench Pareto Frontier (\bon).} We conduct the same experiments as~\Cref{fig: bon-acc-eflops,fig: bon-model-selection,fig: bon-acc-flops}. 
    }
\end{figure*}

We evaluate on AIME25 in~\Cref{fig:aime25qwen3,fig: bon-acc-flops-aime25,fig: bon-acc-eflops-aime25,fig: bon-model-selection-aime25} and LiveCodeBench\footnote{For LiveCodeBench dataset, we have sampled 50 examples from the \textit{v5} subset consisting 167 examples. Our subset comprises 24 hard, 16 medium and 10 easy examples respectively.}in~\Cref{fig:lcbqwen3,fig: bon-acc-flops-lcb,fig: bon-acc-eflops-lcb,fig: bon-model-selection-lcb} (excluding the 0.6B model), following the setting described in~\Cref{sec:analysis}. The empirical results support the \law: across both benchmarks, the 0.6B and 1.7B models are consistently less effective, and the Pareto frontier is almost always dominated by the 14B models.


\subsection{Additional Reasoning Models}
\label{denseaddmodels}

\begin{figure*}
    \centering
    \includegraphics[width=\linewidth]{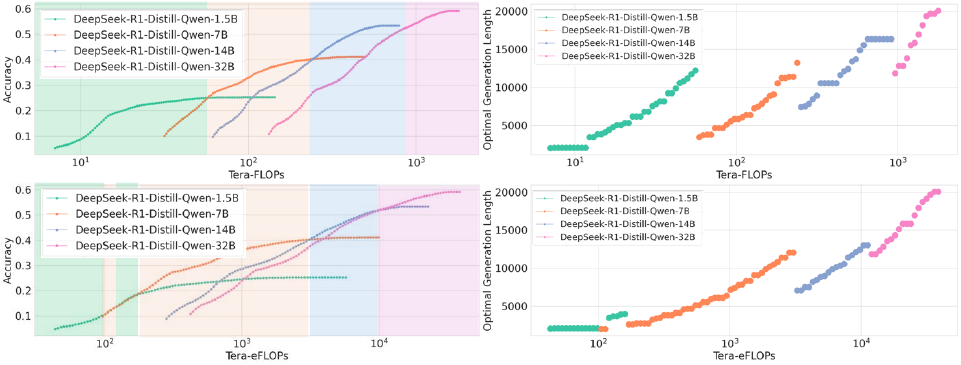}
    \caption{\textbf{AIME24 Pareto Frontier (\longcot).} We conduct the same experiments as~\Cref{fig:modelcost} on DeepSeek Distilled Qwen series.}
    \label{fig:aime24ds}
\end{figure*}
\begin{figure*}
    \centering
    \subfloat[Accuracy (eFLOPs)]{
\includegraphics[width=0.32\linewidth]{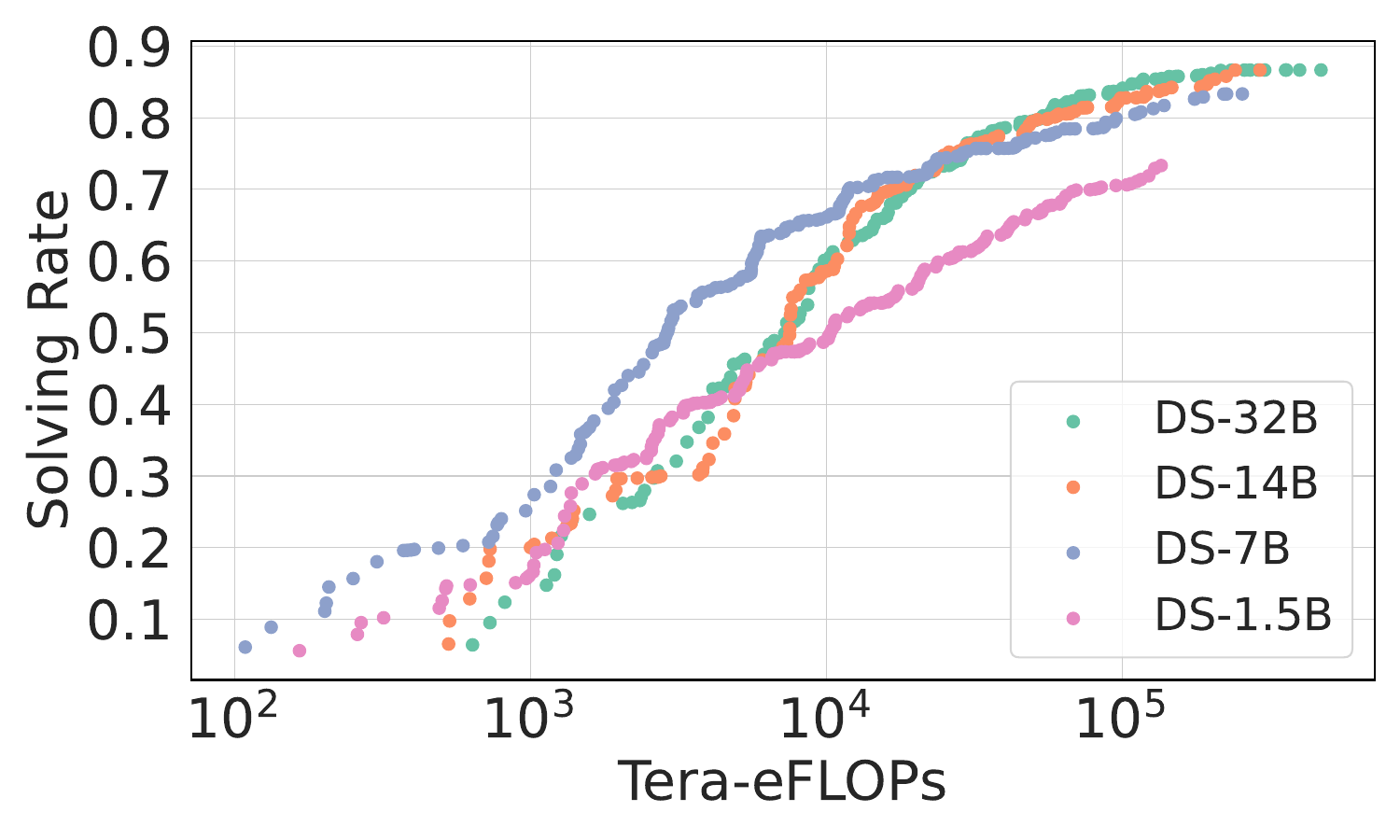}
    \label{fig: bon-acc-eflops-ds}
    }
  \subfloat[Accuracy (FLOPs)]{
\includegraphics[width=0.32\linewidth]{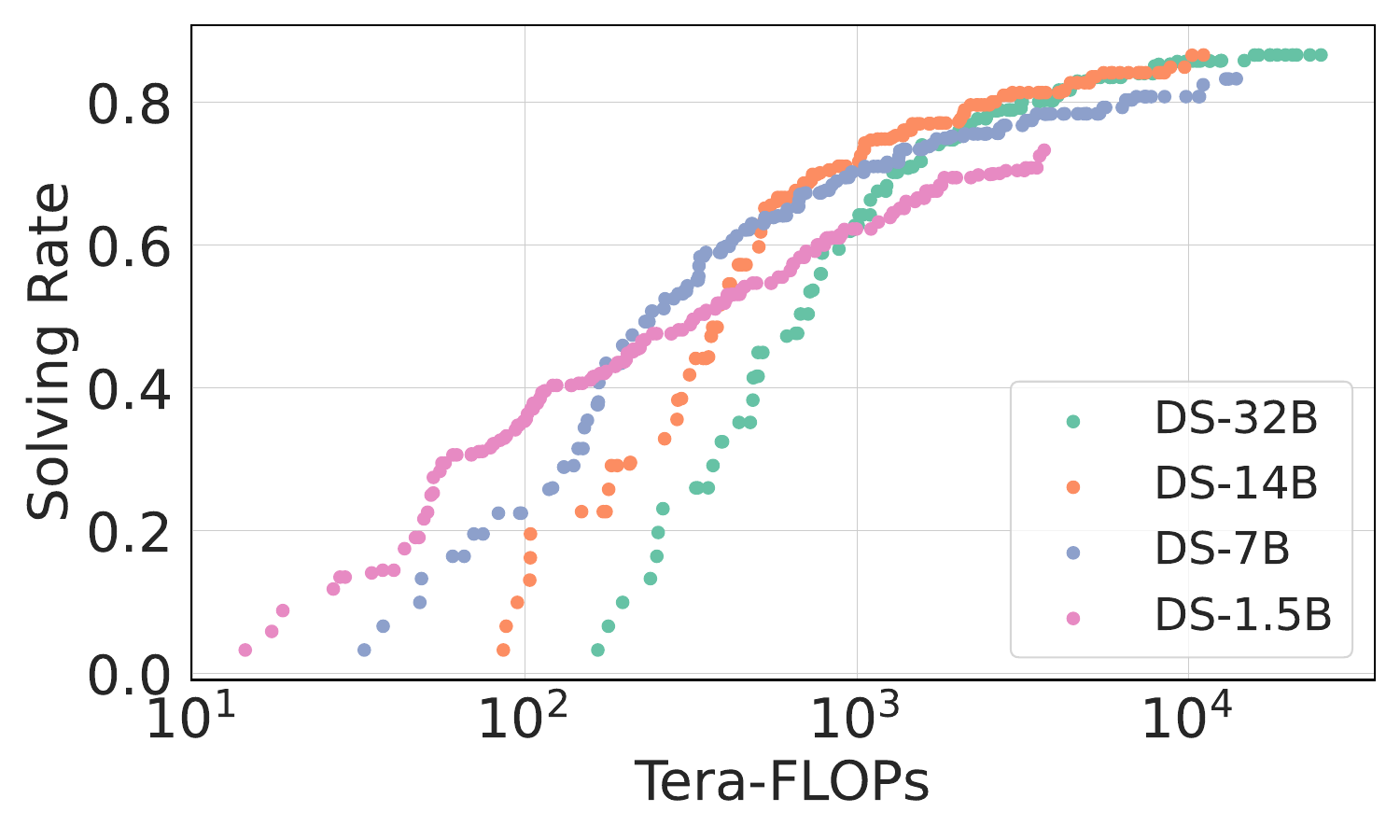}
    \label{fig: bon-acc-flops-ds} 
    }
    \subfloat[Optimal Models]{
\includegraphics[width=0.32\linewidth]{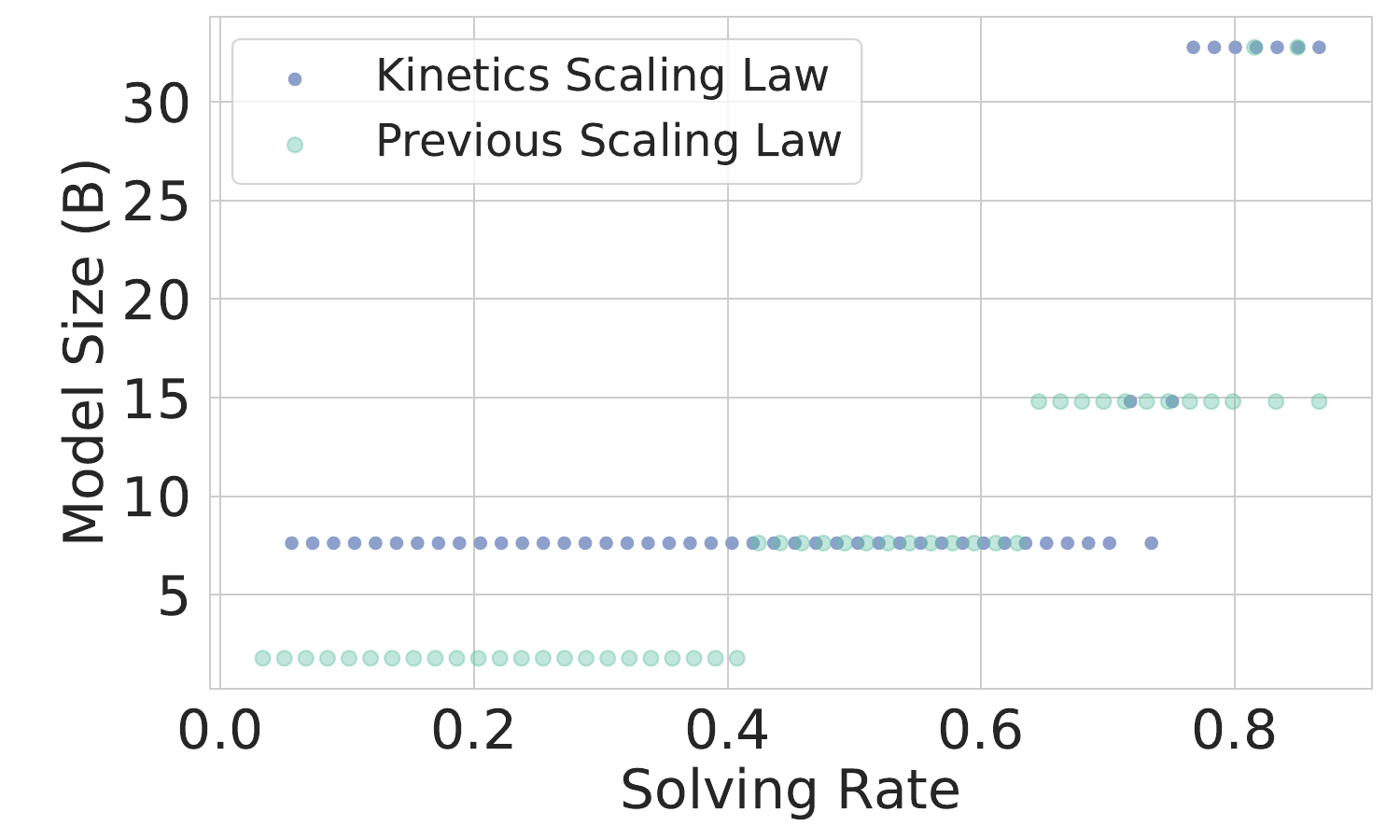}
    \label{fig: bon-model-selection-ds} 
    } 
    \caption{\textbf{AIME24 Pareto Frontier (\bon).} We conduct the same experiments as~\Cref{fig: bon-acc-eflops,fig: bon-model-selection,fig: bon-acc-flops} on DeepSeek Distilled Qwen series.
    }
\end{figure*}

 In~\Cref{fig:aime24ds,fig: bon-acc-flops-ds,fig: bon-acc-eflops-ds,fig: bon-model-selection-ds}, we evaluate DeepSeek-R1 Distilled Qwen models (abbreviated as DS models)~\citep{guo2025deepseek} on AIME24. The DeepSeek series models further demonstrate that previous scaling laws—those based on FLOPs—significantly overestimate the effectiveness of the 1.5B model. As predicted by the \law, increasing the number of generated tokens for the 1.5B model is less effective than scaling up the model size, such as using the 7B or larger variants.

Interestingly, we observe a shift in the emerging model size: unlike Qwen3, where the 14B model dominates, the 7B model becomes the dominant choice in the DeepSeek series. In~\Cref{fig:aime24ds,fig: bon-acc-eflops-ds,fig: bon-model-selection-ds}, the 7B model spans most of the Pareto frontier, and~\Cref{fig:aime24ds} shows that 7B models with \longcot are more efficient and effective than 14B models with short generations. We attribute this to an architectural outlier in the DeepSeek-R1 (Qwen2.5) model series. As shown in~\Cref{tab: model comparison}, the DeepSeek-R1 7B model is significantly more KV memory-efficient than the Qwen3-8B model. Unlike most model series illustrated in~\Cref{fig: kvtrend}, where KV cache size typically grows sublinearly with respect to model parameters, DeepSeek-R1 shows a deviation from this trend: the 14B model has approximately $3.4\times$ more KV memory than the 7B model, while having only $2\times$ more parameters.

 \begin{table}[h]
\centering
\caption{KV memory Size for Qwen3 and DeepSeek-R1 Distilled models (per 32K tokens, unit: GB).}
\begin{tabular}{lcccc}
\toprule
\textbf{Qwen3}& Qwen3-1.7B & Qwen3-8B & Qwen3-14B & Qwen3-32B \\
& 3.5 & 4.5 & 6 & 8 \\
\midrule
\textbf{DeepSeek} & DS-1.5B & DS-7B & DS-14B & DS-32B \\
& 0.875 & 1.75 & 6 & 8 \\
\bottomrule
\label{tab: model comparison}
\end{tabular}
\end{table}

This finding highlights the importance of concrete model architecture design, rather than focusing solely on the number of model parameters. Whether KV memory size is directly related to reasoning performance remains an open question, which we leave for future investigation.

\begin{figure*}[ht]
    \centering
    \subfloat[\bon Scaling Comparison]{
\includegraphics[width=0.35\linewidth]{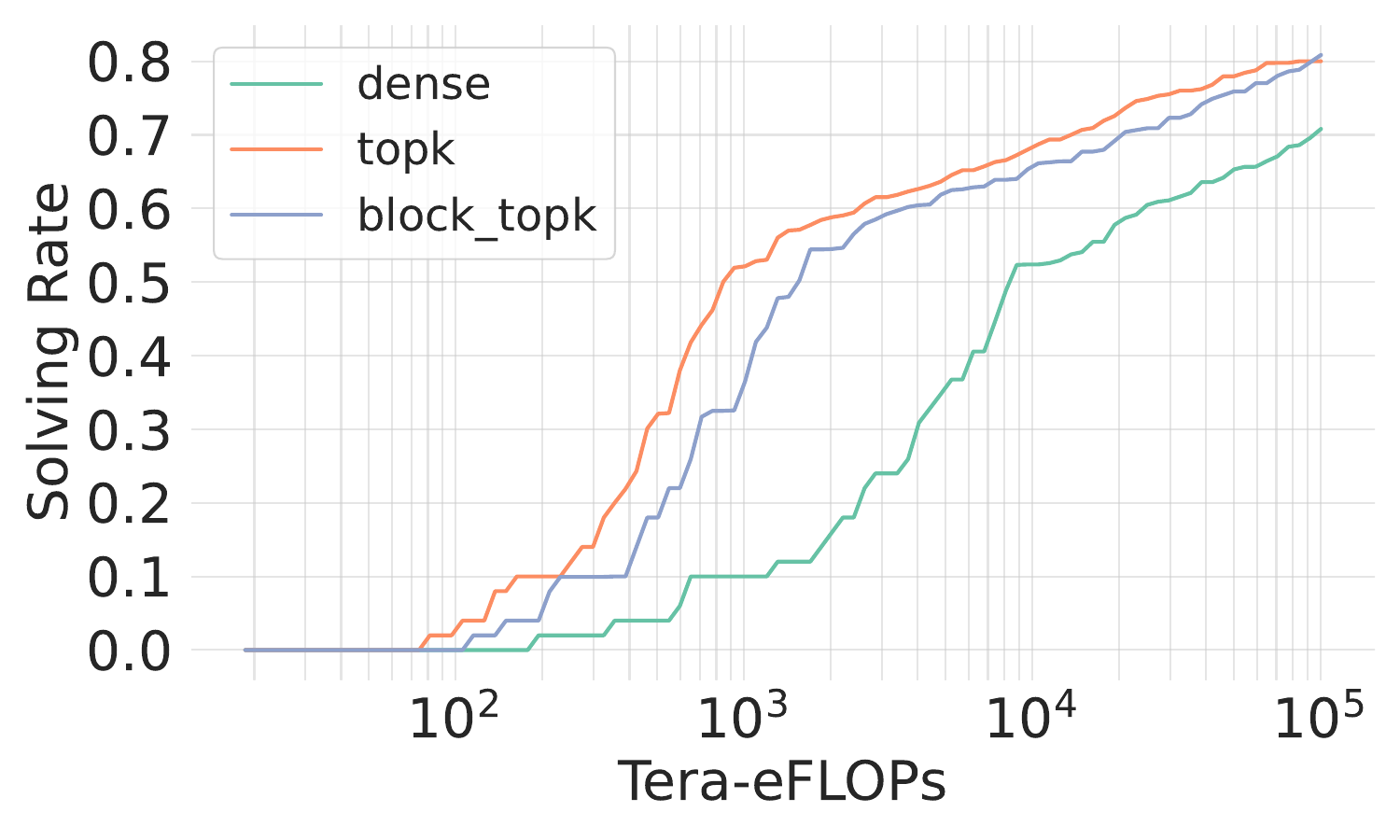}
    \label{fig: bon-lcb-comp-trial}
    }
    \subfloat[\textit{\bon} Top-$K$ Sparse Scaling]{
\includegraphics[width=0.32\linewidth]{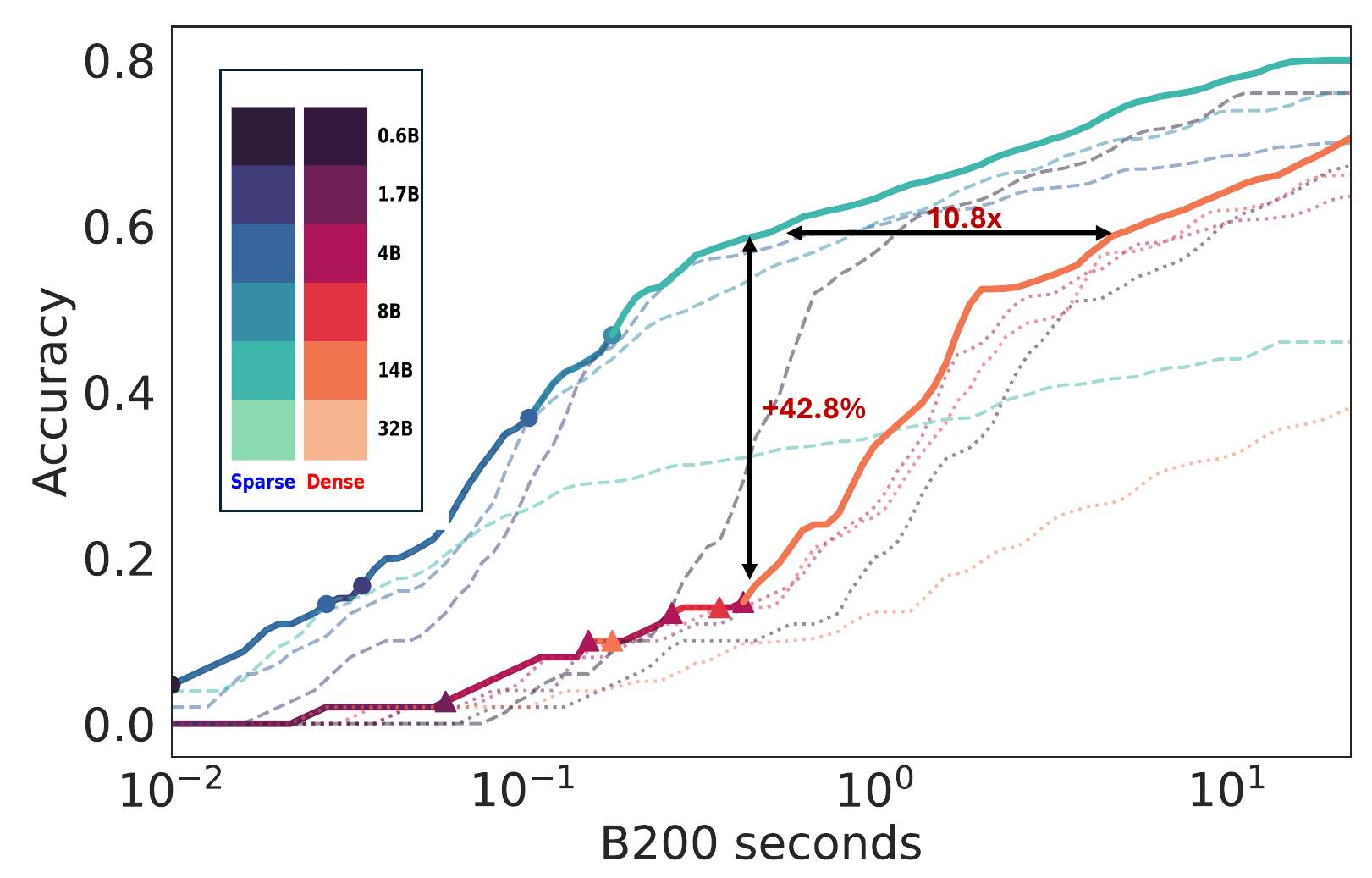}
    \label{fig: bon-lcb-oracle-scaling-trial}
    }
    \subfloat[\bon Block Top-$K$ Scaling]{
\includegraphics[width=0.32\linewidth]{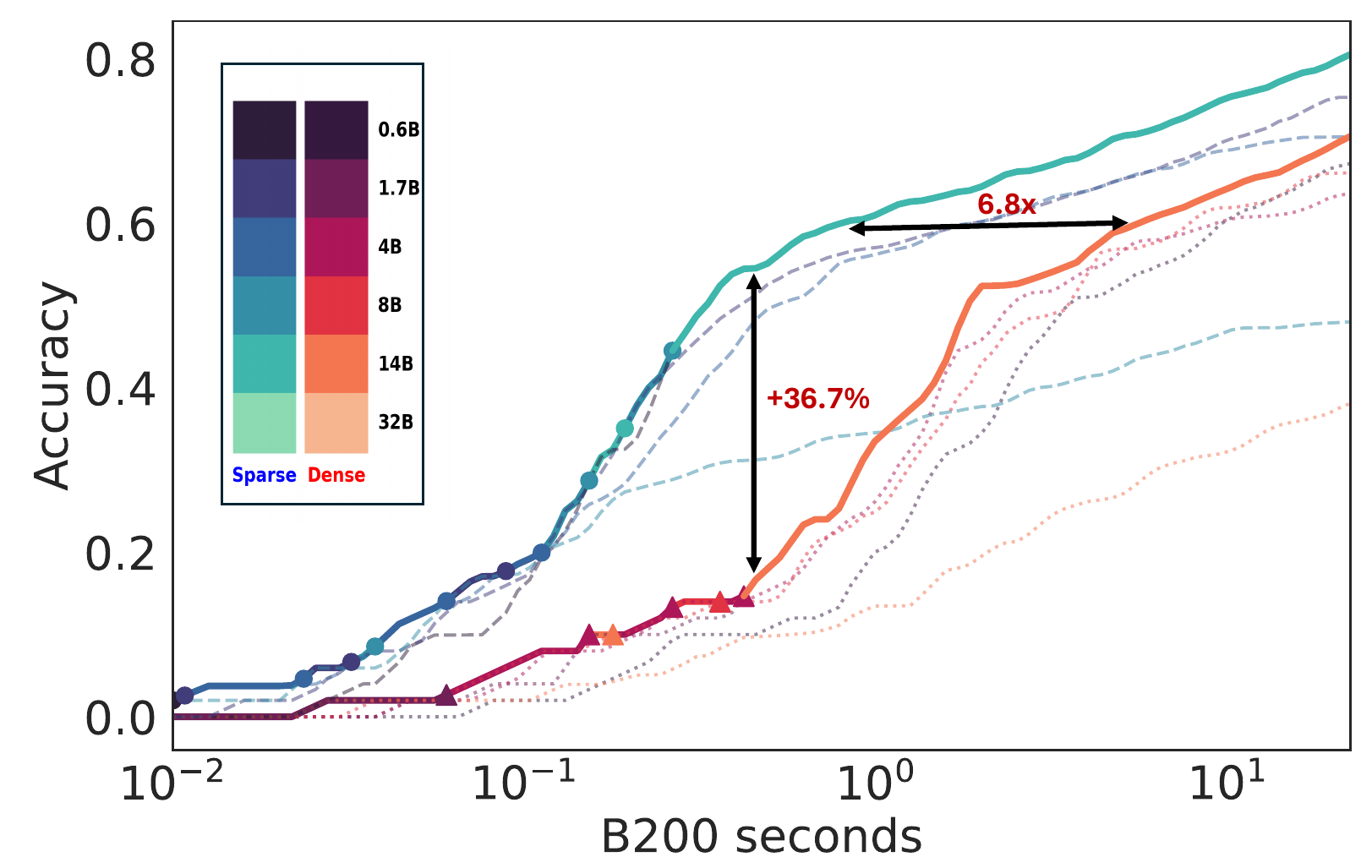}
    \label{fig: bon-lcb-blocktopk-scaling-trial}
    }\\
    \subfloat[\longcot Scaling Comparison]{
\includegraphics[width=0.35\linewidth]{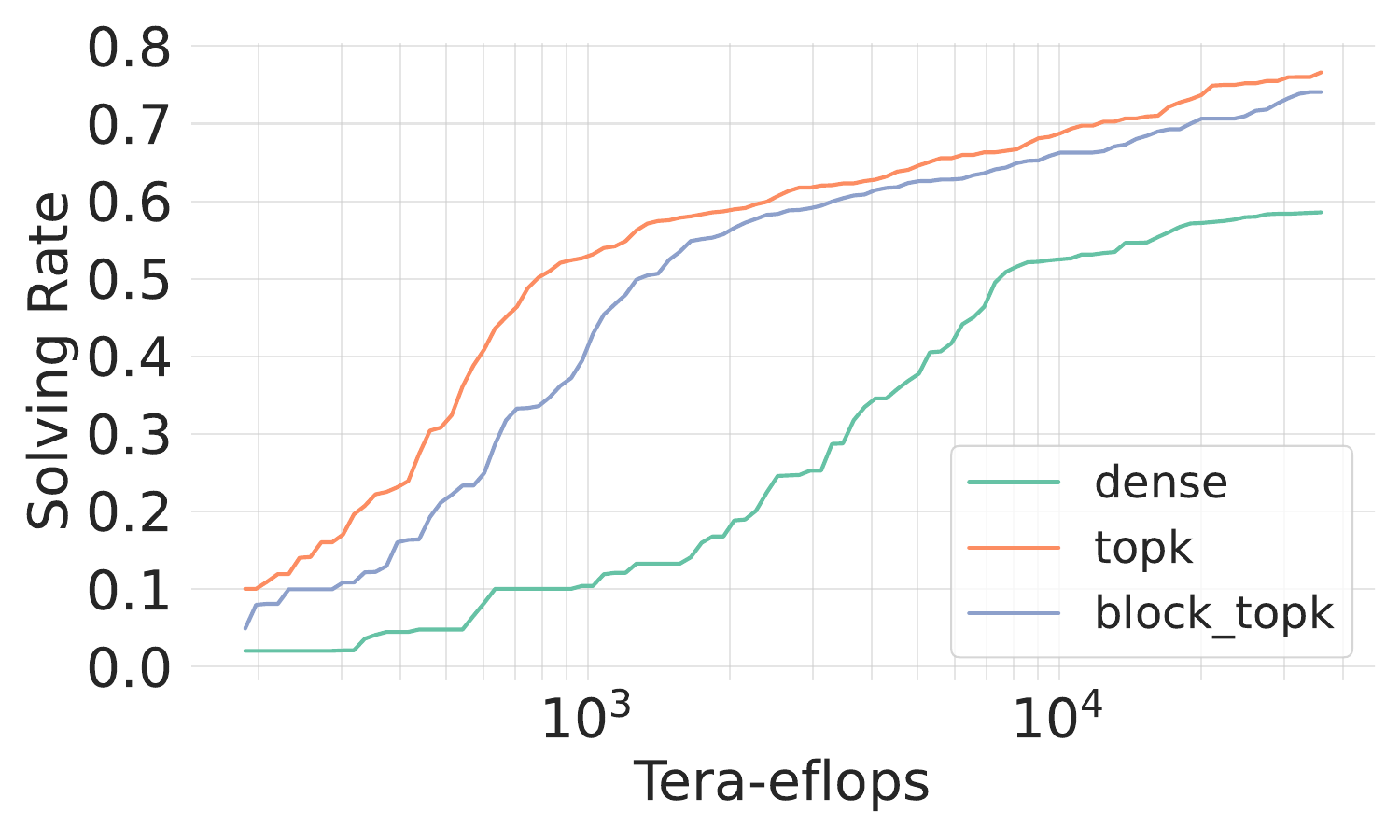}
    \label{fig: cot-lcb-comp-genlen}
    }
    \subfloat[\longcot Top-$K$ Sparse Scaling]{
\includegraphics[width=0.32\linewidth]{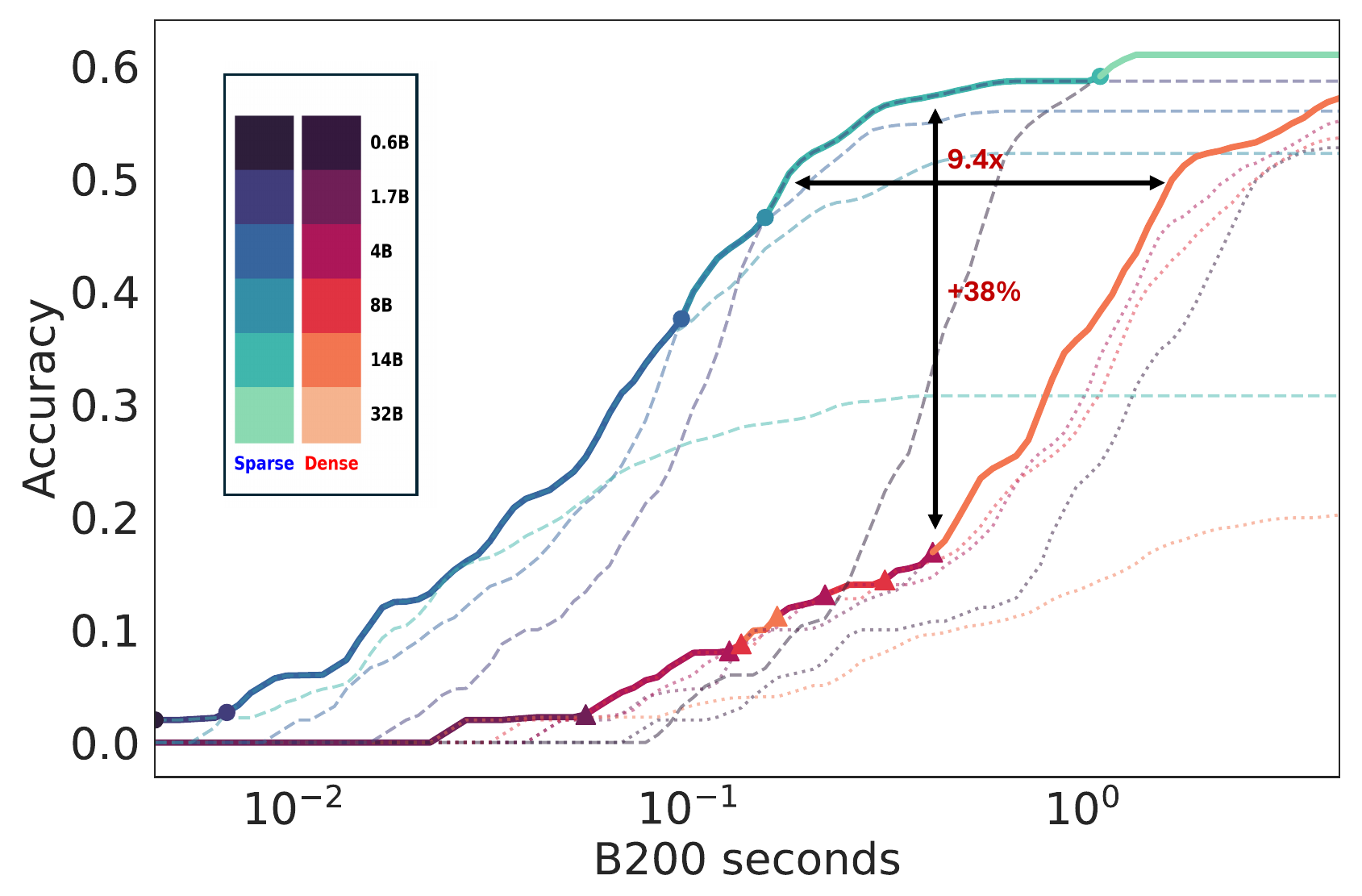}
    \label{fig: cot-lcb-oracle-scaling-genlen}
    }
    \subfloat[\longcot Block Top-$K$ Scaling]{
\includegraphics[width=0.32\linewidth]{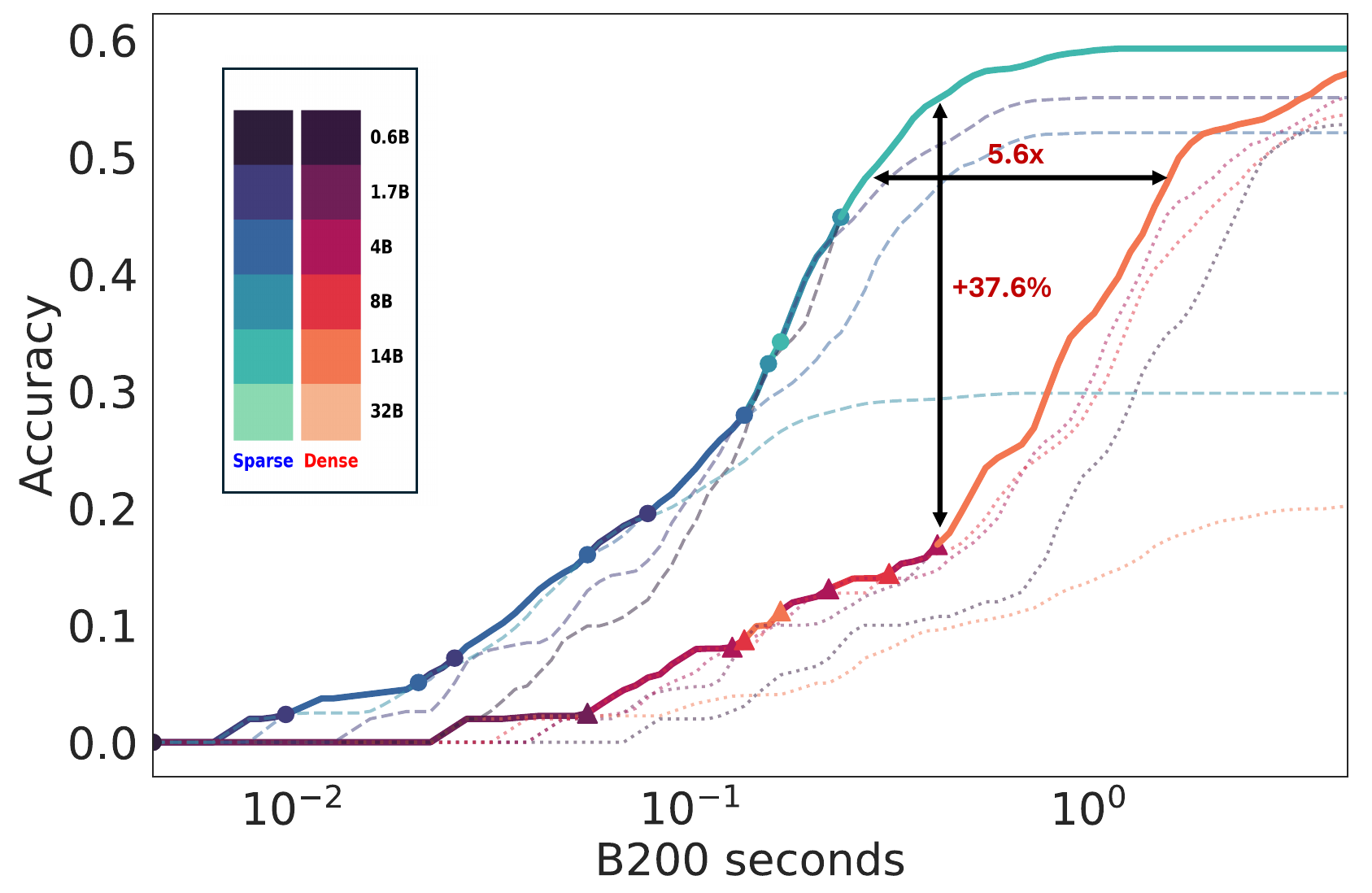}
    \label{fig: cot-lcb-blocktopk-scaling-genlen}
    }\\
    \subfloat[\bon Scaling (Easy)]{
\includegraphics[width=0.32\linewidth]{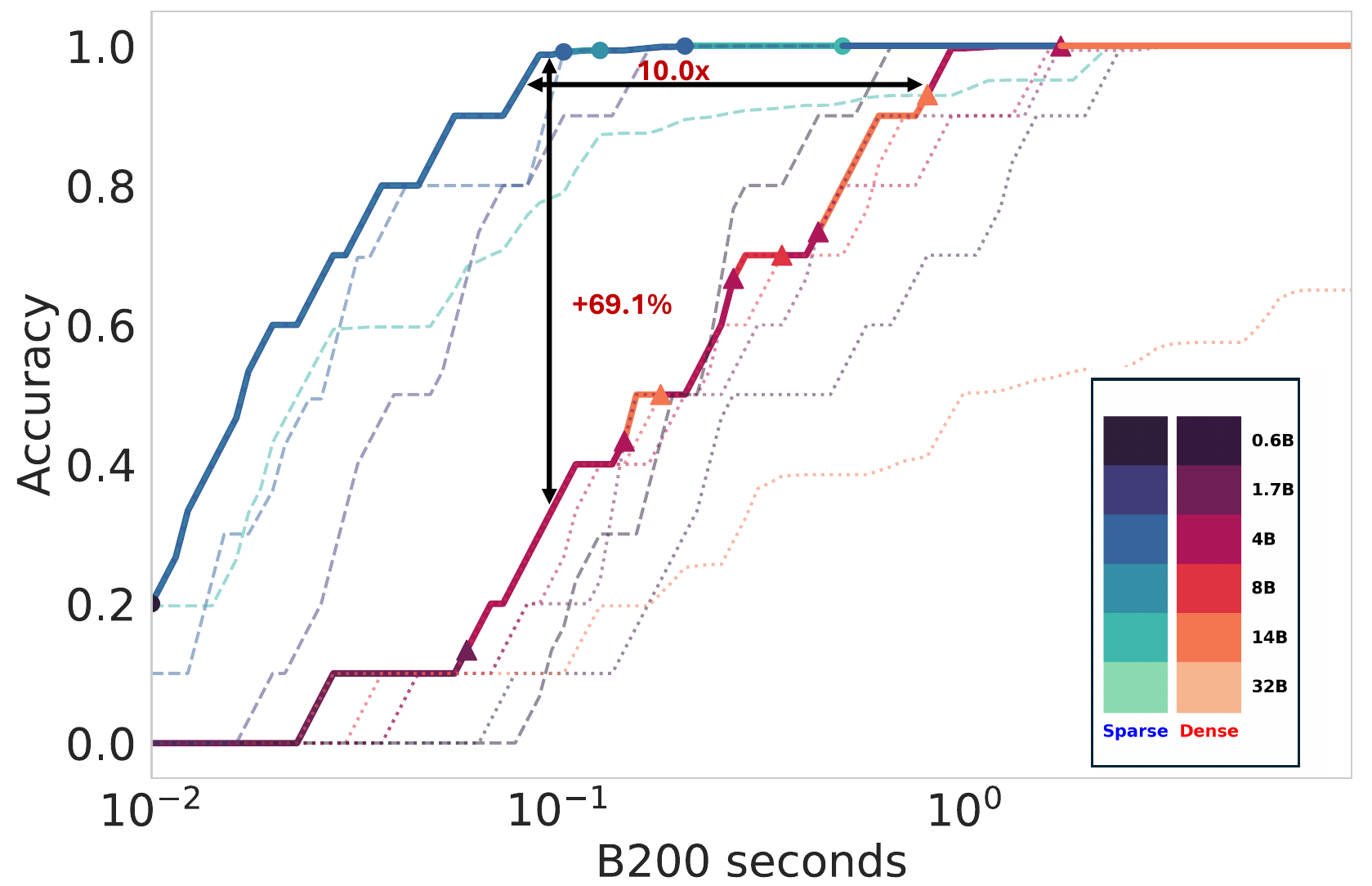}
    \label{fig: bon-lcb-oracle-easy}
    }
    \subfloat[\bon Scaling (Medium)]{
\includegraphics[width=0.32\linewidth]{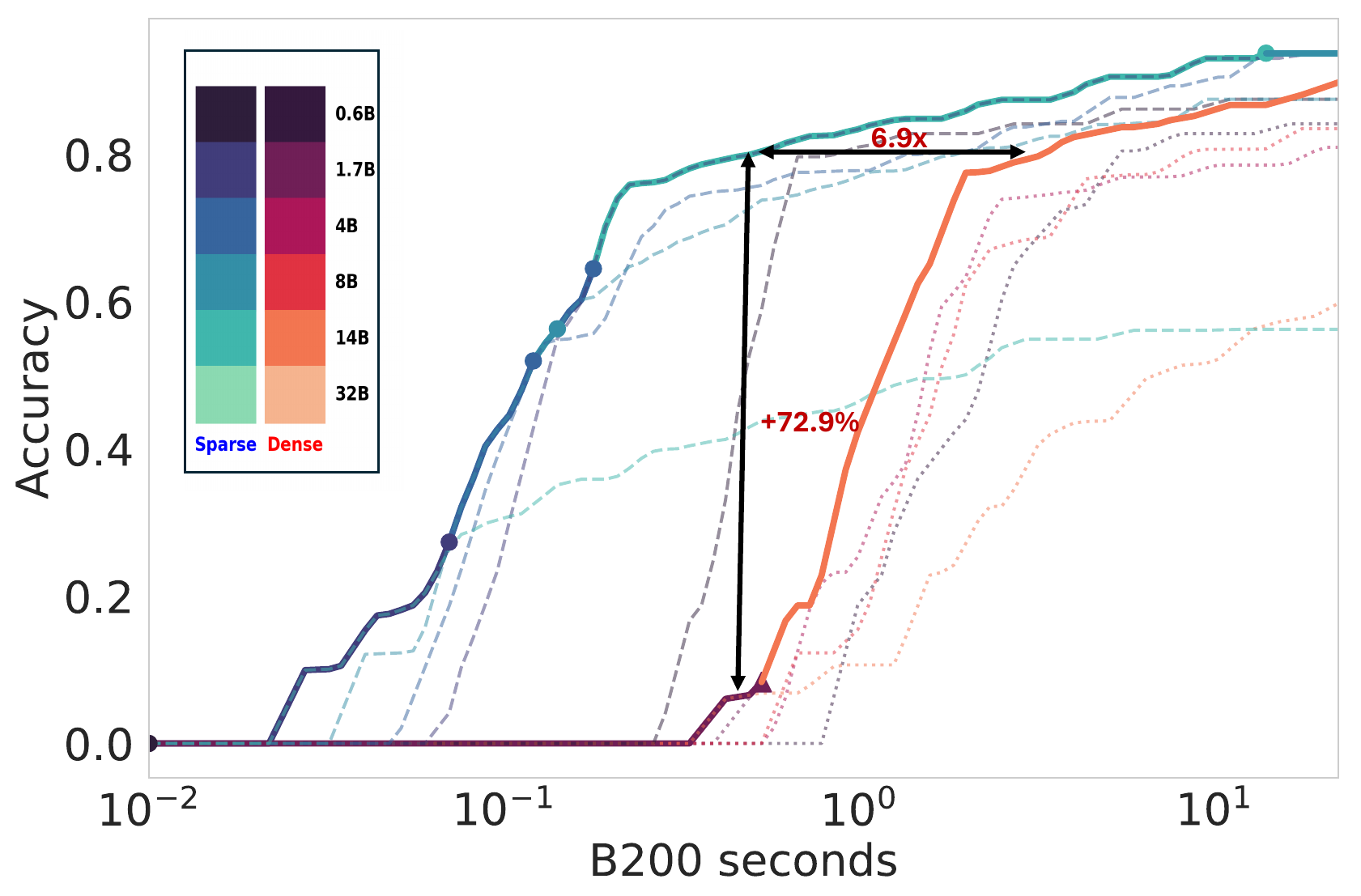}
    \label{fig: bon-lcb-oracle-medium}
    }
    \subfloat[\bon Scaling (Hard)]{
\includegraphics[width=0.32\linewidth]{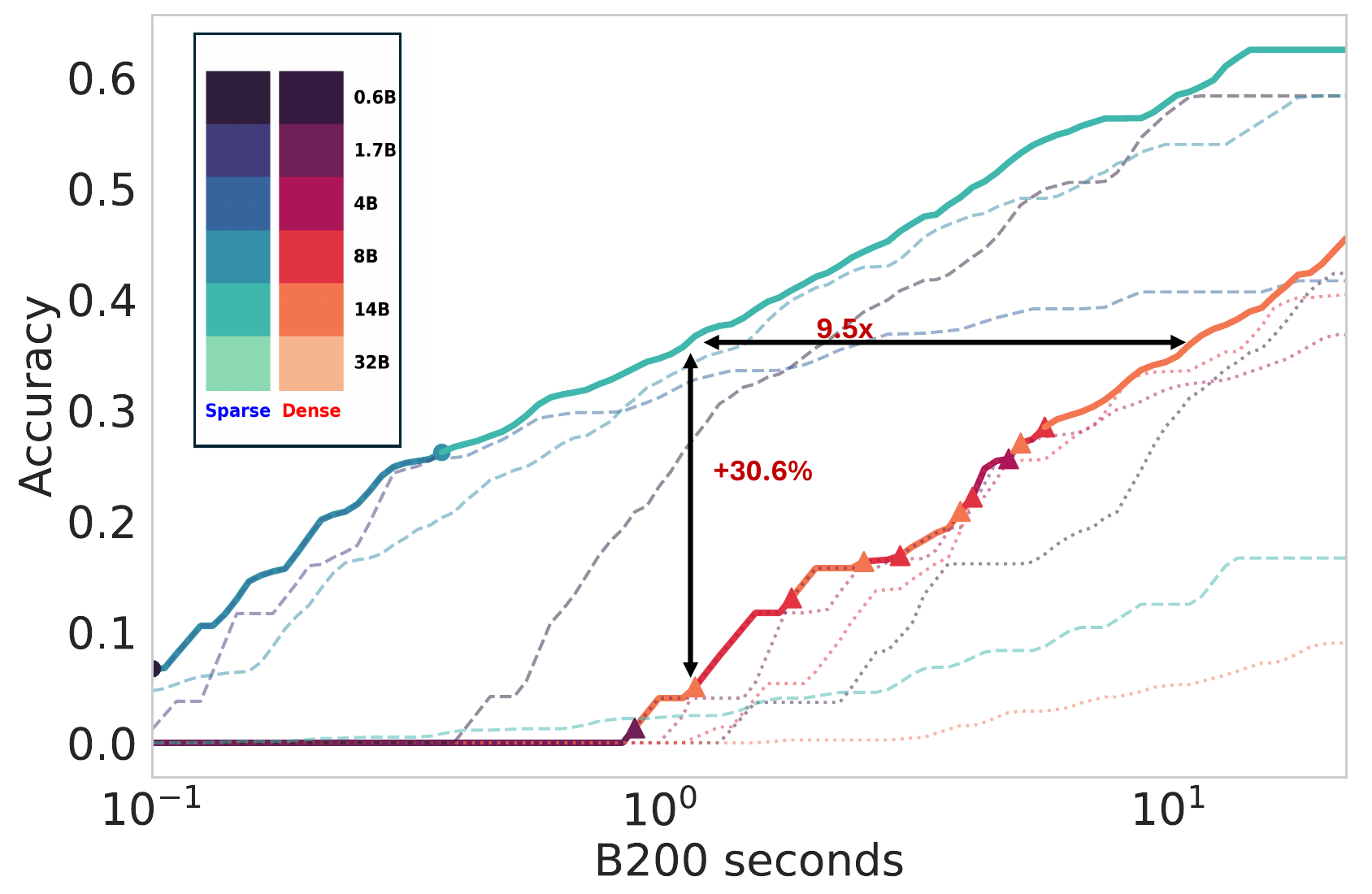}
    \label{fig: bon-lcb-oracle-hard}
    }
\caption{\textbf{LiveCodeBench Sparse Scaling.} We evaluate \sparselaw for Qwen3-14B model using oracle top-$k$ and block-top-$k$ attention on the LiveCodeBench dataset. \textbf{(a)(d)} compare block-top-$k$ and oracle top-$k$ with dense scaling under \bon and \longcot TTS settings. \textbf{(b)(e)} show cost-accuracy trade-offs for top-$k$ attention. \textbf{(c)(f)} show trade-offs for block-top-$k$ attention. \textbf{(g)(h)(i)} compare the oracle top-$k$ scaling for easy, medium and hard difficulty questions.}
\label{fig: lcb sparse scaling}
\end{figure*}

\begin{figure*}[h]
    \centering
    \subfloat[\bon Scaling Comparison]{
\includegraphics[width=0.32\linewidth]{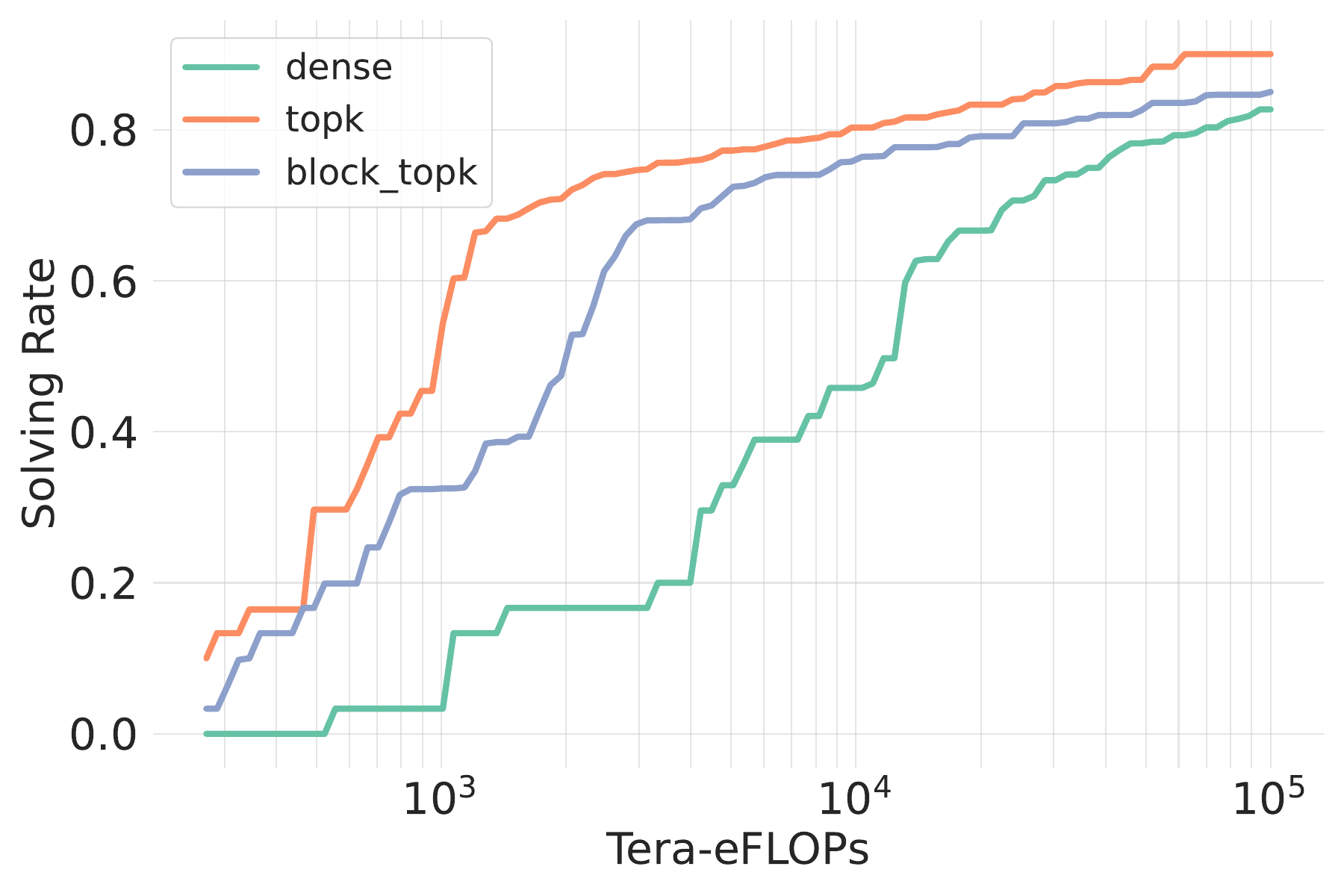}
    \label{fig: bon-aime25-comp-trial}
    }
    \subfloat[\bon Top-$K$ Sparse Scaling]{
\includegraphics[width=0.32\linewidth]{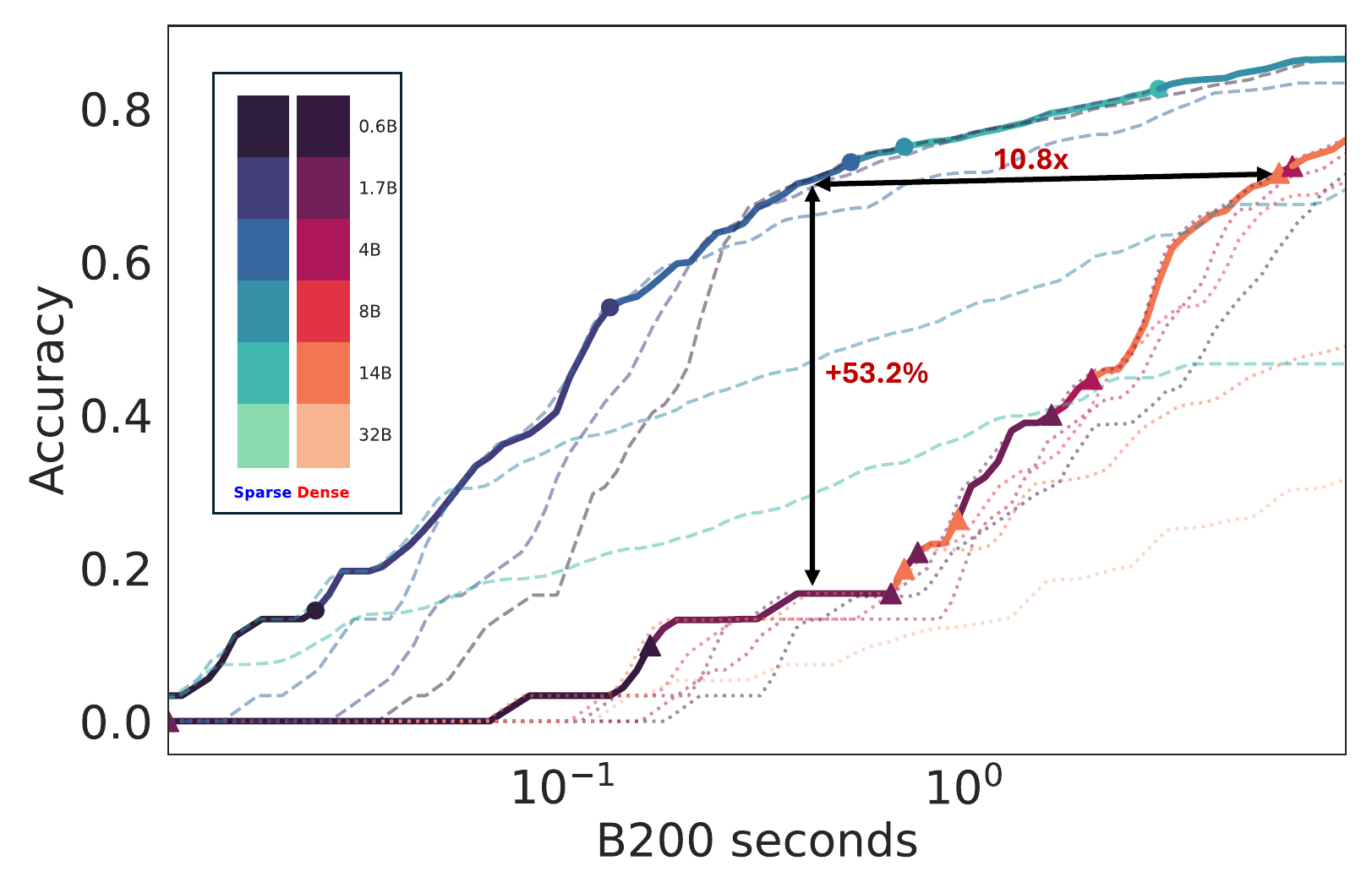}
    \label{fig: bon-aime25-oracle-scaling-trial}
    }
    \subfloat[\bon Block Top-$K$ Scaling]{
\includegraphics[width=0.32\linewidth]{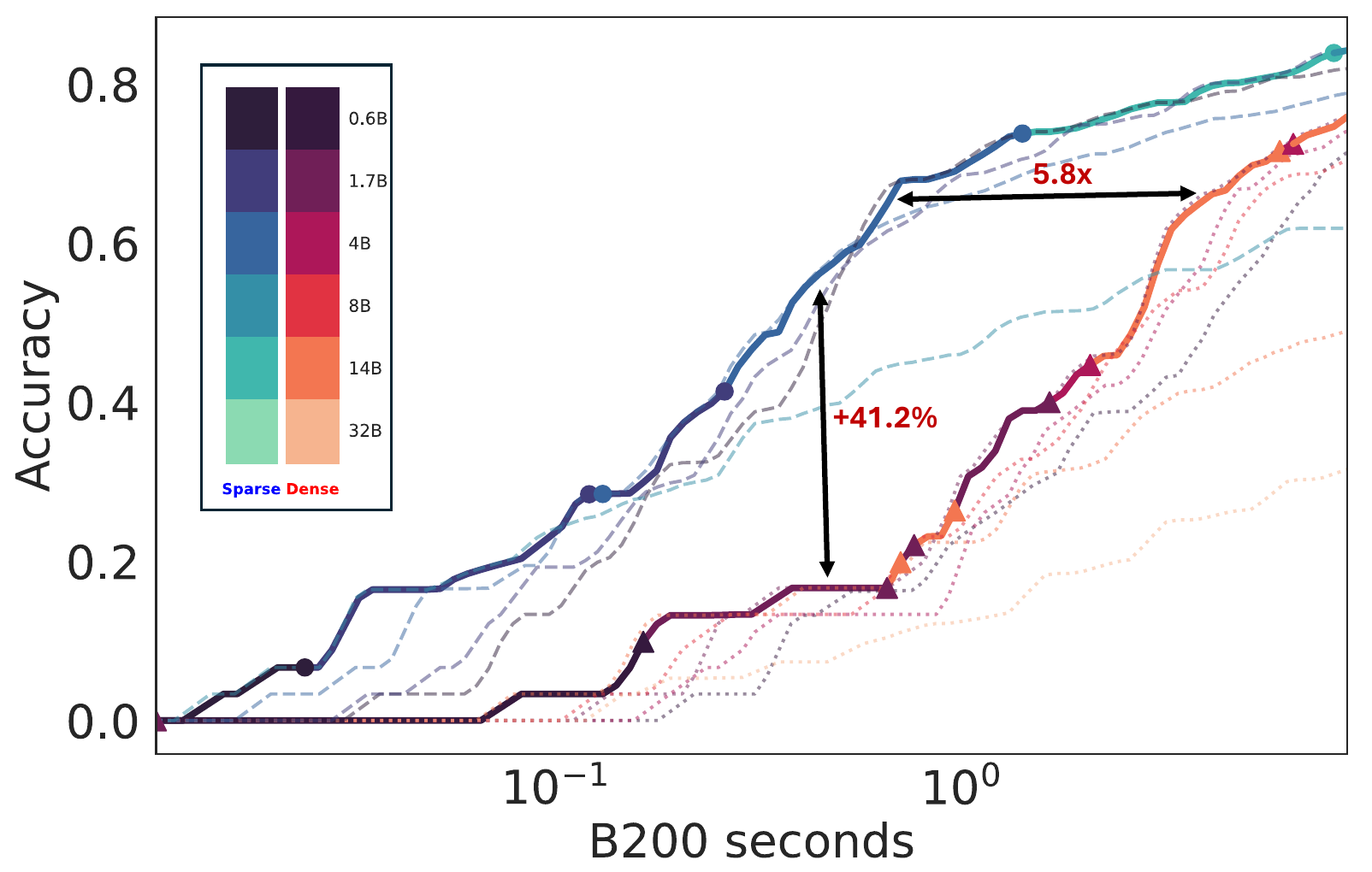}
    \label{fig: bon-aime25-blocktopk-scaling-trial}
    }\\
    \subfloat[\longcot Scaling Comparison]{
\includegraphics[width=0.32\linewidth]{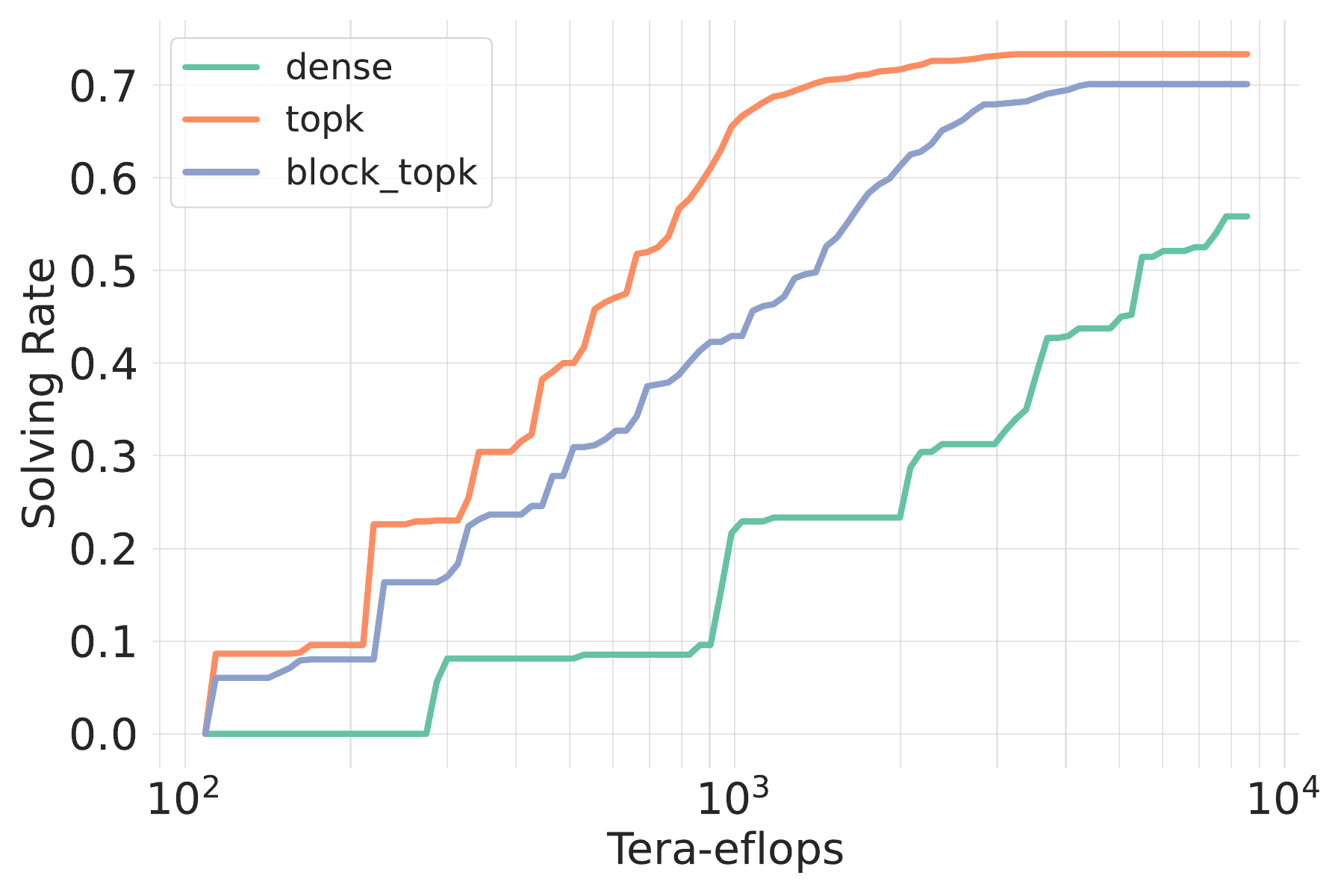}
    \label{fig: bon-aime25-comp-genlen}
    }
    \subfloat[\longcot Top-$K$ Sparse Scaling]{
\includegraphics[width=0.32\linewidth]{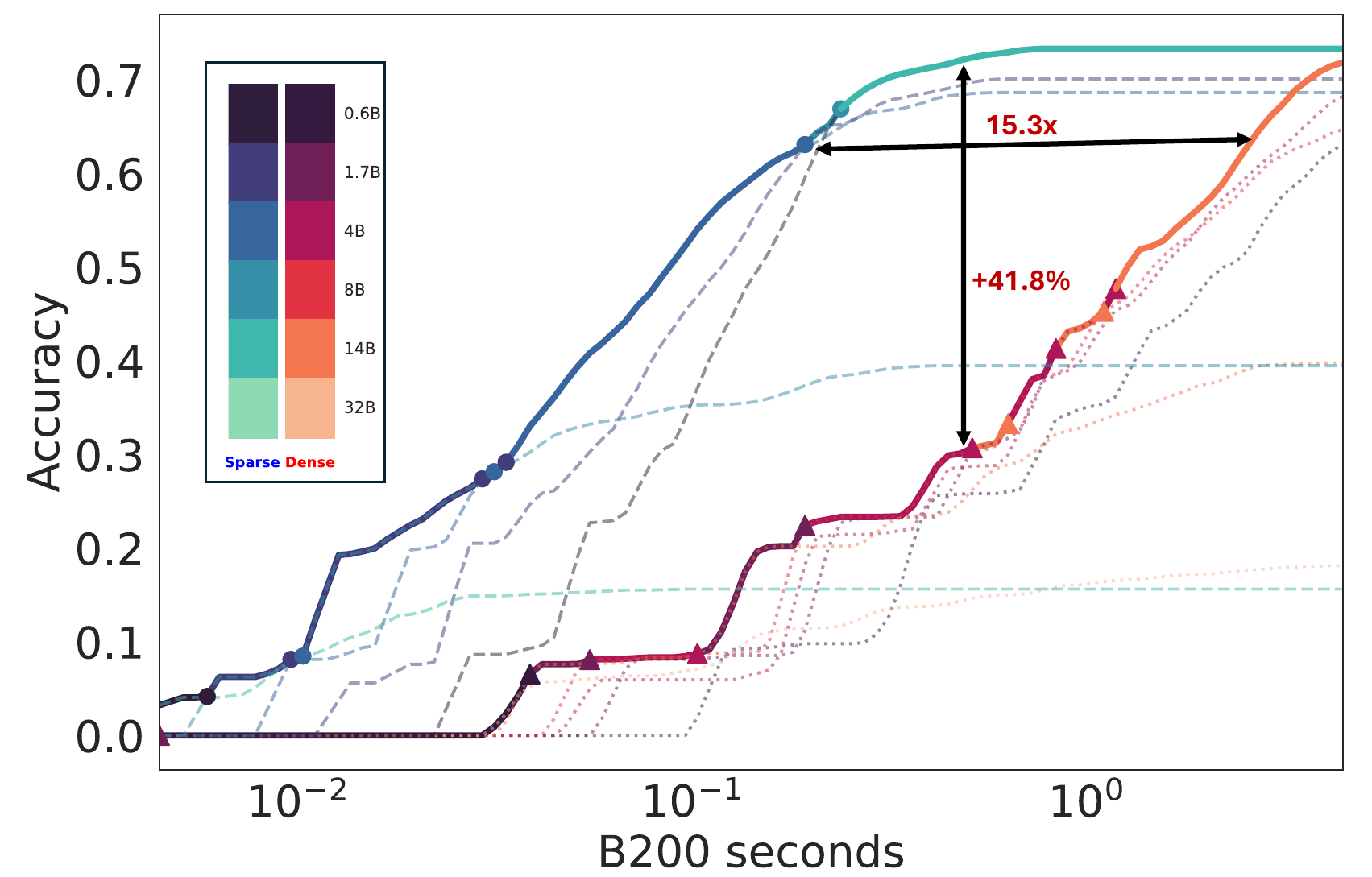}
    \label{fig: bon-aime25-oracle-scaling-genlen}
    }
    \subfloat[\longcot Block Top-$K$ Scaling]{
\includegraphics[width=0.32\linewidth]{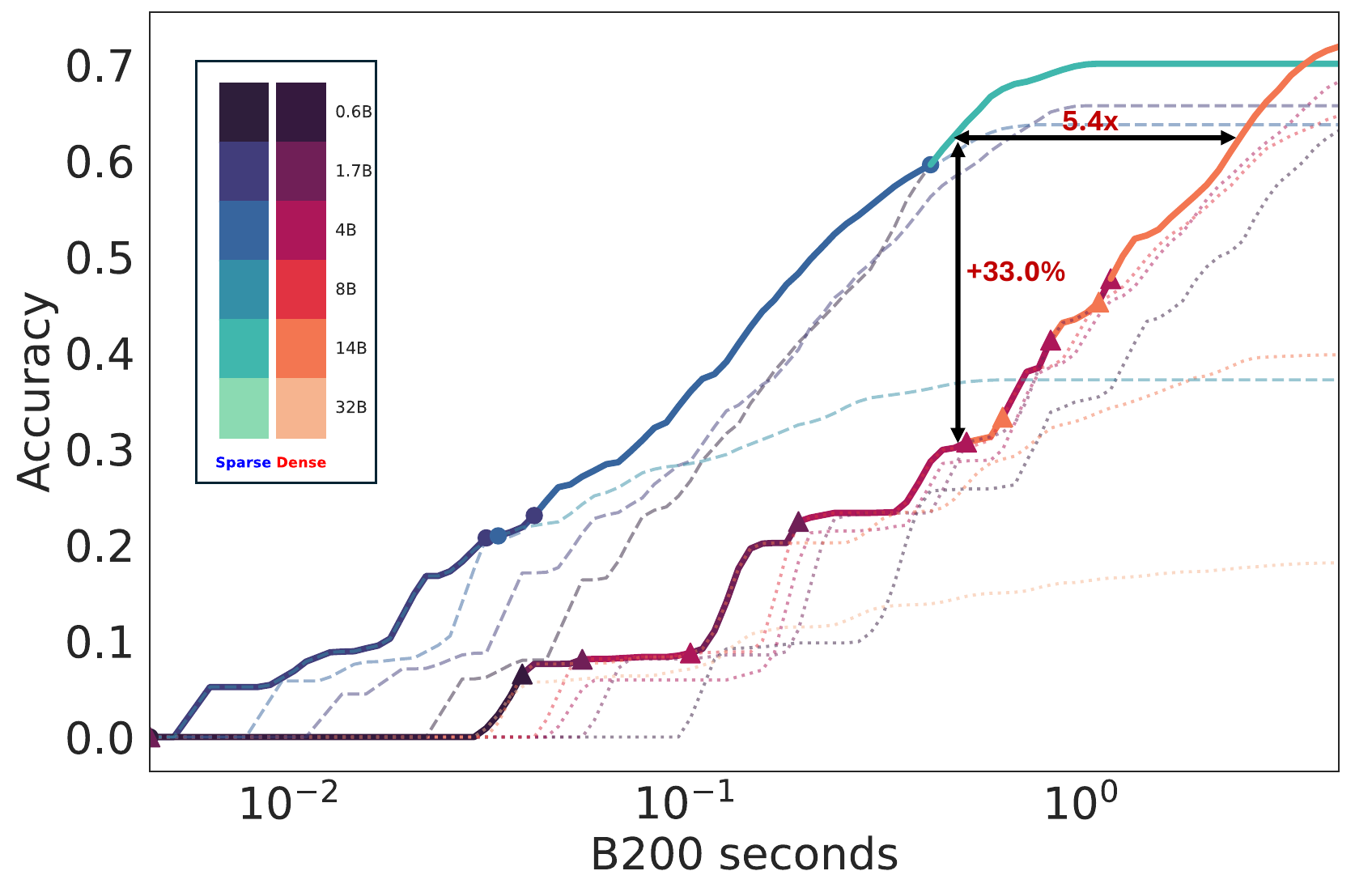}
    \label{fig: bon-aime25-blocktopk-scaling-genlen}
    }
\caption{\textbf{AIME25 Sparse Scaling.} We evaluate \sparselaw for Qwen3-14B model using oracle top-$k$ and block-top-$k$ attention on the AIME25 dataset. \textbf{(a)(d)} compare block-top-$k$ and oracle top-$k$ with dense scaling under \bon and \longcot settings. \textbf{(b)(e)} show cost-accuracy trade-offs for oracle top-$k$ attention. \textbf{(c)(f)} show trade-offs for block-top-$k$ attention.}

\label{fig: aime25 sparse scaling}
\end{figure*}

\begin{figure*}[h]
    \centering
    \subfloat[\longcot Block Top-$K$ MoE]{
\includegraphics[width=0.32\linewidth]{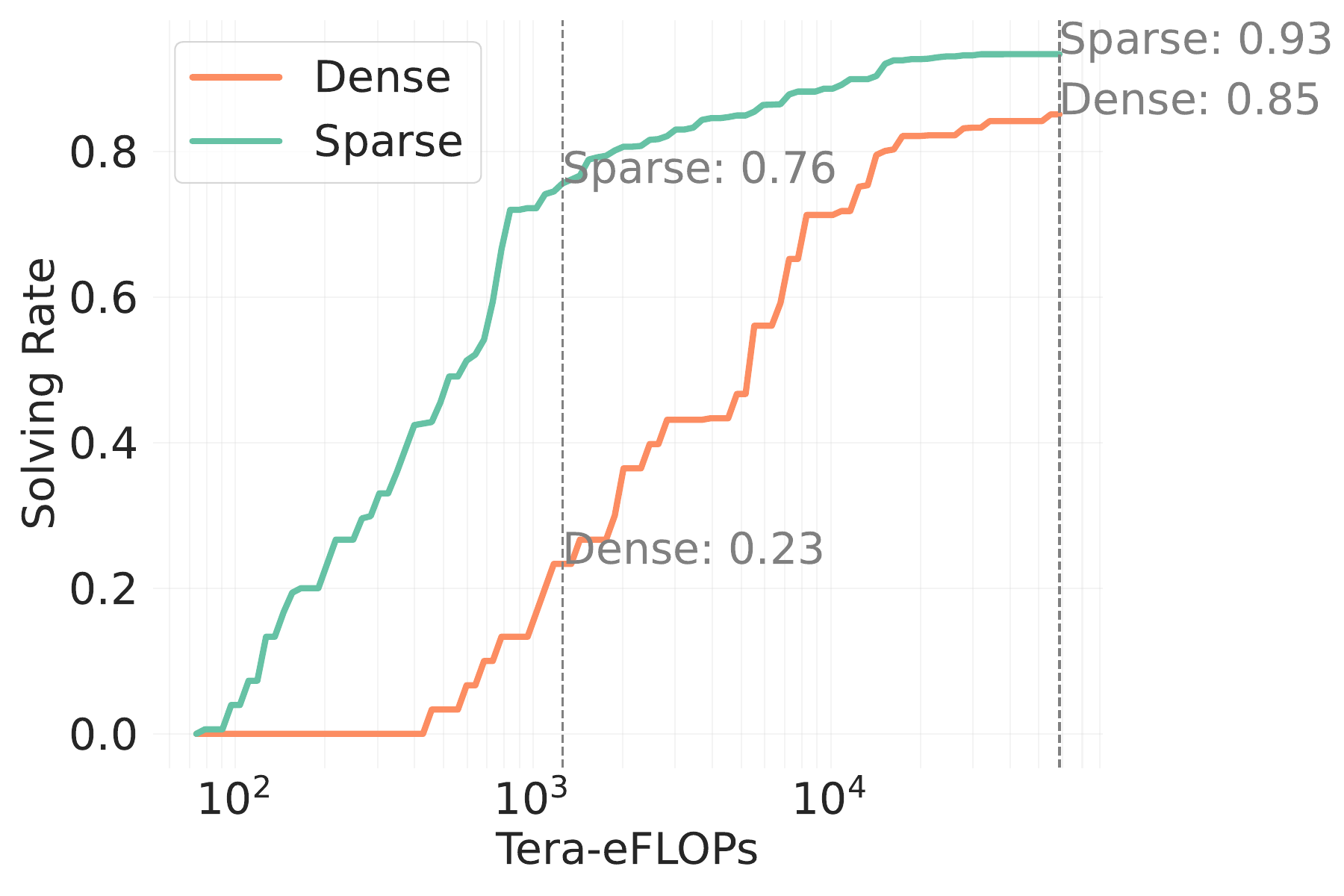}
    \label{fig: cot-moe-blk}
    }
    \subfloat[\bon Block Top-$K$ MoE]{
\includegraphics[width=0.32\linewidth]{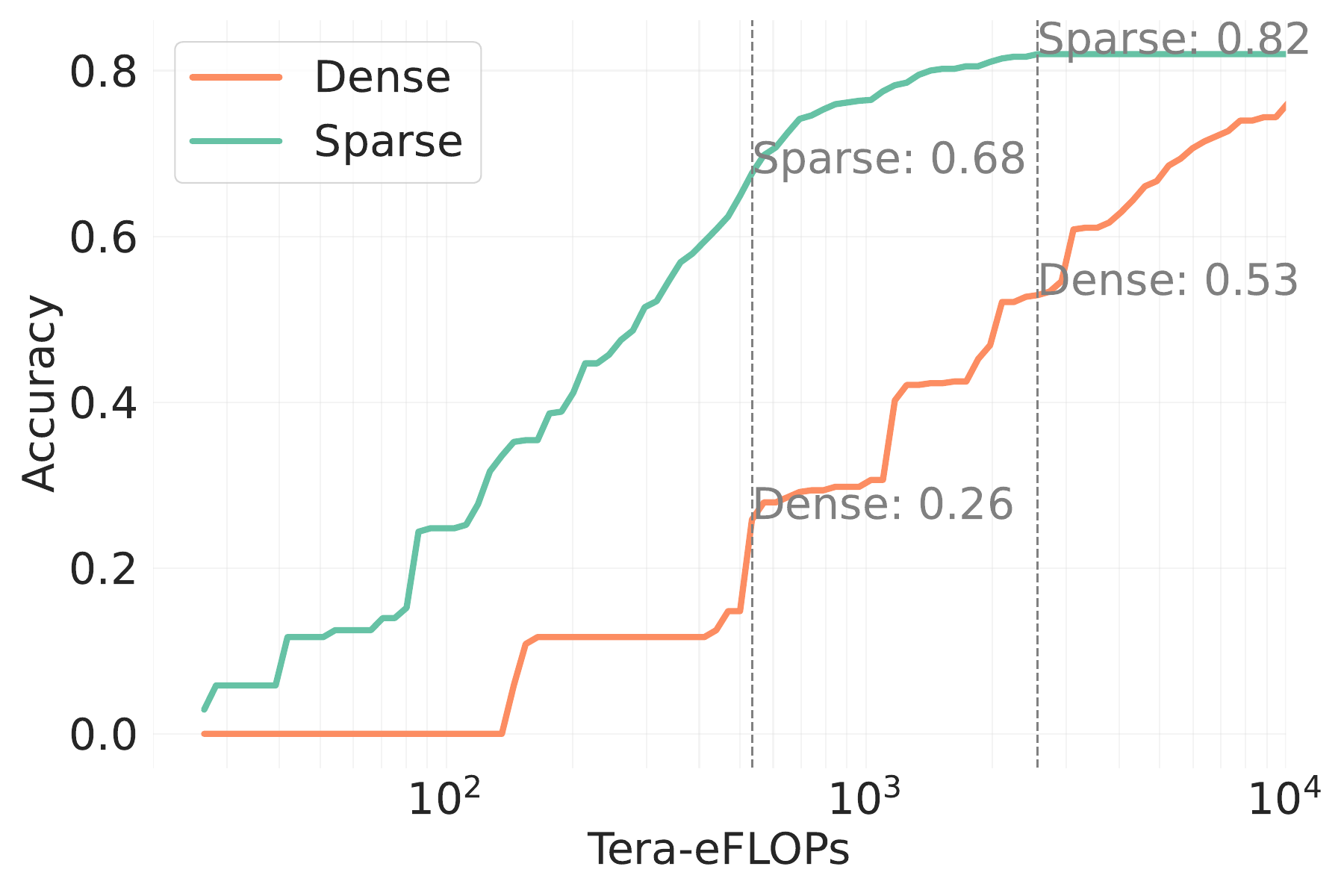}
    \label{fig: bon-moe-blk}
    }
\caption{\textbf{AIME24 MoE Block Top-$K$ scaling.}  we analyze Qwen3-30B-A3B and observe that  Block Top-$K$ 
 yields notable gains. In low-cost regimes, it can enhance problem-solving rates by \textbf{42--53 percentage points}. Even in high-cost regimes, sparse models maintain an advantage of around \textbf{8 points}, while reaching these performance levels much earlier. These findings are consistent with other (0.6$\sim$32B) models.}
\end{figure*}

\section{\sparselaw}
\label{sparsescalinglaw}
We present additional results supporting the \sparselaw across multiple tasks and demonstrate how these insights enable scalable test-time scaling with sparse attention.

\subsection{Additional Benchmarks}
\label{sparsebenchmarks}
Beyond AIME24, we evaluate our approach on LiveCodeBench~\citep{jain2024livecodebench} and AIME25~\citep{aime25}. LiveCodeBench features complex programming problems from recent coding contests, while AIME25 consists of challenging math problems. In both cases, sparse attention—particularly oracle top-$k$—consistently outperforms dense attention. Block top-$k$ attention, a tractable alternative, closely matches the performance of the oracle.

For LiveCodeBench, we sample 50 problems from the \textit{v5} subset (24 hard, 16 medium, 10 easy). As shown in~\Cref{fig: lcb sparse scaling}, oracle top-$k$ attention can achieve $\sim 10\times$ speedup in high-accuracy regimes and improves coverage by $40$–$50\%$ in low-cost regimes. Conversely, the tractable alternative, Block top-$k$ yields $5$–$6\times$ speedup and $30$–$40\%$ coverage gains. We further show how the benefits of sparse attention scale with problem difficulty~(\Cref{fig: bon-lcb-oracle-easy,fig: bon-lcb-oracle-medium,fig: bon-lcb-oracle-hard}).~\Cref{fig: aime25 sparse scaling} confirms similar trends for AIME25, with substantial gains in both accuracy and efficiency under sparse attention. 

We present the evaluations of MoE models (Qwen3-30B-A3B) in~\Cref{fig: bon-moe-blk,fig: cot-moe-blk}, where we draw consistent conclusions on the scalability of sparse attention.

\subsection{Additional Analysis}
\label{sparseanalysis}
\begin{figure*}
    \centering
    \subfloat[AIME25 Gen.]{
\includegraphics[width=0.24\linewidth]{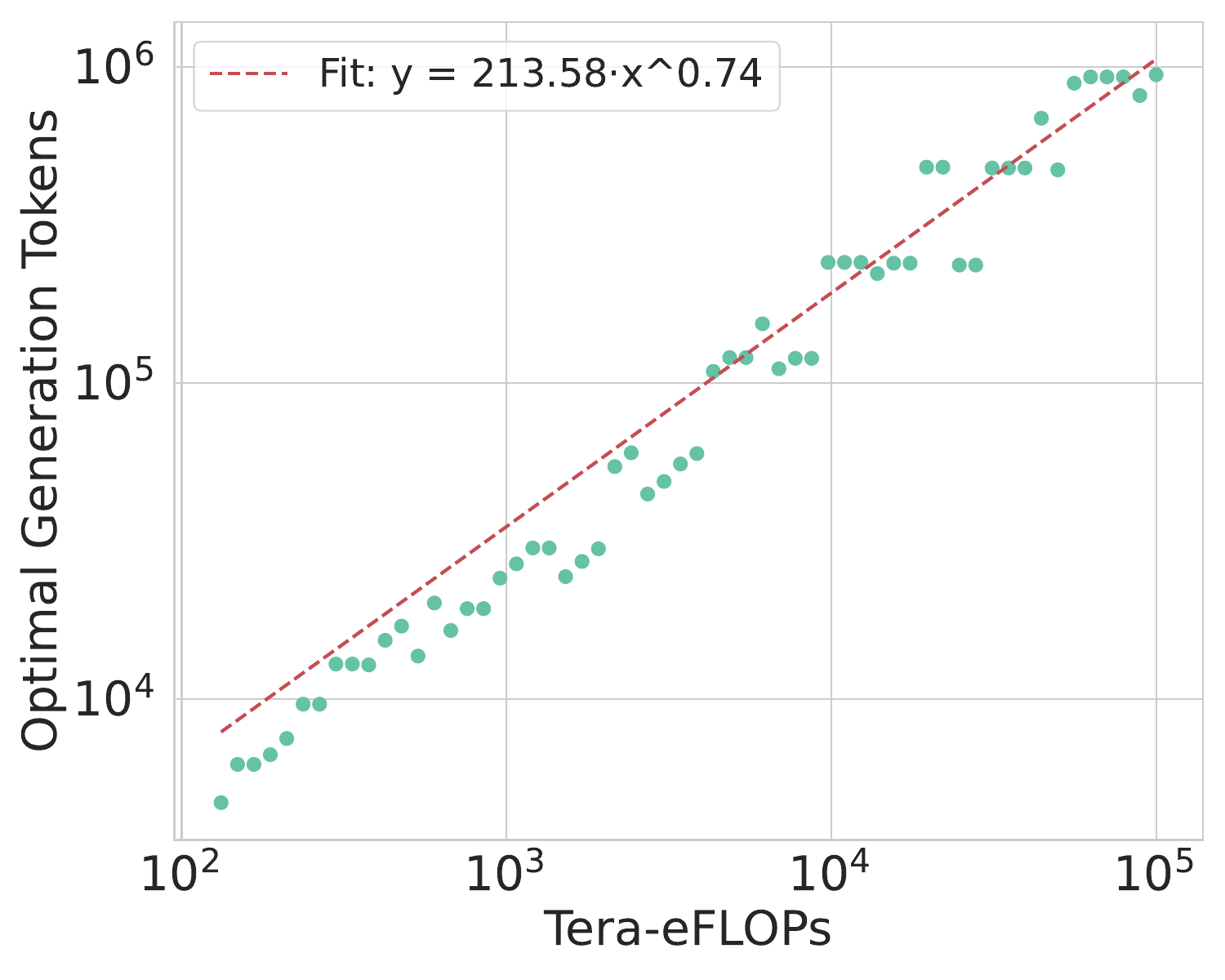}
    \label{fig: bon-8b-kv-aime25}
    }
  \subfloat[AIME25 Budget]{
\includegraphics[width=0.24\linewidth]{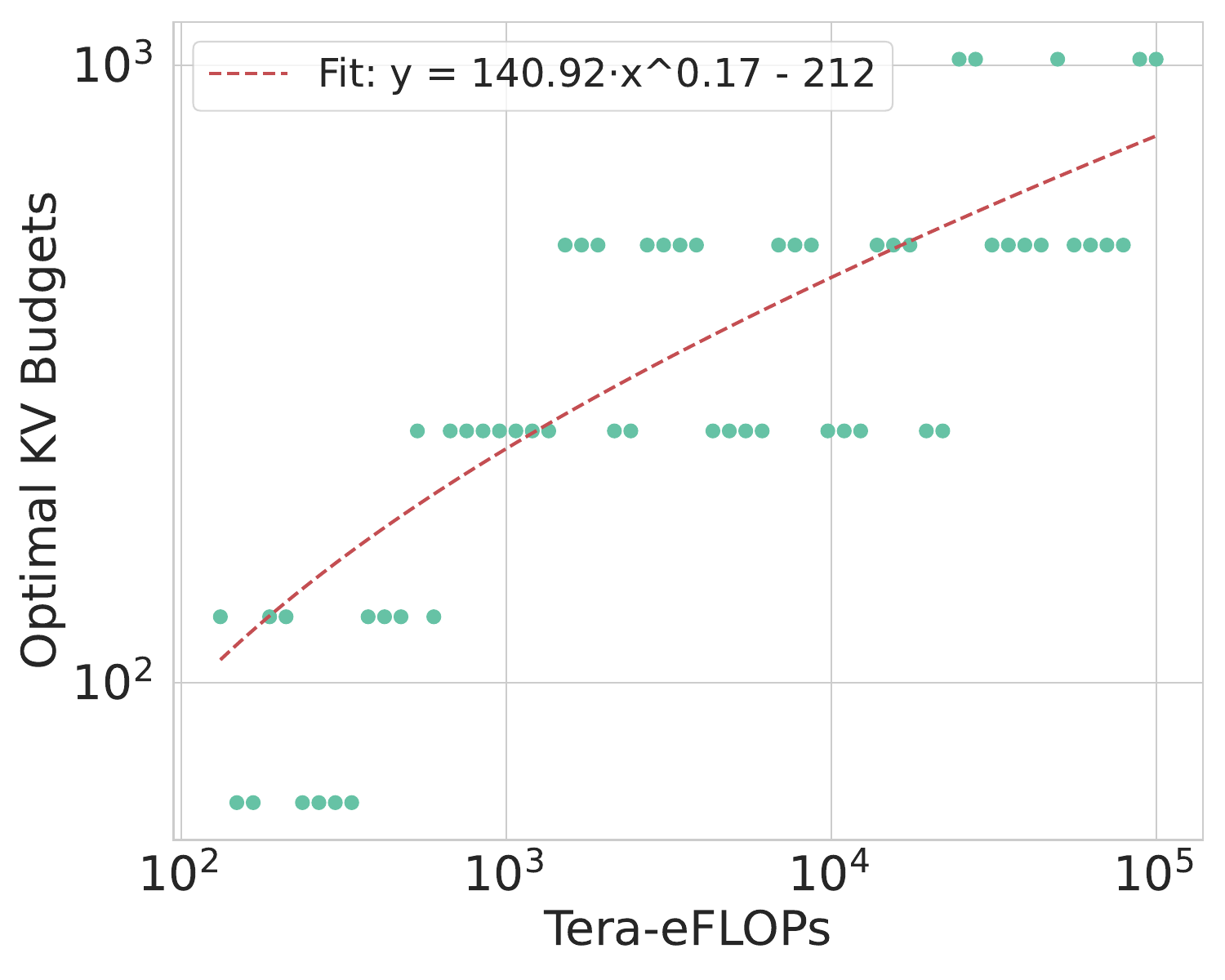}
    \label{fig: bon-8b-tokens-aime25} 
    }
    \subfloat[LiveCodeBench Gen.]{
\includegraphics[width=0.24\linewidth]{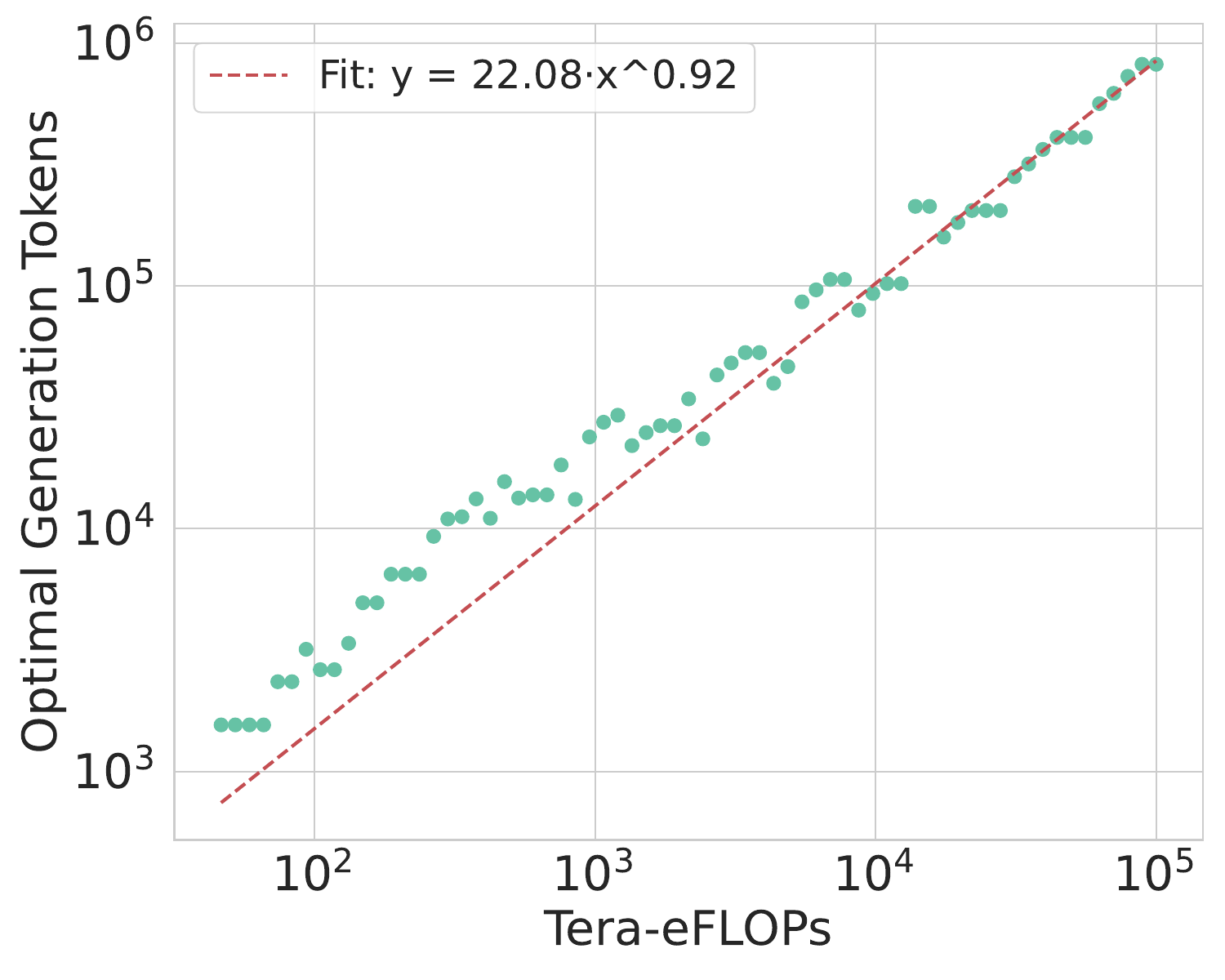}
    \label{fig: bon-8b-kv-lcb} 
    } 
     \subfloat[LiveCodeBench Budget]{
\includegraphics[width=0.24\linewidth]{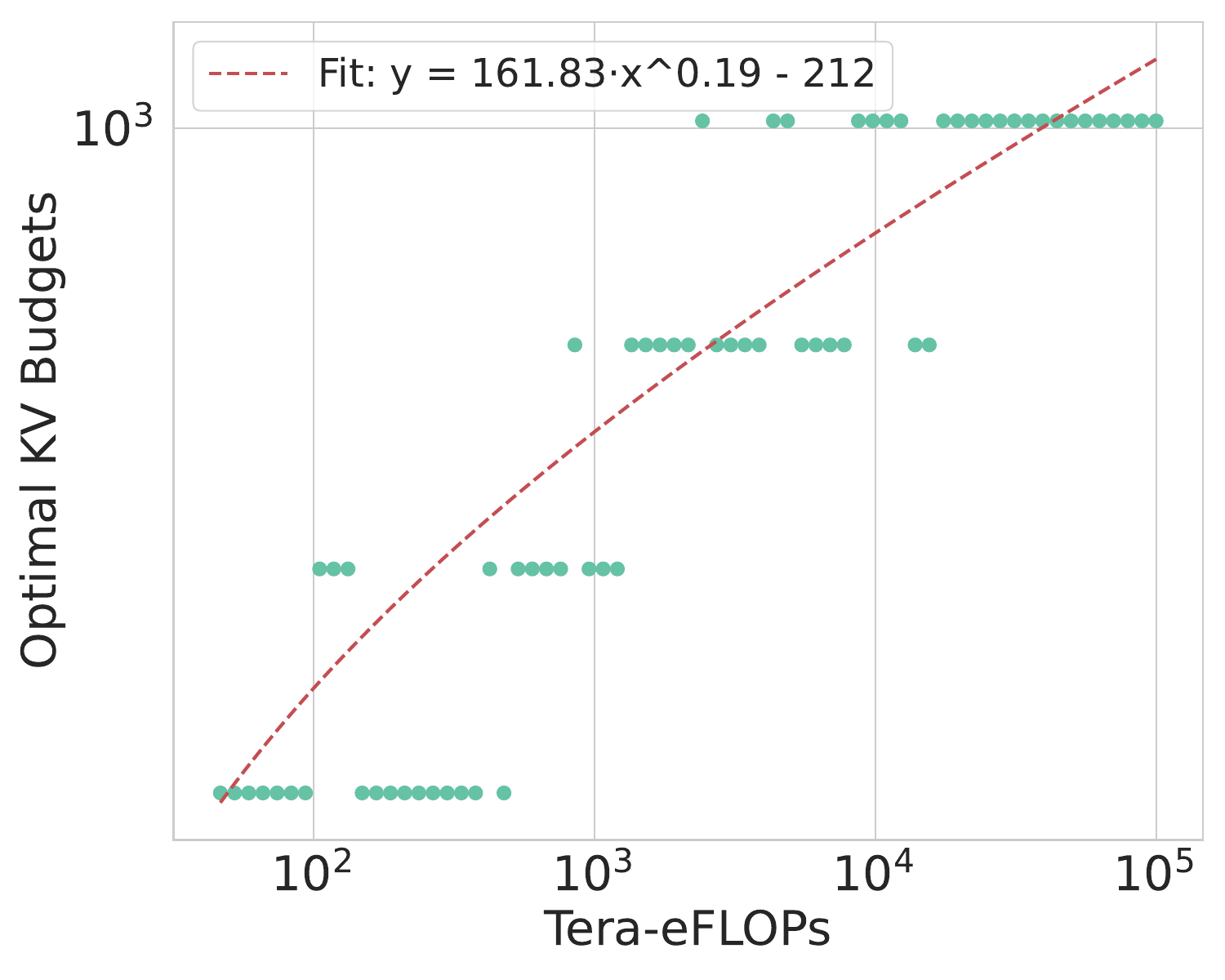}
    \label{fig: bon-8b-tokens-lcb} 
    } 
    \caption{\textbf{Tradeoff Between Generated Tokens and KV Budget.} We characterize the tradeoff between increasing generation length and allocating a larger KV cache budget using Qwen3-8B. For AIME25 (\textbf{(a)(b)}) and LiveCodeBench (\textbf{(c)(d)}), we identify the optimal KV budget and generated tokens (defined as number of reasoning trials times the average generated tokens per trial) to achieve the highest problem-solving rate under every cost constraint $C$.
}   
\vspace{-0.3cm}
\end{figure*}

 Fixing a model (e.g., Qwen3-8B), we investigate the tradeoff between generating more tokens through \bon and increasing the KV budget in~\Cref{fig: bon-8b-kv-aime25,fig: bon-8b-tokens-aime25,fig: bon-8b-kv-lcb,fig: bon-8b-tokens-lcb}. As the figures suggest, on AIME25, each doubling of total compute cost increases the optimal KV budget by \(1.13\times\), while generated tokens grow by \(1.67\times\); on LiveCodeBench, these factors are \(1.14\times\) and \(1.89\times\), respectively. We find that although the concrete numbers depend on the types of tasks, the overall results confirm our suggestions in the main paper that allocating compute toward generating more responses is generally more effective than expanding KV budget, highlighting the scalability of sparse attention. 
 
\section{Experimental Details}
\label{expdetails}
In this section, we explain the details about our experiments.
\subsection{Estimate Cost, Optimal Generation Tokens, Accuracy and Solving Rate}
\label{costaccuracysolving}
When empirically measuring cost, one major challenge is the difficulty of controlling the actual generation length. Although it is possible to set an upper bound on the number of generated tokens, there is no guarantee that the model will utilize the full budget. For instance, in our Best-of-\(N\) experiments, we set the maximum number of generated tokens to 32{,}768, yet the average generation length was only 14K–16K tokens.

Furthermore, it is important to model the relationship between actual inference cost and performance metrics, such as accuracy in \longcot or solving rate in Best-of-\(N\). Relying solely on the maximum allowed generation length to estimate cost can substantially underestimate the efficiency of models that solve problems with much shorter responses—an ability that \textbf{may} reflect higher capability.

To address this challenge, we first sample \( S \) independent reasoning traces \( r_1, r_2, \ldots, r_S \) from model \( M \) on task \( T \), with the maximum allowed number of tokens set to \( n \). We slightly generalize~\Cref{eq: cost model full equations} as:
\begin{align}
    C_{\mathrm{TTS}} &= 2NP\mathbb{E}[L_{\text{out}}] + 2rNL_{\text{in}}D\mathbb{E}[L_{\text{out}}] + rND\mathbb{E}[L_{\text{out}}^2] \nonumber \\
    &\quad + 2IL_{\text{in}}D\mathbb{E}[L_{\text{out}}] + IND\mathbb{E}[L_{\text{out}}^2] \nonumber \\
    &= a\mathbb{E}[L_{\text{out}}] + b\mathbb{E}[L_{\text{out}}^2] + c,
    \label{eq:cost estimate}
\end{align}
where \( a \), \( b \), and \( c \) are constants determined by the model architecture and test-time strategies (e.g., the value of \( n \)). The expectations are estimated from the sampled traces, whose distribution is influenced by the model \( M \), the token limit \( n \), and the task \( T \).

\textbf{For \longcot}, we fix \( N = 1 \) in~\Cref{eq:cost estimate} and vary \( n \). From the sampled traces, we estimate the accuracy (Pass@1), and compute the corresponding cost by substituting the empirical values of \( \mathbb{E}[L_{\text{out}}] \) and \( \mathbb{E}[L_{\text{out}}^2] \) measured under each \( n \).

\textbf{For Best-of-\(N\)}, we fix \( n = 32{,}768 \), and estimate the solving rate (Pass@\(K\)) following the methodology of~\citet{brown2024large}. The corresponding cost is then computed by substituting \( N = K \) into~\Cref{eq:cost estimate}.

Similarly, we can estimate the cost for sparse attention models using~\Cref{eq: sparse,eq: search}. 

Advanced control of generation lengths is an active area of research~\citep{yang2025qwen3technicalreport,muennighoff2025s1,ma2025reasoning}, but it is beyond the scope of this paper.

\textbf{Optimal Generation Tokens.} To address the question: \textit{Given a total cost budget \( C \), what proportion should be allocated to generating longer responses, in contrast to enlarging model sizes or reducing attention sparsity?}, we project the optimal number of generation tokens from the Pareto frontier (e.g.,~\Cref{fig: bon-8b-tokens,fig:modelcost}). Intuitively, each point on the frontier corresponds to a specific model and cost, but does not directly specify a generation length, since this can vary across tasks. Estimating the optimal generation length requires further analysis.

It is important to note that we do not consider inter-request resource scheduling strategies, such as early stopping or dynamic reallocation across requests~\citep{fu2024efficiently}, since we aim to ensure fairness across all inputs. Instead, the cost constraint \( C \) is interpreted as the maximum allowable cost per request (not the average), even if some requests achieve saturated accuracy below that threshold.

Under this assumption, the effective cost at any point on the frontier is determined by the task that incurs the \textit{maximum} cost. For previous scaling laws and \sparselaw studies, where the generation costs are linear to generated tokens, the optimal generation tokens is calculated with $\max_{T\in\mathcal{T}}N_T\mathbb{E}[L_{\text{out}}]$. For \law, the optimal generation tokens is calculated with $\max_{T\in\mathcal{T}}\sqrt{\mathbb{E}[L^2_{\text{out}}]}$ (we only analysis \longcot). This adjustment accounts for the quadratic dependence of cost on output length, better measuring the cost allocated to generating longer responses.

\subsection{Oracle Resource Allocation}
\label{optimalresourceallocation}
We describe the procedure for identifying oracle resource allocations and establishing the Pareto frontier for sparse attention models in~\Cref{alg:avg_acc_genlen,alg:avg_acc_select}, as a supplement to~\Cref{sec:sparse scaling laws}. Given a fixed cost constraint \( C \), we perform a grid search over key parameters: KV budgets and either reasoning trials or maximum generation lengths. 

Empirically, we sweep over KV budgets \{32, 64, 128, 256, 512, 1024\}; reasoning trials \{1, 2, 4, 8, 16, 32\} (with a reduced upper limit for the 14B and 32B models to save computation time); and generation lengths \{2\text{k}, 4\text{k}, 6\text{k}, 8\text{k}, 10\text{k}, 12\text{k}, 14\text{k}, 16\text{k}, 18\text{k}, 20\text{k}, 22\text{k}, 24\text{k}, 26\text{k}, 28\text{k}, 30\text{k}, 32\text{k}\}.

By varying the cost constraint \( C \) in~\Cref{alg:avg_acc_genlen,alg:avg_acc_select}, we obtain the performance of sparse attention models under optimal resource allocation, as shown in~\Cref{fig: bon-total,fig: bon-moe,fig: bon-32b,fig: cot-total,fig: cot-moe,fig: cot-32b,fig: sparse comparison,fig: bon-bs-frontier,fig: cot-bs-frontier}.

\begin{algorithm}
\KwData{Tasks $\mathcal{T}$, KV budgets $\{B_1, \dots, B_j\}$, trial counts $\{N_1, \dots, N_i\}$, cost limit $C$}
\KwResult{Average of maximum accuracy per task under cost $C$}
$\text{AccumBestAcc} \gets 0$\, $\text{Count} \gets 0$\;
\For{task $T$ in $\mathcal{T}$}
{
  \For{KV budget $B_b$}{
    Generate $S \ge \max\{N_1,..,N_i\}$ responses using $B_b$ for task $T$\;
    \For{trial count $N_a$}{
      compute cost $c_{b,a}^{(T)}$\;
      \If{$c_{b,a}^{(T)} \le C$}{
        Compute accuracy $\text{Acc}_{b,a}^{(T)} = \text{Pass@}N_a$;\\
        \If{$\text{Acc}_{b,a}^{(T)} > \text{BestAcc}$}{
            $\text{BestAcc} \gets \text{Acc}_{b,a}^{(T)}$;
        }
      }
    }
    }
    $\text{AccumBestAcc} \mathrel{+}= \text{BestAcc} $;
    $\text{Count} \mathrel{+}= 1$;
}
$\text{AvgBestAcc} = \text{AccumBestAcc} / \text{Count}$\;
\Return $\text{AvgBestAcc}$\;
\caption{\bon oracle resource allocation under cost $C$}
\label{alg:avg_acc_select}
\end{algorithm}

\begin{algorithm}
\KwData{Tasks $\mathcal{T}$, KV budgets $\{B_1, \dots, B_j\}$, gen. lengths $\{n_1, \dots, n_i\}$, samples $S$, cost limit $C$}
\KwResult{Average of maximum accuracy per task under cost $C$}
$\text{AccumBestAcc} \gets 0$\, $\text{Count} \gets 0$\;
\For{task $T$ in $\mathcal{T}$}
{
  $\text{BestAcc} \gets 0$\;
  \For{gen. length $n_a$}{
    \For{KV budget $B_b$}
    {
      Generate $S$ responses using $(B_b, n_a)$; compute cost $c_{b,a}^{(T)}$\;
      \If{$c_{b,a}^{(T)} \le C$}{
        Compute accuracy $\text{Acc}_{b,a}^{(T)} = \text{Pass@}1$;\\
        \If{$\text{Acc}_{b,a}^{(T)} > \text{BestAcc}$}{
            $\text{BestAcc} \gets \text{Acc}_{b,a}^{(T)}$;
        }
      }
    }
    }
    $\text{AccumBestAcc} \mathrel{+}= \text{BestAcc} $;
    $\text{Count} \mathrel{+}= 1$;
}
$\text{AvgBestAcc} = \text{AccumBestAcc} / \text{Count}$\;
\Return $\text{AvgBestAcc}$\;
\caption{\longcot oracle resource allocation under cost $C$}
\label{alg:avg_acc_genlen}
\end{algorithm}
\subsection{Top-$K$ Attention and Block Top-$K$ Attention}
\label{topkandblocktopk}
In this section, we explain the sparse attention algorithms discussed in the main paper, namely \textit{Top-$K$ Attention} and \textit{Block Top-$K$ Attention}.

During the decoding phase of a large language model (LLM), the self-attention mechanism computes a weighted average of past values as follows:
\begin{equation}
    o = \mathrm{Softmax}\left(\frac{qK^\top}{\sqrt{d}}\right)V = wV,
    \quad q \in \mathbb{R}^{1 \times d}, \quad K, V \in \mathbb{R}^{n \times d}, \quad w \in \mathbb{R}^{1 \times n},
\end{equation}
where $d$ is the head dimension and $n$ is the context length. The key and value matrices are given by $K = [k_1, k_2, \dots, k_n]$, $V = [v_1, v_2, \dots, v_n]$, where each $k_i, v_i \in \mathbb{R}^{1 \times d}$ are cached from previous decoding steps.

\textbf{Top-$K$ Attention.} Top-$K$ Attention is a sparsification method where only the $K$ most relevant tokens (i.e., those with the highest attention scores) are selected to compute the output. Formally, instead of computing the full softmax, we define a sparse attention weight vector:
\begin{equation}
    w_i =
    \begin{cases}
        \frac{\exp(s_i)}{\sum_{j \in \mathcal{I}_K} \exp(s_j)} & \text{if } i \in \mathcal{I}_K, \\
        0 & \text{otherwise},
    \end{cases}
    \quad \text{where} \quad s_i = \frac{q k_i^\top}{\sqrt{d}}, \quad \mathcal{I}_K = \text{TopK}_K(s),
\end{equation}
Here, $\mathcal{I}_K$ denotes the indices of the top $K$ attention scores $s_i$. By masking out the less important positions, this approach reduces the computational and memory cost of attention from $\mathcal{O}(n)$ to $\mathcal{O}(K)$, where $K \ll n$.

\textbf{Block Top-$K$.} Block Top-$K$ Attention is a block-level sparse attention mechanism. Instead of selecting individual tokens based on attention scores, this method selects entire blocks of tokens, thereby reducing the number of attention computations.

Specifically, assume the full sequence of $n$ keys is divided into $m = \frac{n}{\texttt{BLOCK\_SIZE}}$ consecutive blocks, each of size \texttt{BLOCK\_SIZE}:
\[
K = [k_1, \dots, k_n] \rightarrow \{K_1, K_2, \dots, K_m\}, \quad K_i \in \mathbb{R}^{\texttt{BLOCK\_SIZE} \times d}
\]

For each block $K_i$, we first compute the average key vector:
\[
\bar{k}_i = \frac{1}{\texttt{BLOCK\_SIZE}} \sum_{j=1}^{\texttt{BLOCK\_SIZE}} k_{i,j}
\]

Next, we compute the attention score between the query $q$ and each block’s average key:
\[
s_i = \frac{q \bar{k}_i^\top}{\sqrt{d}}, \quad \text{for } i = 1, 2, \dots, m
\]

We then select the top $K' = \frac{K}{\texttt{BLOCK\_SIZE}}$ blocks based on the scores $s_i$, denoted by the index set $\mathcal{J}_{K'} = \text{TopK}_{K'}(s)$. Attention is computed only over the tokens within the selected blocks. The sparse attention weights are defined as:
\[
w_i =
\begin{cases}
\frac{\exp(s_i)}{\sum_{j \in \mathcal{I}_K} \exp(s_j)} & \text{if } i \in \mathcal{I}_K \subseteq \text{tokens in selected blocks}, \\
0 & \text{otherwise}
\end{cases}
\]

For both algorithms, $K$ is the KV budget. For GQA, we conduct an average pooling across all the query heads in a group, ensuring that the total number of retrieved key-value vectors does not exceed the allocated KV budget.

\section{Extended Related Work}
\label{relatedwork}
\textbf{Efficient Attention.} 
Sparse attention~\citep{kitaev2020reformer,zandieh2023kdeformer,chen2021scatterbrain,chen2024magicpig,zhang2023h2o,xiao2024efficientstreaminglanguagemodels,yuan2025native,nawrot2025sparse,child2019generating,li2024snapkvllmknowslooking,cai2024lococo,mazare2025inference} has been comprehensively studied 
to reduce the attention cost when processing long sequeces. In parallel, approaches like FlashAttention~\citep{flash_attn,flash_attn2} accelerate attention by maximizing hardware efficiency. To address the quadratic complexity of standard attention, researchers have also explored linear attention architectures~\citep{mamba,s4,linear_attention,choromanski2020rethinking}. Additionally, quantization and low-precision methods~\citep{liu2024kivi,NEURIPS2024_028fcbcf,lin2024qserve} have been broadly applied for improving inference efficiency.

\textbf{Efficient Inference.} Orca~\citep{280922}, vLLM~\citep{kwon2023efficientmemorymanagementlarge}, and SGLang~\citep{zheng2024sglang} are widely adopted to enhance the efficiency of LLM serving. Our analysis builds on the practical designs and implementations of these systems. In parallel, speculative decoding~\citep{leviathan2023fast,chen2023accelerating,miao2023specinfer,sadhukhan2024magicdec} has been proposed to mitigate the memory-bandwidth bottleneck during LLM decoding. Additionally, model compression and offloading~\citep{dettmers2022gpt3,lin2024awq,svirschevski2024specexec,sheng2023flexgen,frantar2022gptq} techniques are playing a crucial role in democratizing LLM deployment.

\textbf{Efficient Test-time Strategies.} Optimizing reasoning models to generate fewer tokens has been shown to directly reduce inference-time cost~\citep{sky,arora2502training,ma2025cot}. Recent work such as CoCoNut~\citep{hao2024training} and CoCoMix~\citep{tack2025llm} explores conducting reasoning in a latent space, thereby reducing decoding time. Methods like ParScale~\citep{ParScale}, Tree-of-Thoughts~\citep{yao2023tree}, and Skeleton-of-Thoughts~\citep{ning2023skeleton} aim to improve efficiency by enabling parallel reasoning. Architectural innovations such as CoTFormer~\citep{mohtashami2023cotformer} further enhance efficiency by adaptively allocating computational resources across tokens. Efficient reward-model-based~\citep{wu2024inference,snell2024scaling,sun2024fast} test-time scaling algorithms are also comprehensively studied. 
\end{document}